\documentclass[12pt,journal,draftcls,a4paper,twoside,onecolumn]{IEEEtran} 
\usepackage{setspace}
\usepackage{amsmath}                        
\usepackage{graphicx}
\usepackage{amssymb}
\usepackage{cite}
\usepackage{algorithmic}
\usepackage[linesnumbered,lined,ruled]{algorithm2e}
\usepackage{array}
\usepackage{stfloats}
\usepackage{subfigure}
\usepackage{multicol}
\usepackage[all,poly]{xy}
\usepackage{multirow}
\usepackage{color}
\usepackage{graphicx,psfrag,color}
\usepackage{amsthm}

\def\bfn{{\mathbf{n}}}

\def\bfp{{\mathbf{p}}}

\def\bfr{{\mathbf{r}}}

\def\bfx{{\mathbf{x}}}
\def\bfy{{\mathbf{y}}}
\def\bfz{{\mathbf{z}}}

\def\bfH{{\mathbf{H}}}
\def\bfI{{\mathbf{I}}}

\def\bbR{{\mathbb{R}}}

\def\bsr{{\boldsymbol{r}}}

\newcounter{algo}
\renewcommand{\thealgo}{\arabic{algo}}

\def\bstheta{{\boldsymbol{\theta}}}
\def\bsxi{{\boldsymbol{\xi}}}
\def\bsgamma{{\boldsymbol{\gamma}}}

\newlength{\tempdima}

\newcommand{\rowname}[1]
{\rotatebox{90}{\makebox[\tempdima][c]{\textbf{#1}}}}

\newcommand{\zhao}[1]{\textcolor[rgb]{0.00,0.00,0.00}{#1}}
\newcommand{\zhaoone}[1]{\textcolor[rgb]{0.00,0.00,0.00}{#1}}
\newcommand{\rev}[1]{\textcolor[rgb]{0.00,0.00,0.00}{#1}}

\newcommand{\revAQ}[1]{\textcolor[rgb]{0.00,0.00,0.00}{#1}}

\begin{document}
\title{Joint Segmentation and Deconvolution of Ultrasound Images Using a Hierarchical Bayesian Model based on Generalized Gaussian Priors}

\author{\IEEEauthorblockN{Ningning Zhao, Adrian Basarab, Denis Kouam{\'e} and Jean-Yves Tourneret} 

\thanks{Ningning Zhao and Jean-Yves Tourneret are with University of Toulouse, IRIT/INP-ENSEEIHT, Toulouse, France (e-mail: \{nzhao, jean-yves.tourneret\}@enseeiht.fr). 

Adrian Basarab and Denis Kouam{\'e} are with University of Toulouse, IRIT, CNRS UMR 5505, 118 Route de Narbonne, F-31062, Toulouse Cedex 9, France (e-mail: \{adrian.basarab, denis.kouame\}@irit.fr).} }

\maketitle
\begin{abstract}
\rev{This paper proposes a joint segmentation and deconvolution Bayesian method for medical ultrasound (US) images. 
Contrary to piecewise homogeneous images, US images exhibit heavy characteristic speckle patterns correlated with the tissue structures. 
The generalized Gaussian distribution (GGD) has been shown to be one of the most relevant distributions for 
characterizing the speckle in US images. Thus, we propose a GGD-Potts model defined by a label map coupling US image segmentation and deconvolution.} 
The Bayesian estimators of the unknown model parameters, including the US image, the label map and all the hyperparameters are difficult to be expressed in closed form. Thus, we investigate a Gibbs sampler to generate samples distributed according to the posterior of interest. 
These generated samples are finally used to compute the Bayesian estimators of the unknown parameters. 
The performance of the proposed Bayesian model is compared with existing approaches via several experiments conducted on realistic synthetic data and \textit{in vivo} ultrasound images. 
\end{abstract}


\section{Introduction}
\IEEEPARstart{U}{ltrasound} (US) imaging is a well-established medical imaging modality widely used for clinical diagnosis, 
visualization of anatomical structures, tissue characterization and blood flow measurements. 
The popularity of US imaging compared to other imaging modalities such as computed tomography (CT) or magnetic resonance imaging (MRI) is mainly due to its efficiency, low cost and safety \cite{Atam2011}. Despite these advantages and the recent advances in instrumentation \cite{tanter2014ultrafast} and beam-forming \cite{rindal2014understanding}, it also has some limitations, mainly related to its poor signal-to-noise ratio, limited contrast and spatial resolution.
Furthermore, US images are characterized by speckle, which considerably reduces their quality and may lead to interpretation issues. 
For this reason, several despeckling methods can be found in the US literature \cite{Gungor2015, Michailovich2006}.
Despite its negative effect, speckle has also been extensively used as a source of information in applications such as image segmentation and tissue characterization \cite{Noble2006_Survey,Pereyra2012}. 
Specifically, it has been shown that the statistical properties of the speckle are strictly correlated with the 
tissue structures \cite{MartinoAlessandrini2011, Szabo_book}. 
Thus, methods allowing image restoration using the statistical properties of the speckle noise are also an interesting research track in US imaging \cite{JamesNg2007, MartinoAlessandrini2011}.

\subsection{\rev{Problem Statement}}
Under the first order Born approximation and the assumption of weak scattering classically assumed for soft tissues \cite{Jensen1993}, 
the radio-frequency (RF) US images can be modeled as the convolution between a blurring operator/point spread function (PSF) and a tissue reflectivity function (TRF), see, e.g., \cite{Jensen1993,Taxt1995,Michailovich2007,JamesNg2007,MartinoAlessandrini2011}. 
The resulting linear model is given by
\begin{equation}
y(\bsr) = h(\bsr) \otimes x(\bsr) + n(\bsr), \; \; \bsr\in \mathcal{R}
\label{eq:linearmodelR}
\end{equation}
where $\otimes$ is the two dimensional convolution operator, $y(\bsr)$ is the observed image pixel \textcolor{black}{at the location $\bsr$}, $x(\bsr)$ is the TRF to be estimated, $h(\bsr)$ is the system PSF, $n(\bsr)$ is the measurement noise and $\mathcal{R}$ is the image domain. Equivalently, after lexicographical ordering the corresponding images $y(\bsr)$, $x(\bsr)$, $n(\bsr)$ and 
forming the huge matrix $\bfH \in \bbR^{N \times N}$ associated with $h(\bfr)$, we obtain the following equivalent model
\begin{equation}
\bfy = \bfH\bfx + \bfn.
\label{eq:linearmodel}
\end{equation}
\rev{Due to the physical corrections related to image formation (e.g., time gain compensation, dynamic beamforming), in most of soft tissues, $h(\bsr)$ can be assumed shift invariant. 
\revAQ{Moreover, cyclic convolution is considered in this paper for computational purpose,} leading to a block circulant matrix of circulant blocks (BCCB) $\bfH$\footnote{\rev{Some existing works \cite{JamesNg2007,Michailovich2005,MartinoAlessandrini2011,NageOLeary1998} assume that the PSF in US imaging is shift-variant mainly along the axial direction. In this case, US images are generally divided into several local regions along the axial direction. In each region, the local PSF is assumed shift-invariant. The global blurring matrix is built in this case by combining these local shift-invariant PSFs.}}.} 
Note that the PSF is unknown in practical applications and that its estimation has been extensively explored in US imaging.
A typical approach in US imaging, also adopted in this paper, is to estimate the PSF in a pre-processing step before applying the deconvolution algorithm (see, e.g., \cite{MartinoAlessandrini2011}, \cite{Jensen1993}).

\subsection{\rev{Related Work}}
US image deconvolution aims at estimating the TRF $\bfx$ from the RF data $\bfy$, which is a typical ill-posed problem.
Imposing a regularization constraint is one traditional way to cope with this problem. The regularization constraint usually reflects the prior knowledge about $\bfx$. In US imaging, Gaussian and Laplacian distributions have been widely explored as prior information for the TRF $\bfx$, leading to $\ell_2$-norm \cite{Jirik2008} and $\ell_1$-norm \cite{Michailovich2007}, \cite{Yu2012} constrained optimization problems. 

\rev{Due to the tight relationship between image deconvolution and segmentation, it is interesting to consider these two operations jointly. This idea has been recently exploited for piecewise homogeneous images using the Mumford-Shah model \cite{Bar2004,MumfordShah2011,Chan2014}, the Potts model \cite{Ayasso2010,Storath2015} or the generalized linear models \cite{Paul2013} in Bayesian or variational frameworks. 
Moreover, segmentation-based regularizations have been considered in \cite{Mignotte2006} to improve the image reconstruction performance. 
However, due to the intrinsic granular appearance of US data, these methods are not efficient to simultaneously 
restore and segment US images. In order to develop US image \rev{deconvolution and segmentation} methods, it is common to take advantage of the statistical properties of the TRF. \rev{Except the traditional Gaussian and Laplace distributions mentioned above,} distributions that have been considered for US images include the homodyned K \cite{Hruska2009}, Nakagami \cite{LarrueNaka2011} and generalized Gaussian distributions \cite{Bernard2007}.}
Alessandrini \textit{et. al.} recently investigated a deconvolution method for US images based on generalized Gaussian distributions (GGDs) using the expectation maximization (EM) algorithm \cite{MartinoAlessandrini2011_AI, MartinoAlessandrini2011}. This method assumed that the US image can be divided into different regions characterized by GDDs with different parameters. 
Despite its accuracy when compared to several state-of-the-art US image deconvolution methods, the framework in \cite{MartinoAlessandrini2011_AI} has two major drawbacks that we propose to tackle in this paper. First, it is well-known that the EM algorithm can easily converge to a local minimum of the cost function and is sensible to the initial values of the parameters to be tuned, which may lead to inaccurate estimates. Second, the EM algorithm can only be applied to cases where a mask (or label map) 
of the homogeneous regions is available. \rev{Note that a US image deconvolution method based on Markov chain Monte Carlo (MCMC) methods was recently investigated in \cite{NZHAO2015}. However, the proposed method was also using an \textit{a priori} label map for the different image regions. 
Due to the tight relationships between segmentation and deconvolution, we think that combining these two operations can increase their performance, which is the objective of this paper.}

\subsection{\rev{Proposed method}}
\rev{Compared with the US image deconvolution method of \cite{NZHAO2015}, this paper defines a Potts Markov random field for the hidden image labels, assigns GGD priors to the image TRF, and investigates a joint segmentation and deconvolution method for US images. \rev{Thus, the proposed algorithm generalizes the results of \cite{NZHAO2015} to situations where a label map is unknown.}}
\rev{Additional motivations for the proposed model are provided below.
First, it uses \rev{a GGD-Potts model} to regularize the ill-posed joint deconvolution \rev{and segmentation} problem. 
Second, \rev{it exploits the local statistical properties of different image regions, which are usually related with the anatomical image structures.} 
Finally, the proposed model is able to capture the spatial correlations between neighboring pixels.}
To our knowledge, the proposed method represents a first attempt for a joint segmentation and deconvolution in US imaging.
The complicated form of the resulting posterior distribution makes it too difficult to compute closed form expressions of the corresponding Bayesian estimators. Therefore, a Markov chain Monte Carlo (MCMC) method based on a Gibbs sampler is investigated to sample the posterior distribution of interest and build the estimators of its unknown parameters. 

The rest of the paper is organized as follows. The statistical hierarchical Bayesian model proposed for image segmentation and deconvolution is introduced in Section II.  Section III studies a hybrid Gibbs sampler, which generates samples asymptotically distributed according to the posterior distribution of this model. Simulation results obtained on synthetic data, realistic simulated and \textit{in vivo} US images are presented in Section IV. Conclusions are finally reported in Section V.

\section{\rev{Bayesian Model for Joint Deconvolution and Segmentation}}

\rev{This section introduces the Bayesian model investigated in this paper for the joint deconvolution and segmentation of US images.}
We assume that the US TRF $\bfx=(x_1,\cdots,x_N)^T$ can be divided into $K$ statistical homogeneous regions, denoted as $\{\mathcal{R}_1,...,\mathcal{R}_K\}$ and we introduce a hidden label field $\bfz=(z_1,\cdots,z_N)^T \in \mathbb{R}^N$ mapping the image into these $K$ regions. 
More precisely, $z_i=k$ if and only if the corresponding pixel $x_i$ belongs to the region $\mathcal{R}_k$, where $k\in\{1,\cdots,K\}$ and $i\in\{1,\cdots,N\}$. The conditional distribution of pixel $x_i$ is then defined as
\begin{equation}
x_i|z_i=k \sim \mathcal{GGD}(\xi_k,\gamma_k)
\label{prior_x_v1}
\end{equation}
where $\xi_k$ and $\gamma_k$ are the shape and scale parameters of the GGD associated with the region $\mathcal{R}_k$. \revAQ{We remind that a univariate GGD with shape parameter $\xi$ and scale parameter $\gamma$ denoted as $\mathcal{GGD} (\xi, \gamma)$ has the following pdf,}
\begin{equation}
p(x) = \frac{1}{2\gamma^{1/\xi}\Gamma(1+1/\xi)} \exp\left(-\frac{{|x|^{\xi}}}{\gamma} \right), \quad x \in \mathbb{R}.
\end{equation}

Assuming that the pixels are independent conditionally to the knowledge of their classes, the TRF is distributed according to a mixture of GGDs with the following probability density function (pdf)
\begin{equation}
p(x_i) = \sum_{k=1}^K w_k \mathcal{GGD}(\xi_k,\gamma_k) \;\;\textrm{with} \;\; w_k=P(z_i = k).
\end{equation}
In addition, we assign a Potts model to the hidden field $\bfz$ to exploit the dependencies between pixels that are nearby in the image \cite{Ayasso2010,Pereyra2012,Pereyra2013}. The resulting model is referred to as GGD-Potts model. In the following, we define a hierarchical Bayesian model based on this 
GGD-Potts model for the joint segmentation and deconvolution of US images. Using the Bayes rule for the joint posterior of the \revAQ{unknown parameters,}
the following result can be obtained
\revAQ{
\begin{equation}
p(\bfx,\bfz,\bstheta|\bfy) \propto p(\bfy|\bfx,\bstheta)p(\bfx|\bfz,\bstheta)p(\bfz|\bstheta)p(\bstheta)
\label{Bayes}
\end{equation}
}
where $\varpropto$ means ``proportional to", $\bstheta$ is a parameter vector containing all the model parameters and hyperparameters except $\bfx$ and $\bfz$, \revAQ{\textit{i.e.,} the noise variance, the shape and scale parameters}. The likelihood $p(\bfy|\bfx,\bstheta)$ depending on the noise model and the prior distributions $p(\bfx|\bfz,\bstheta)$, $p(\bfz|\bstheta)$ based on the GGD-Potts model are detailed hereinafter.

\subsection{Likelihood}
Assuming an additive white Gaussian noise (AWGN) with a constant variance $\sigma_{n}^{2}$, the likelihood function associated with the linear model \eqref{eq:linearmodel} is  
\begin{equation}
p(\bfy|\bfx, \sigma_n^2)=\frac{1}{(2\pi\sigma_{n}^{2})^{N/2}}
\textrm{exp}\left(-\frac{1}{2\sigma_{n}^{2}}\|\bfy-\bfH\bfx\|_{2}^{2}
\right)
\end{equation}
where $\|\cdot\|_2$ is the \zhaoone{Euclidean} $\ell_2$-norm.

\subsection{Prior Distributions}
\subsubsection{Tissue reflectivity function (TRF) $\bfx$}
As explained beforehand, a mixture of GGD priors is assigned to the TRF.
Assuming that the pixels are independent conditionally to the knowledge of their classes, we obtain the following prior for the target image
\begin{align}
p(\bfx|\bfz,\bsxi,&\bsgamma) 
= \prod_{k=1}^K \prod_{i=1}^{N_k}\frac{1}{2\gamma_{k}^{1/\xi_k}\Gamma(1+1/\xi_k)} \exp\left(-\frac{{|x_i|^{\xi_k}}}{\gamma_{k}} \right) \notag\\
 = &\prod_{k=1}^K  \frac{1} {{\left[2\gamma_{k}^{1/\xi_{k}}\Gamma(1+1/\xi_{k})\right]}^{N_k}} \exp\left(-\frac{\sum_{i=1}^{N_k}|x_{i}|^{\xi_k}}{\gamma_k} \right)\notag\\
 =& \prod_{k=1}^K \frac{1} {{\left[2\gamma_{k}^{1/\xi_{k}}\Gamma(1+1/\xi_{k})\right]}^{N_k}} \exp\left(-\frac{\|\bfx_{k}\|_{\xi_k}^{\xi_k}}{\gamma_k} \right)
 \label{eq:prior_x}
\end{align}
where $\bsxi=(\xi_1,\cdots,\xi_K)^T$ and $\boldsymbol{\gamma}=(\gamma_1,\cdots,\gamma_K)^T$, $\xi_k$ and $\gamma_k$ are the shape and scale parameters of the $k$th region $\mathcal{R}_k$, $N_k$ is the number of pixels in $\mathcal{R}_k$, $\bfx_{k}$ contains all the pixels assigned to $\mathcal{R}_k$, $\Gamma(\cdot)$ is the gamma function and $\|\bfx_k\|_{\xi}=(\sum_{i=1}^{N_k}|x_i|^{\xi})^{1/\xi}$ denotes the $\ell_{\xi}$-norm. 
\subsubsection{Noise variance $\sigma_n^2$}
In the presence of an AWGN, it is standard to assign a conjugate inverse gamma ($\mathcal{IG}$) prior to the noise variance, i.e.,
\begin{align}
p(\sigma_n^2) \sim& \mathcal{IG}(\alpha,\nu) \notag\\
=&\frac{\nu^{\alpha}}{\Gamma(\alpha)} (\sigma_n^2)^{-\alpha-1} \exp\left( -\frac{\nu}{\sigma_n^2} \right) \mathcal{I}_{\mathbb{R+}}(\sigma_n^2)
\end{align}
where $\mathcal{I}_{A}$ is the indicator function on the set $A$. This prior has two adjustable parameters $\alpha$, $\nu$ which make it very flexible and thus appropriate to the variance of most statistical models. 
The values of $\alpha$ and $\nu$ have been fixed by cross validation in our experiments leading to $(\alpha, \nu)=(0.1,0.1)$.

\subsubsection{Labels \zhaoone{$\bfz$}}
A Potts model (generalization of the Ising model) is considered as prior for the hidden image label field. 
The Potts Markov random field (MRF) has been shown to be appropriate for image segmentation \cite{Murray2006,Pereyra2013}. 
It establishes dependencies between pixels that are nearby in an image \cite{Pereyra2012,Pereyra2013}.
More specifically, adjacent labels of the image are dependent and tend to belong to the same class. The conditional distribution of $z_n$ (associated with pixel $x_n$) for the Potts MRF is defined as
\begin{equation}
p(z_n|\bfz_{-n})=p(z_n|\bfz_{\mathcal{V}(n)})
\label{eq:neighbors}
\end{equation}
where $\bfz_{-n}=(z_1,...,z_{n-1},z_{n+1},...,z_N)$ and $\mathcal{V}(n)$ contains the neighbors of label $z_n$. In this paper, a first order neighborhood structure (\zhaoone{i.e., 4 nearest pixels}) is considered.
The whole set of random variables $\bfz$ forms a random field.

Using the Hammersley-Clifford theorem \cite{Besag1974}, the prior of $\bfz$ can be expressed as a Gibbs distribution, i.e.,
\begin{equation}
p(\bfz)=\frac{1}{C(\beta)}\exp\left[\sum_{n=1}^N \sum_{n' \in\mathcal{V}(n)} \beta \delta(z_n-z_{n'})\right]
\end{equation}
where $\beta$ is the granularity coefficient or smooth parameter, $\delta(\cdot)$ is the Kronecker function and $C(\beta)$ is the normalizing constant (often referred to as partition function). The value of $\beta$ has been fixed by cross validation, leading to $\beta = 1$. 

\subsubsection{Shape and scale parameters}
The prior used for the US TRF defined in \eqref{eq:prior_x} depends on the shape and scale parameters of the GGD, which are usually referred to as hyperparameters. Following the works in \cite{Lotfi2010}, we have chosen the following priors for these hyperparameters
\begin{align}
p(\bsxi)&= \prod_{k=1}^K  p(\xi_k) = \prod_{k=1}^K \frac{1}{3} \mathcal{I}_{[0,3]}(\xi_k)\\
p(\bsgamma)&=\prod_{k=1}^K p(\gamma_k) =  \prod_{k=1}^K \frac{1}{\gamma_k}\mathcal{I}_{\mathbb{R+}}(\gamma_k)
\end{align}
where $k\in \{1,...,K\}$. Note that the range $[0,3]$ covers all 
the possible values of $\xi_k$ and that $p(\gamma_k)$ is the uninformative Jeffreys prior for $\gamma_k$. 
\subsection{Joint posterior distribution}
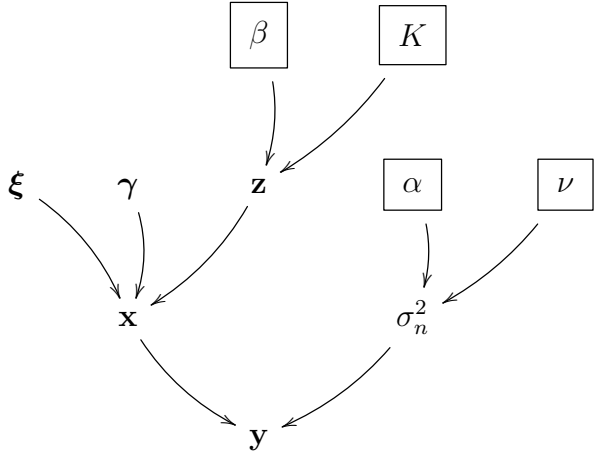
\begin{figure}[h!]
\centerline{
\xymatrix{
 & & *+<0.05in>+[F-]+{\beta} \ar@/^/[d] & *+<0.05in>+[F-]+{K} \ar@/^/[ld]  & &  & \\
 \bsxi \ar@/^/[rd] & \bsgamma \ar@/^/[d] & \bfz \ar@/^/[ld]   & *+<0.05in>+[F-]+{\alpha} \ar@/^/[d] & *+<0.05in>+[F-]+{\nu} \ar@/^/[ld]  &\\
 & \bfx \ar@/_/[rd]   & &  \sigma_n^2\ar@/^/[ld] \\
 & & \bfy & &
 }
}
\caption{\rev{Hierarchical Bayesian model} for the parameter and hyperparameter priors, \rev{where the TRF $\bfx$ is modeled by a mixture of GGDs, the hidden label field $\bfz$ follows a Potts MRF and the parameters appearing in the boxes are fixed in advance.}}
\label{fig:DAG}
\end{figure}

The joint posterior distribution of the unknown parameters
{$\bfx,\sigma_n^2,\bsxi,\bsgamma,\bfz$} can be determined as follows
\begin{eqnarray}
p(\bfx,\sigma_n^2,\bsxi,\bsgamma,\bfz|\bfy)
&\varpropto & p(\bfy|\bfx,\sigma_n^2,\bsxi,\bsgamma,\bfz)p(\bfx,\sigma_n^2,\bsxi,\bsgamma,\bfz) \notag  \\
&\varpropto & p(\bfy|\bfx,\sigma_n^2,\bsxi,\bsgamma,\bfz)p(\bfx|\bsxi,\bsgamma,\bfz)p(\sigma_n^2) \notag  \\
&&\times p(\bsxi)p(\bsgamma)p(\bfz) \notag  \\
&\varpropto & \frac{1}{(2\pi\sigma_{n}^{2})^{N/2}}\exp\left(-\frac{1}{2\sigma_n^2}\|\bfy-\bfH\bfx\|_2^2\right) 
\times\frac{1}{(\sigma_{n}^{2})^{\alpha+1}} \exp\left(-\nu/\sigma_n^2\right)   \notag\\    
&&\times \prod_{k=1}^K \Bigg\{ a_k^{N_k}  \exp\left(-\frac{\|\bfx_{k}\|_{\xi_k}^{\xi_k}}{\gamma_k} \right)  
\times \exp\bigg[\sum_{n=1}^N \sum_{n' \in\mathcal{V}(n)} \beta \delta(z_n-z_{n'})\bigg]   \notag\\
&&\times \frac{1}{3}\mathcal{I}_{[0,3]}(\xi_k) \frac{1}{\gamma_k} \mathcal{I}_{\mathbb{R+}}(\gamma_k)  \Bigg\} 
\label{eq:jointpost1}
\end{eqnarray}
where $a_k= \frac{1} {2\gamma_{k}^{1/\xi_{k}}\Gamma(1+1/\xi_{k})}$ and the hyperparameters are supposed to be \textit{a priori} independent.
Fig. \ref{fig:DAG} summarizes the proposed hierarchical Bayesian model as a directed acyclic graph (DAG), in which the relationships between the parameters and hyperparameters are indicated.

\section{\rev{Sampling the posterior and computing  the Bayesian estimators}}
Computing closed-form expressions of the MMSE or MAP estimators for the unknown parameters $\bfx, \sigma_n^2, \bsxi, \bsgamma, \bfz$ from \eqref{eq:jointpost1} is clearly complicated. In this case, a possible solution is to consider MCMC methods in order to generate samples asymptotically distributed according to the distribution of interest and to use the generated samples to build estimators of the unknown parameters. \rev{In this section, a hybrid Gibbs sampler is investigated to generate samples asymptotically distributed according to \eqref{eq:jointpost1}. These samples are used to compute the Bayesian estimators of the US TRF $\bfx$, hidden label field $\bfz$, noise variance $\sigma_n^2$ and GGD parameters $\bsxi, \bsgamma$.}


\subsection{\rev{Hybrid Gibbs sampler}}
The proposed hybrid Gibbs sampler is a 5-step algorithm summarized in Algorithm \ref{algrithm1}. The algorithm is explained in detail in what follows.
\begin{algorithm}
\label{algrithm1}
Sampling the noise variance $\sigma_n^2$ according to the conditional distribution \eqref{eq:post_noise}.

Sampling the shape parameter $\bsxi$ according to the conditional distribution
\eqref{eq:shape_conditional} with an RWMH algorithm.

Sampling the scale parameter $\bsgamma$ using \eqref{eq:scale_conditional}.

Sampling the labels $\bfz$ according to the normalized conditional distribution \eqref{norm_post_label}.

Sampling the TRF $\bfx$ using an HMC method.
\caption{Hybrid Gibbs Sampler}
\end{algorithm}
\subsubsection{Sampling the noise variance}
The conditional distribution of $\sigma_n^2 | \bfy,\bfx,\bsxi,\bsgamma,\bfz$ is the following inverse gamma distribution whose expression is derived in Appendix \ref{app:IG}
\begin{align}
p(\sigma_n^2|\bfy,\bfx,\bsxi,\bsgamma,\bfz) & \varpropto p(\bfy|\bfx,\sigma_n^2,\bsxi,\bsgamma,\bfz) p(\sigma_n^2) \notag\\
= \mathcal{IG}& \left(\alpha+N/2,\theta+\frac{1}{2}\|\bfy-\bfH\bfx\|^2_2\right).
\label{eq:post_noise}
\end{align}
Generating samples according to \eqref{eq:post_noise} is straightforward.
\subsubsection{Sampling the shape parameter vector $\bsxi$}
The conditional distribution of the shape parameter vector $\bsxi$ satisfies the following relation 
\begin{eqnarray}
p(\bsxi|\bfy,\bfx,\sigma_n^2,\bsgamma,\bfz) 
&\varpropto & p(\bfy|\bfx,\sigma_n^2,\bsxi,\bsgamma,\bfz)p(\bfx|\bsxi,\bsgamma,\bfz)p(\bsxi) \notag\\
 &\varpropto & p(\bfx|\bsxi,\bsgamma,\bfz)p(\bsxi). 
\label{post_xi1}
\end{eqnarray}
Assuming that the shape parameters are \textit{a priori} independent, we have 
\begin{align}
p(\xi_k|\bfx,\bsgamma,\bfz,\bsxi_{-k})\varpropto p(\bfx_k| \xi_k,\gamma_k, \bfz_k) p(\xi_k)  \notag\\
\varpropto a_k^{N_k} \exp\left(-\frac{\|\bfx_k\|_{\xi_k}^{\xi_k}}{\gamma_k} \right) \mathcal{I}_{[0,3]}(\xi_k)
\label{eq:shape_conditional}
\end{align}
where $\bsxi_{-k}=(\xi_1,...,\xi_{k-1}, \xi_{k+1},...,\xi_K)$ for $k \in \{1,...,K\}$, $\bfx_k$ contains the pixels 
belonging to class $k$ and $\bfz_k$ is built from the corresponding labels.
Unfortunately, the conditional distribution \eqref{eq:shape_conditional} is not easy to sample directly. Thus, we propose to consider a random walk
Metropolis Hastings (RWMH) move \cite{Hastings1970}. 
\rev{More implementation details about this move and the resulting algorithm are given in Appendix \ref{app:RWMH}.} 

\subsubsection{Sampling the scale parameter vector $\bsgamma$} 
The conditional distribution of the scale parameter vector $\bsgamma$ satisfies the following relation
\begin{eqnarray}
p(\bsgamma|\bfy,\bfx,\sigma_n^2,\bsxi,\bfz) 
&\varpropto & p(\bfy|\bfx,\sigma_n^2,\bsxi,\bsgamma,\bfz)p(\bfx|\bsxi,\bsgamma,\bfz)p(\bsgamma) \notag\\
&\varpropto & p(\bfx|\bsxi,\bsgamma,\bfz)p(\bsgamma).
\label{post_scale1}
\end{eqnarray}
Assuming that the scale parameters are independent, we have 
\begin{align}
p(\gamma_k |\bfx,\bsxi,\bfz,\bsgamma_{-k}) 
&\varpropto p(\bfx_k| \xi_k,\gamma_k, \bfz_k) p(\gamma_k)  \notag\\
&\varpropto  \mathcal{IG} \left(\frac{N_k}{\xi_k},\|\bfx_k\|_{\xi_k}^{\xi_k}\right)
\label{eq:scale_conditional}
\end{align}
where $\bsgamma_{-k}=(\gamma_1,...,\gamma_{k-1} , \gamma_{k+1},...,\gamma_K)$ for $k \in \{1,...,K\}$.
Drawing samples from the inverse gamma distribution \eqref{eq:scale_conditional} is straightforward.
\rev{More details about the derivation of \eqref{eq:scale_conditional} are provided in Appendix \ref{app:IG}.}
\subsubsection{Sampling the labels $\bfz$}
The conditional distribution of the labels $\bfz$ can be computed using Bayes rule
\begin{eqnarray}
p(\bfz|\bfy,\bfx,\sigma_n^2,\bsxi,\bsgamma )&\varpropto& p(\bfy|\bfx,\sigma_n^2,\bsxi,\bsgamma,\bfz)p(\bfx|\bsxi,\bsgamma,\bfz)p(\bfz) \notag\\
&\varpropto& p(\bfx|\bsxi,\bsgamma,\bfz)p(\bfz).
\label{post_label2}
\end{eqnarray}
Considering the dependency between a label and its neighbors, the conditional distribution of the label $z_n$ (corresponding to the image pixel $x_n$) is given as follows
\begin{equation}
p(z_n=k|\bfz_{-n},\bfx,\bsxi,\bsgamma)
\varpropto p(x_n|z_n=k,\bsxi,\bsgamma)p(z_n=k|\bfz_{\mathcal{V}(n)})
\label{eq:post_label}
\end{equation} 
where $\bfz_{-n}$ is the vector $\bfz$ whose $n$th element has been removed and $\bfz_{\mathcal{V}(n)}$ represents the neighbors of label $z_n$. 
Note that a $4$-pixel neighborhood structure has been adopted in this paper.
Denoting the left hand side of \eqref{eq:post_label} as $\pi_{n,k}$, we have
\begin{equation}
\pi_{n,k} \varpropto a_k \exp\left(-\frac{|x_n|^{\xi_k}}{\gamma_k}\right)\exp\left( \sum_{n'\in\mathcal{V}(n)}\beta\delta(k-z_{n'}) \right).
\label{post_label}
\end{equation}
The normalized conditional probability of the label $z_n$ is defined as
\begin{equation}
\tilde{\pi}_{n,k}=\frac{\pi_{n,k}}{\sum_{k=1}^K \pi_{n,k}}.
\label{norm_post_label}
\end{equation}
Finally, the label $z_n$ can be drawn from the set $\{1,...,K\}$ with the respective probabilities $\{\tilde{\pi}_{n,1},...,\tilde{\pi}_{n,K}\}$.

\subsubsection{Sampling the TRF $\bfx$}
The conditional distribution of the target image we want to estimate is defined as follows
\begin{equation}
p(\bfx|\bfy,\sigma_n^2,\bsxi,\bsgamma,\bfz)\varpropto
\exp\left(-\frac{\|\bfy-\bfH\bfx\|_2^2}{2\sigma_n^2}-\sum_{k=1}^K \frac{\|\bfx_k\|_{\xi_k}^{\xi_k}}{\gamma_k}\right).
\label{eq:xpost}
\end{equation}
Sampling according to \eqref{eq:xpost} is the critical point of the proposed algorithm. Due to the high dimensionality of $\bfx$, classical Gibbs or MH moves are inefficient. Thus we propose to implement an efficient sampling strategy referred to as Hamiltonian Monte Carlo (HMC) method. The principles of this method have been presented in \cite{Neal2011} with an application to neural networks. 
\zhao{It is widely reported that HMC generally outperforms other standard Metropolis-Hastings algorithms, particularly in high-dimensional scenarios \cite{hoffman-gelman2013}. This empirical observation is in agreement with recent theoretical studies showing that HMC has better scaling properties than the Metropolis adjusted Langevin algorithm (MALA) and RWMH \cite{beskos2013}.}
\rev{The main steps of the HMC method with details about the way to adjust its parameters are reported in Appendix \ref{app:HMC}.}

\subsection{Parameter estimation}
Bayesian estimators of the unknown parameters are computed using the generated samples obtained by the hybrid Gibbs sampler. 
Since the labels are discrete variables, marginal MAP estimators are chosen for the labels. 
The MMSE estimators for the other variables (the TRF $\bfx$, noise variance $\sigma_{n}^2$ and GGD parameters $\bsxi$, $\bsgamma$) are calculated. 
For example, the MMSE estimator of the TRF $\bfx$ is computed by
\begin{equation}
\hat{\bfx}_{\textrm{MMSE}}|\hat{\bfz}_{\textrm{MAP}} \triangleq E\{\bfx |\bfz = \hat{\bfz}_{\textrm{MAP}} \} = \int p(\bfx|\bfz = \hat{\bfz}_{\textrm{MAP}})d \bfx.
\end{equation}
For each pixel, we can approximate this estimator as follows
\begin{equation}
\hat{x}_{n,\textrm{MMSE}}|\hat{z}_{n,\textrm{MAP}} \simeq \frac{1}{M} \sum_{i=1}^M x_n^{(i)}|z_n^{(i)}=\hat{z}_{n,\textrm{MAP}}
\end{equation}
where $M$ is the number of iterations after the so-called burn-in period (see Section \ref{sec:converge} devoted to the sampler convergence for more details) that satisfy $z_n^{(i)}=\hat{z}_{n,\textrm{MAP}}$, the superscript $i$ represents the $i$th generated sample and the subscript $n$ is used for the $n$th pixel. Note that $\hat{\bfz}_{\textrm{MAP}}$ is the marginal MAP estimator of the label map and that $\hat{\bfx}_{\textrm{MMSE}}$ is the MMSE estimator of the reflectivity. Note also that a similar estimator was implemented in \cite{GKail2012} for image blind deconvolution.

\rev{\subsection{Computational Complexity} \label{sec:compute_complex}
The computational cost of the proposed Gibbs sampler is mainly due to the generation of  the TRF $\bfx$ and the label map $\bfz$. In each sampling iteration, the computational complexity for sampling the TRF $\bfx$ using the HMC is of the order $\mathcal{O}((L+1)N\log N)$, where $L$ is the number of Leapfrog iterations and $N$ is the number of image pixels.
The computational complexity for sampling the label map $\bfz$ is of the order $\mathcal{O}(KN)$, where $K$ is the number of label classes. 
Thus, in total, the computation complexity for drawing a cycle of samples in the Gibbs sampler is of the order $\mathcal{O}((K+ (L+1)\log N)N)$. Note that in general $(L+1)\log N \gg K$. Thus, the most time consuming step is for sampling the TRF.}
\section{Experimental results}
This section presents several experiments conducted on simulated and real data using our algorithm. We have also compared our approach with several existing deconvolution algorithms previously applied in US imaging. All the experiments have been conducted using MATLAB R2013a on a computer with Intel(R) Core(TM) i7-4770 CPU @3.40GHz and 8 GB RAM.

\subsection{\rev{Evaluation metrics}}
\rev{Different evaluation metrics were considered for simulated and \textit{in vivo} US images since the TRF ground truth is only available for simulated images. These metrics are presented below.}

\subsubsection{\rev{Simulated US images}}
\paragraph{\rev{Image deconvolution}}
The performance of the TRF estimation is assessed in terms of improvement in SNR (ISNR), normalized root mean square error (NRMSE), peak signal-to-noise ratio (PSNR) and image structural similarity (MSSIM).
The metrics are defined as follows
\begin{align}
&\textrm{ISNR} = 10\log_{10}\frac{\|\bfx-\bfy\|^2}{\|\bfx-\hat{\bfx}\|^2},\\
&\textrm{NRMSE} = \sqrt{\frac{\|\bfx-\hat{\bfx}\|^2}{\|\bfx\|^2}}, \\
&\textrm{PSNR} = 10\log_{10}\frac{\textrm{max}(\bfx,\hat{\bfx})^2}{\textrm{MSE}},\\
&\textrm{MSSIM}(\bfx,\hat{\bfx})  = \frac{1}{W} \sum_{j=1}^{W} \textrm{SSIM}(\bfx_j,\hat{\bfx}_j)
\end{align}
where the vectors $\bfx,\bfy,\hat{\bfx}$ are the ground truth of the TRF, the RF image and the restored TRF, respectively. Note that $W$ is the number of local windows, $\bfx_j$ and $\hat{\bfx}_j$ represent the local reflectivities of $\bfx$ and $\hat{\bfx}$ located in one of these windows and SSIM is the structural similarity measure of each window (defined in \cite{Wang2004}). 
\paragraph{\rev{Image segmentation}}
The performance of the label estimator is assessed using the overall accuracy (OA), defined as the ratio between the number of correctly estimated labels over the total number of labels.

\subsubsection{\rev{\textit{In vivo} US images}}
\rev{Since the ground truth of the TRF and the label map are not available for \textit{in vivo} US data,} the quality of the deconvolution results is evaluated using two other metrics commonly used in US imaging: the resolution gain (RG) \cite{Yu2012} 
and the contrast-to-noise ratio (CNR) \cite{Mahloojifar2010, ACJensen2012}. 
The resolution gain (RG) is the ratio of the normalized autocorrelation (higher than $-3$ dB) of the original RF US image to the normalized autocorrelation (higher than $-3$ dB) of the deconvolved image/restored TRF. 
The definition of the CNR is given by
\begin{equation}
\textrm{CNR} = \frac{|\mu_{1}-\mu_{2}|}{\sqrt{\sigma_{1}^2+\sigma_{2}^2}}
\end{equation}
where $\mu_1$, $\mu_2$, $\sigma_1$ and $\sigma_2$ are the means and standard deviations of pixels located in two regions extracted from the image. 
The two regions are manually chosen so that they belong to different tissue structures. Moreover, as in most US studies, they are at the same depth in order to avoid issues related to wave attenuation.
Note that the higher the values of RG and CNR, the better the deconvolution performance.

\rev{\subsection{Sampler convergence}
\label{sec:converge}
The convergence of the proposed Gibbs sampler can be monitored by determining the so-called burn-in period which refers to the first elements of the Markov chain that are discarded and not used to compute the estimators. The potential scale reduction factor (PSRF) \cite{Gelman1992} requires to define several chains in parallel with different initializations. It is defined by
\begin{equation}
\textrm{PSRF}_v= \frac{M-1}{M} + \frac{C+1}{C M} \frac{B_v}{W_v}
\end{equation}
where $C$ is the number of Markov chains considered, $M$ is the number of iterations after the burn-in period, 
$B_v$ and $W_v$ are the intra-chain and inter-chain variances of the variable $v$, whose definitions are given by
\begin{align}
B_v & = \frac{M}{C-1}\sum_{c=1}^C \left(\bar{v} - \bar{v}_c \right)^2, \\
W_v & = \frac{1}{C}\sum_{c=1}^C \frac{1}{M-1} \sum_{i=1}^M \left(\bar{v}_c - v_c^{(i)}\right)^2
\end{align} 
where $\bar{v} = \frac{1}{C}\sum_{c=1}^C \bar{v}_c$, $\bar{v}_c=\frac{1}{M}\sum_{i=1}^M  v_c^{(i)}$ and $v_c^{(i)}$ is the $i$th sample of the variable $v$ in the $c$th chain. Values of the PSRF below $1.2$ indicate a good convergence of the sampler as suggested in \cite{Gelman1992}. In this work, we checked that the PSRFs of all the variables of interest were below $1.2$.
}

\subsection{Synthetic data}
\subsubsection{\rev{Deconvolution}}
\rev{We first study the deconvolution performance on synthetic data with controlled ground truth, which allows the quality of the different estimators to be appreciated.} 
\begin{figure}[htpt]
\centering
\subfigure[$\xi=2$,$\gamma=2$]{\includegraphics[width=0.3\linewidth]{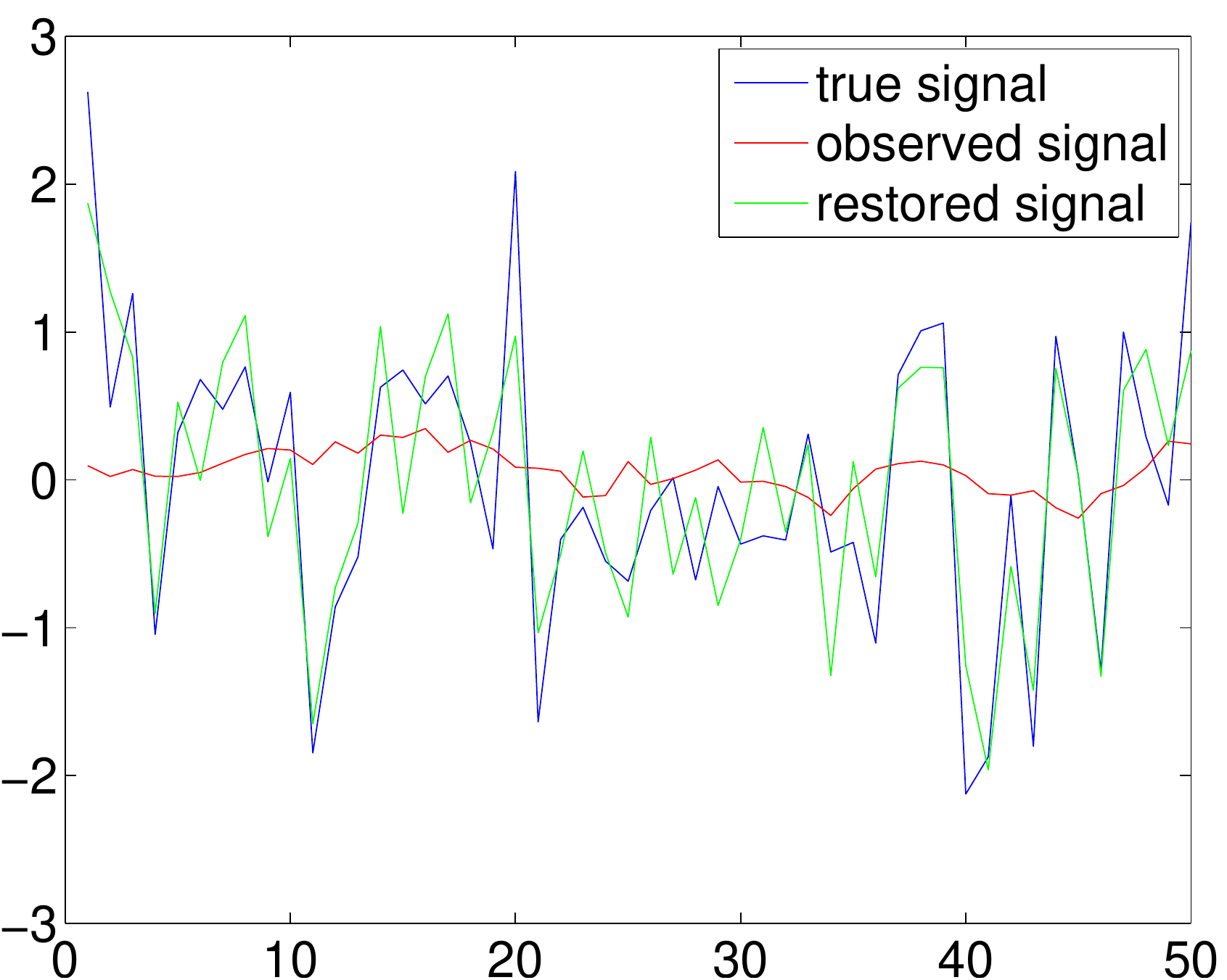}}
\hfil
\subfigure[$\xi=1.5$,$\gamma=1.2$]{\includegraphics[width=0.3\linewidth]{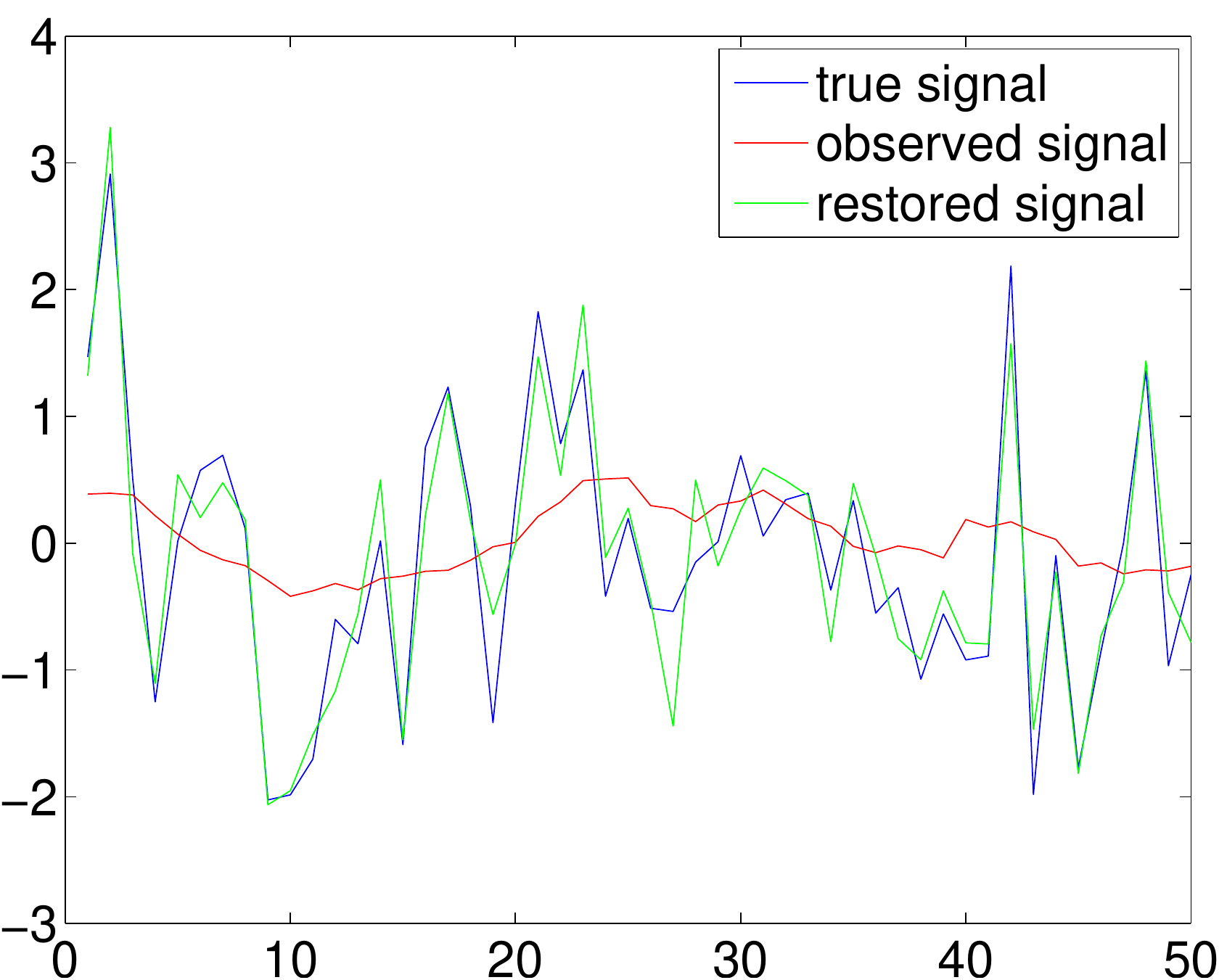}}
\hfil
\subfigure[$\xi=0.6$,$\gamma=0.4$]{\includegraphics[width=0.3\linewidth]{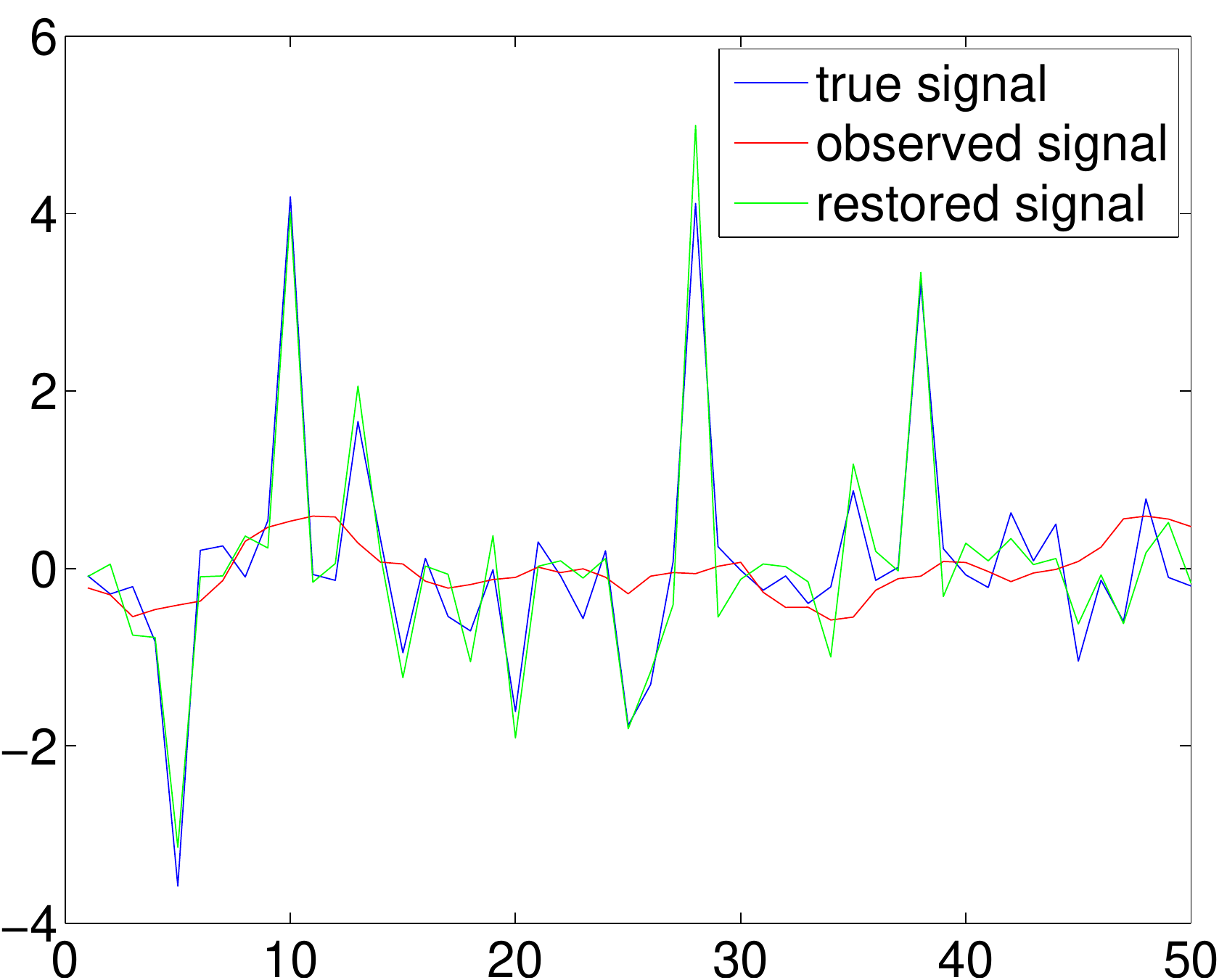}}
\caption{Deconvolution results for one column of the synthetic image (the red curves are the observed lines, the blue curves are the ground truth and the green curves are the restored signals using the proposed method).}
\label{fig:rflines}
\end{figure}
Precisely, three groups of 2D synthetic images with the same image size $N=50\times 50$ are generated assuming that the image pixels are independent and identically distributed (\textit{i.i.d.}) according to GGDs with different shape and scale parameters, as reported in Table \ref{tab1:estimate_paras}. Each image has been corrupted by a $5\times 5$ Gaussian blurring kernel with variance $\sigma_b=3$ and an AWGN. The level of AWGN is characterized by the blurred signal-to-noise ratio (BSNR) expressed in decibels as follows
\begin{equation}
\textrm{BSNR}=10 \log_{10} \left(\frac{\|\bfH\bfx - E(\bfH\bfx)\|_2^2}{N \sigma_n^2}\right)
\end{equation}  
where $E(\cdot)$ is the \textcolor{black}{empirical average} and $N$ is the total number of image pixels. The BSNR was set to $40$ dB for the synthetic data. 
Regarding the MCMC algorithm, $50$ chains of $6000$ iterations including a burn-in period of $2000$ iterations were run for each simulation scenario. In each Monte Carlo chain, the stepsize was initialized to $\epsilon=10^{-5}$ and the number of leapfrog steps was uniformly sampled in the interval $[50,70]$.

\begin{figure}[htpt]
\centering
\subfigure[$\sigma_{n}^2$]{\includegraphics[width=0.3\linewidth]{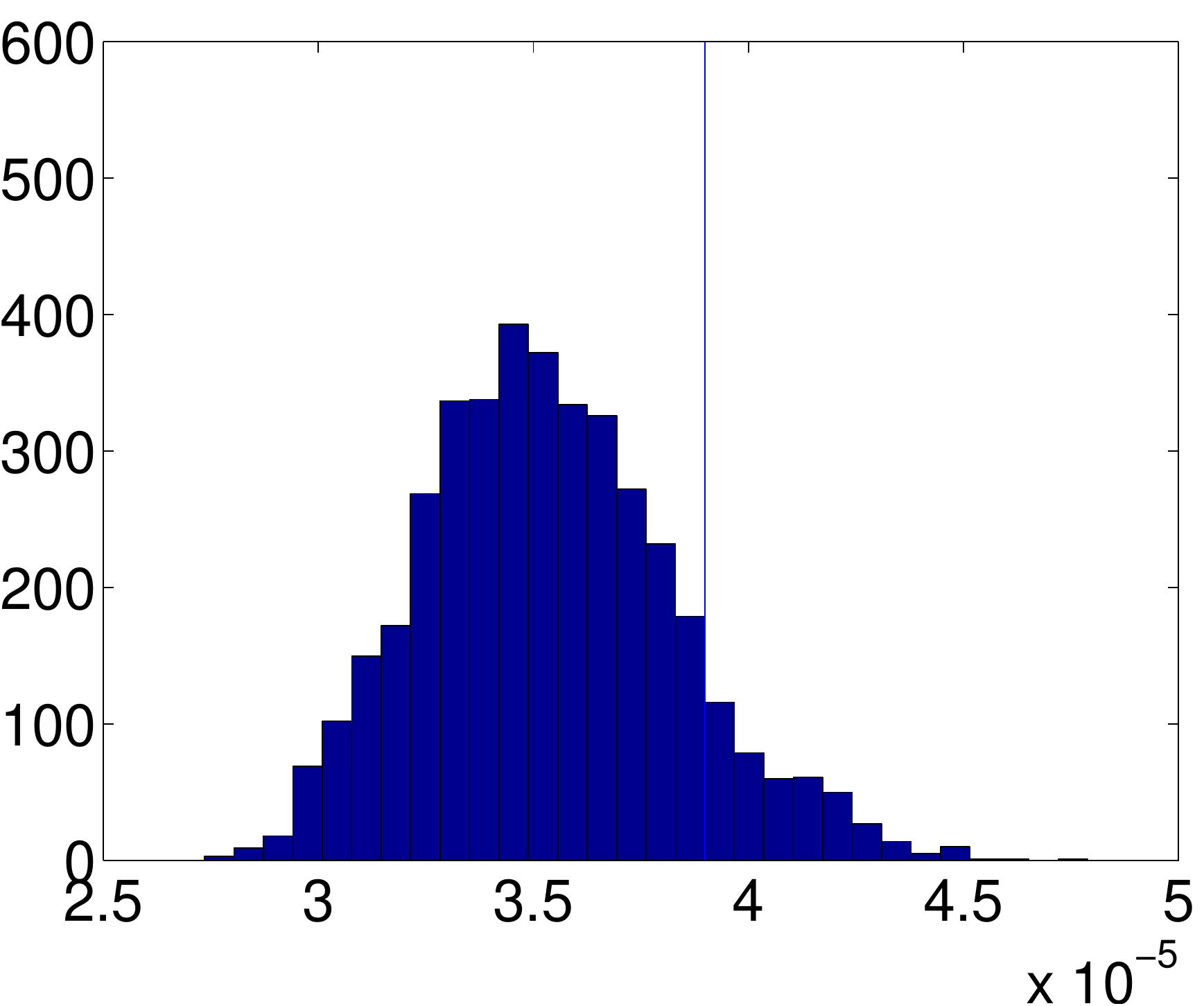}}
\subfigure[$\sigma_{n}^2$]{\includegraphics[width=0.3\linewidth]{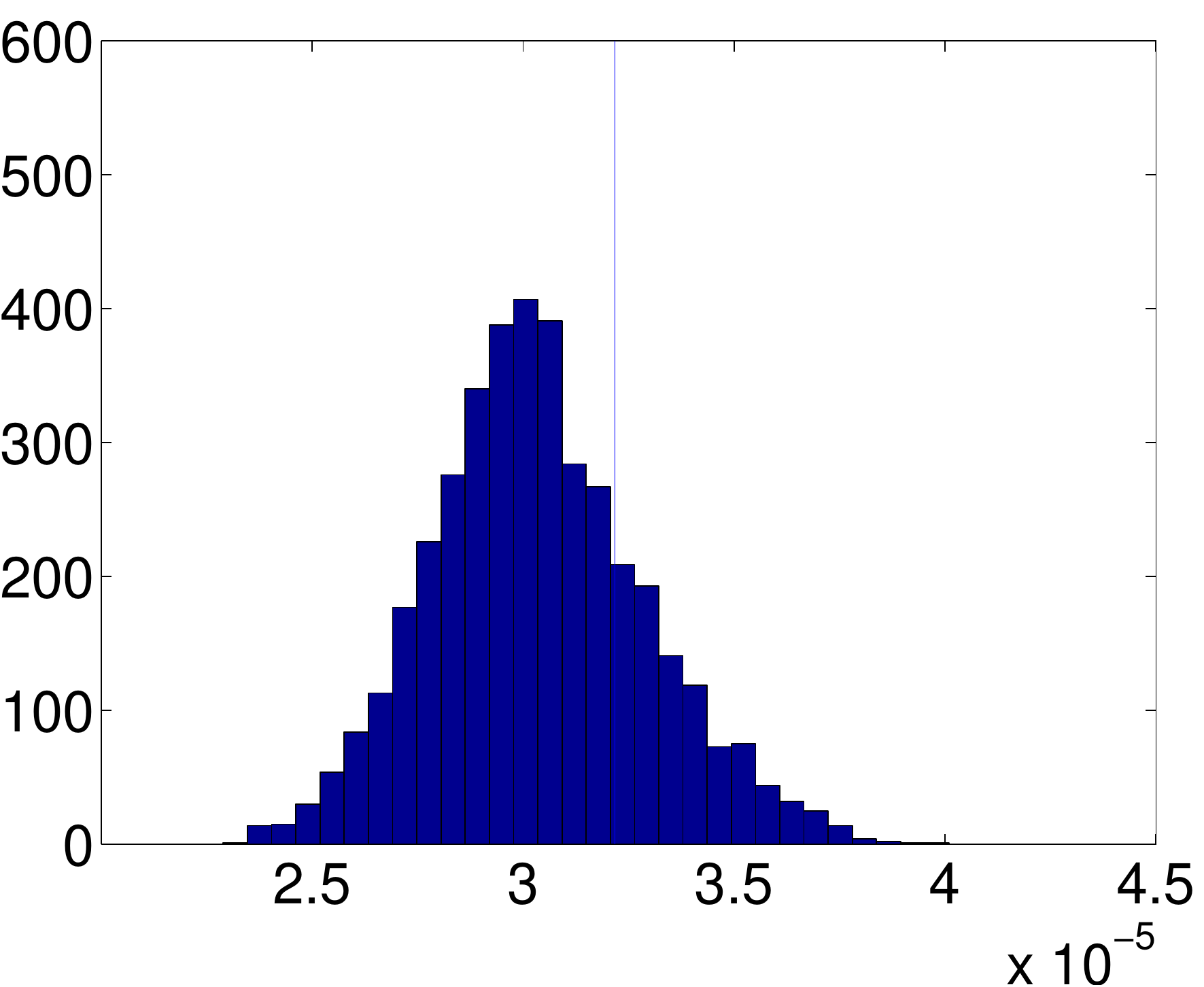}}
\subfigure[$\sigma_{n}^2$]{\includegraphics[width=0.3\linewidth]{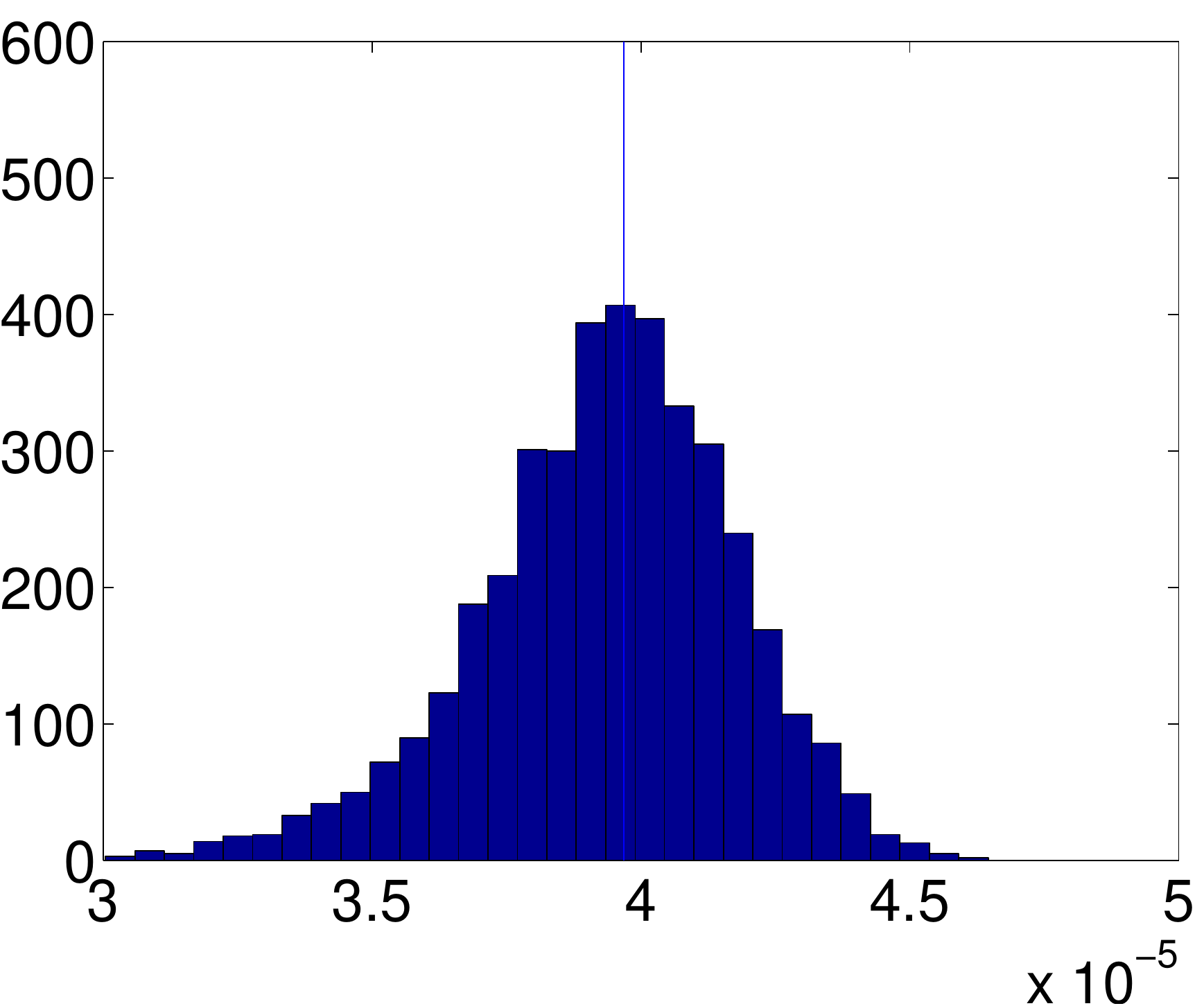}}
\subfigure[$\xi$]{\includegraphics[width=0.3\linewidth]{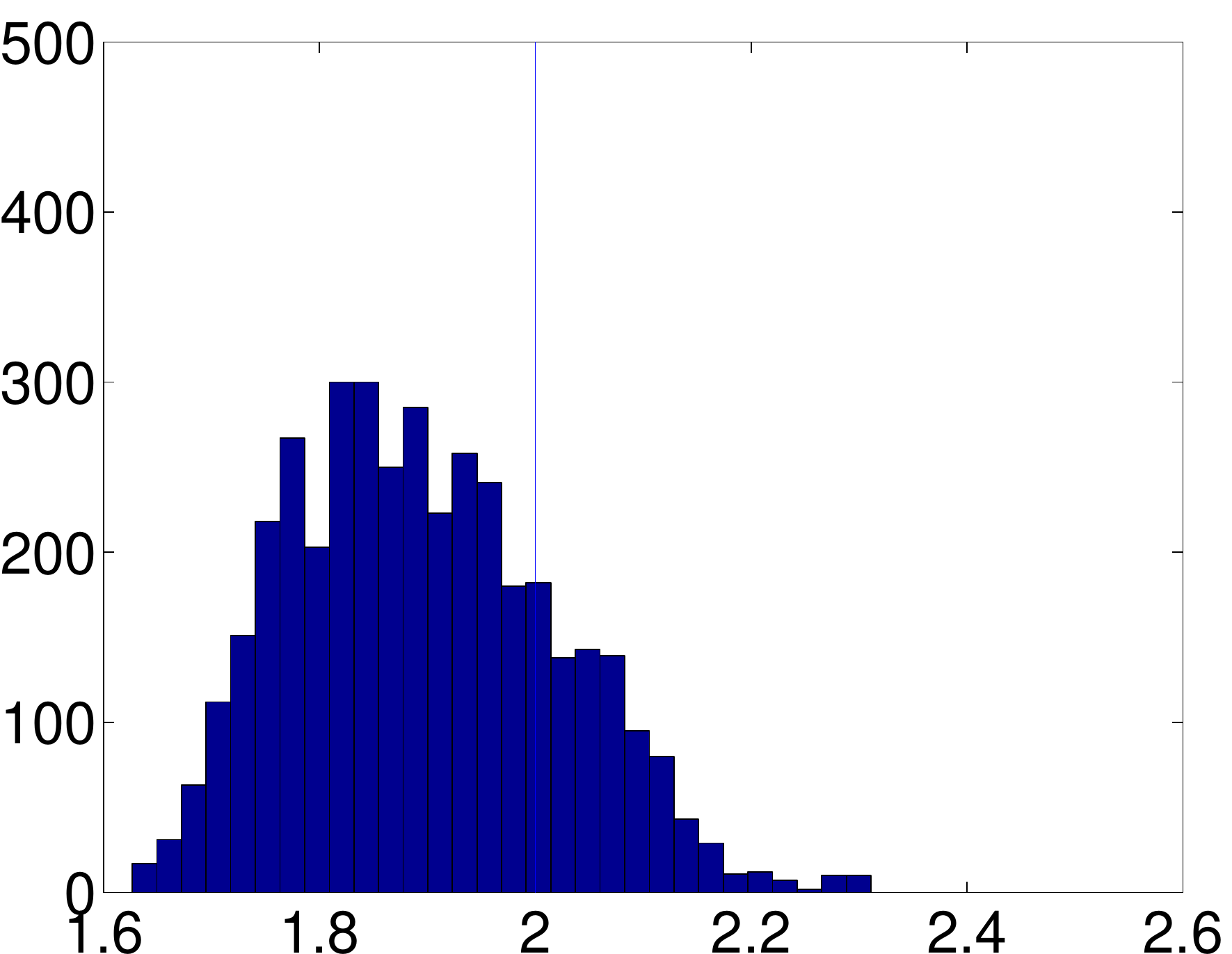}}
\subfigure[$\xi$]{\includegraphics[width=0.3\linewidth]{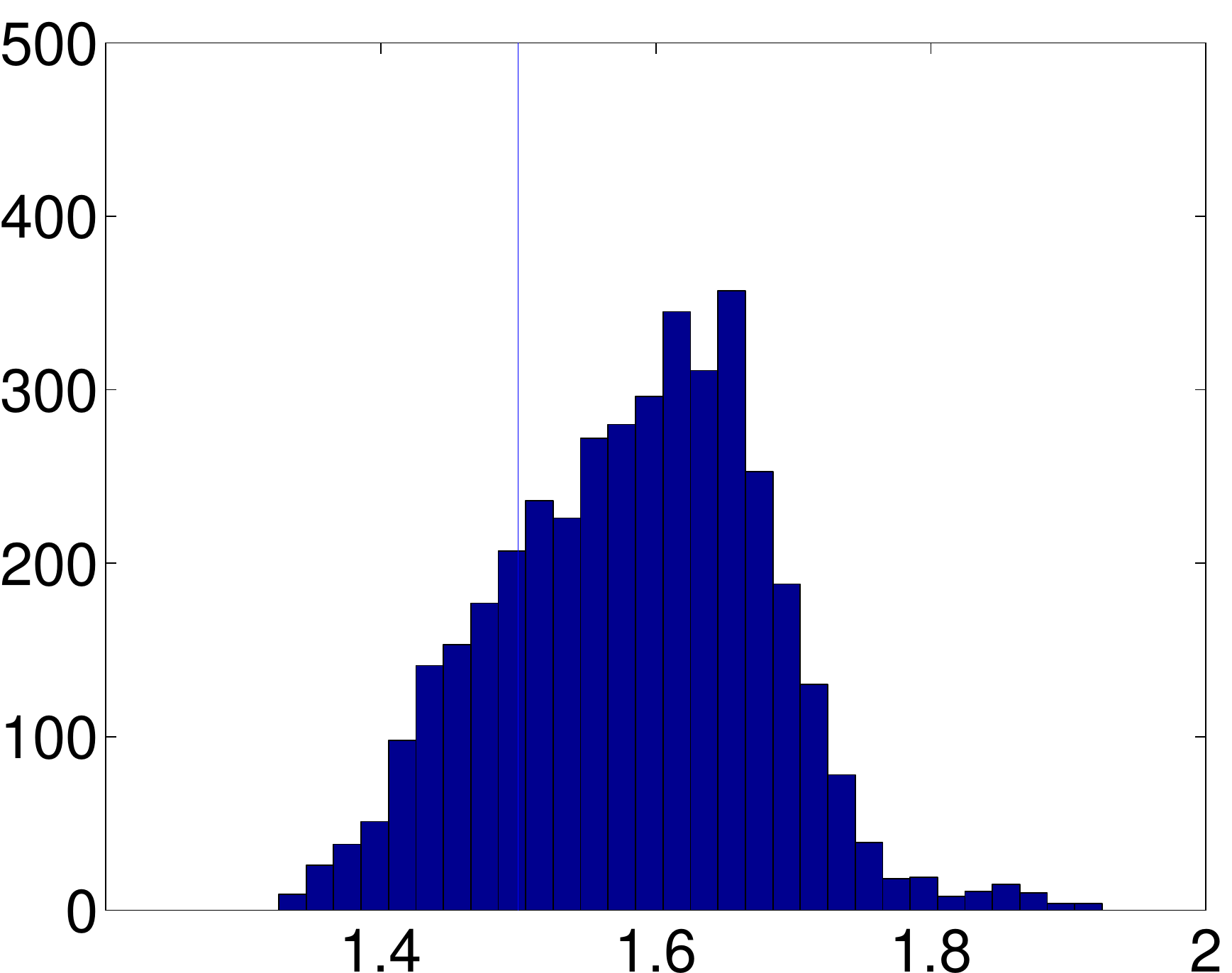}}
\subfigure[$\xi$]{\includegraphics[width=0.3\linewidth]{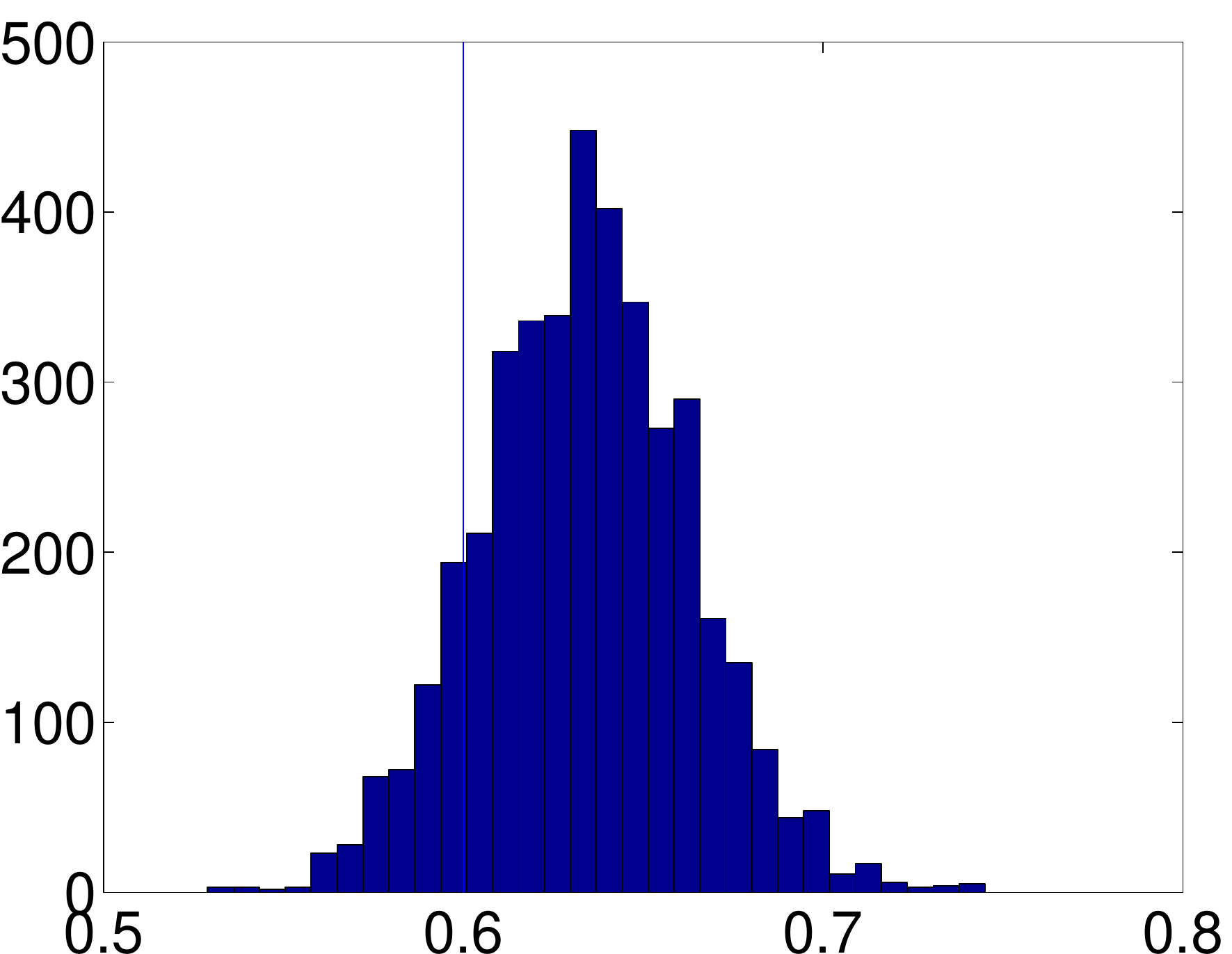}}
\subfigure[$\gamma$]{\includegraphics[width=0.3\linewidth]{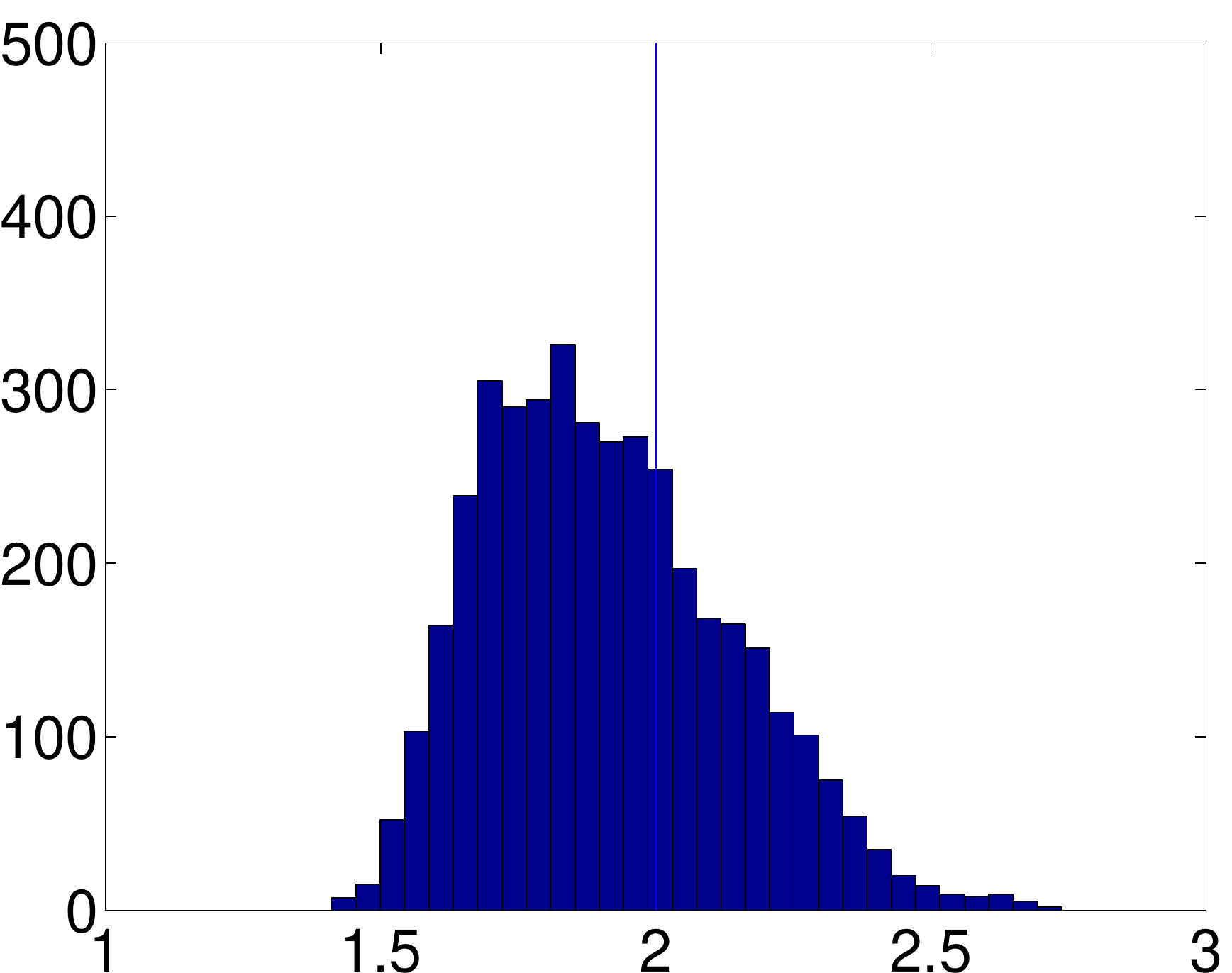}}
\subfigure[$\gamma$]{\includegraphics[width=0.3\linewidth]{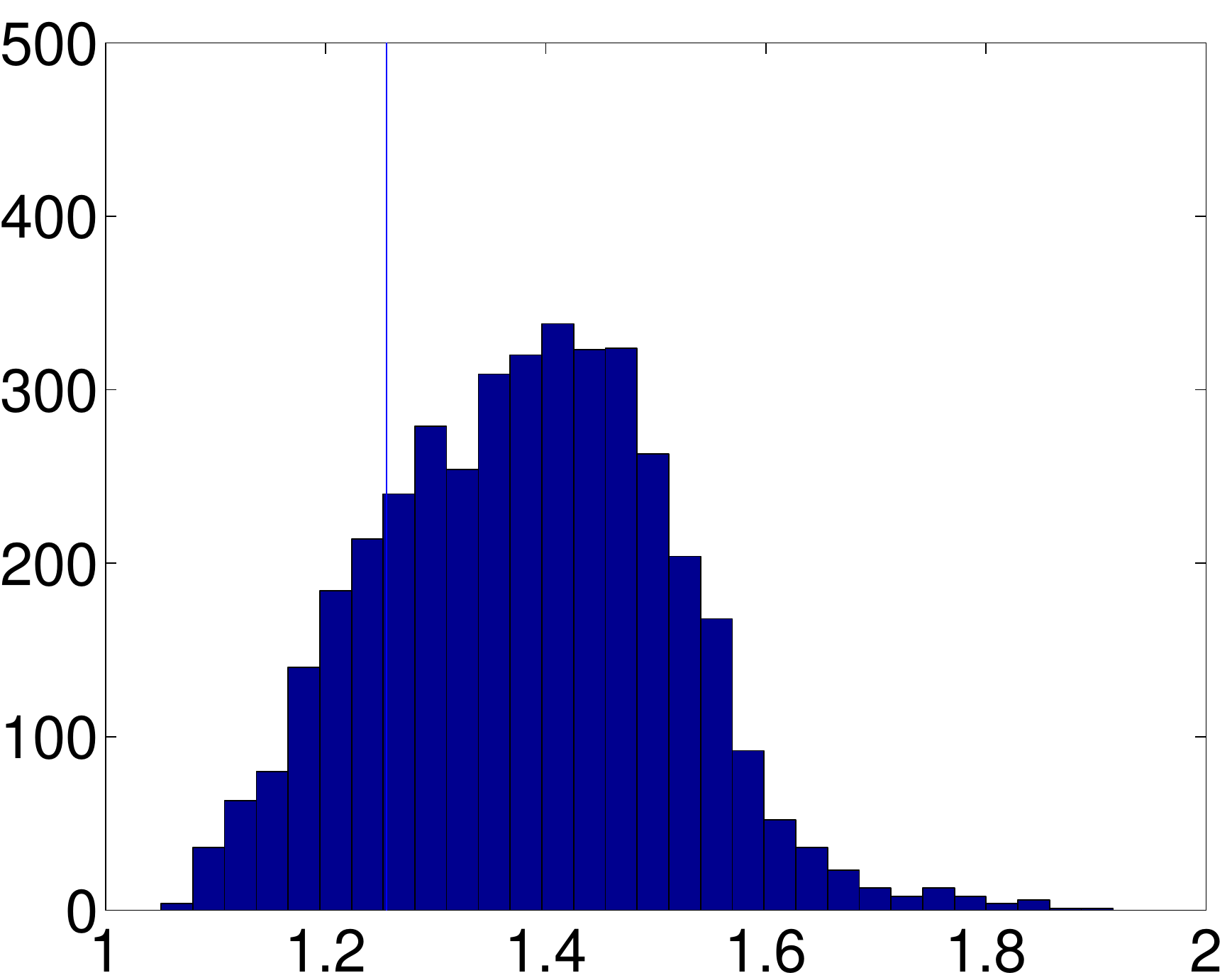}}
\subfigure[$\gamma$]{\includegraphics[width=0.3\linewidth]{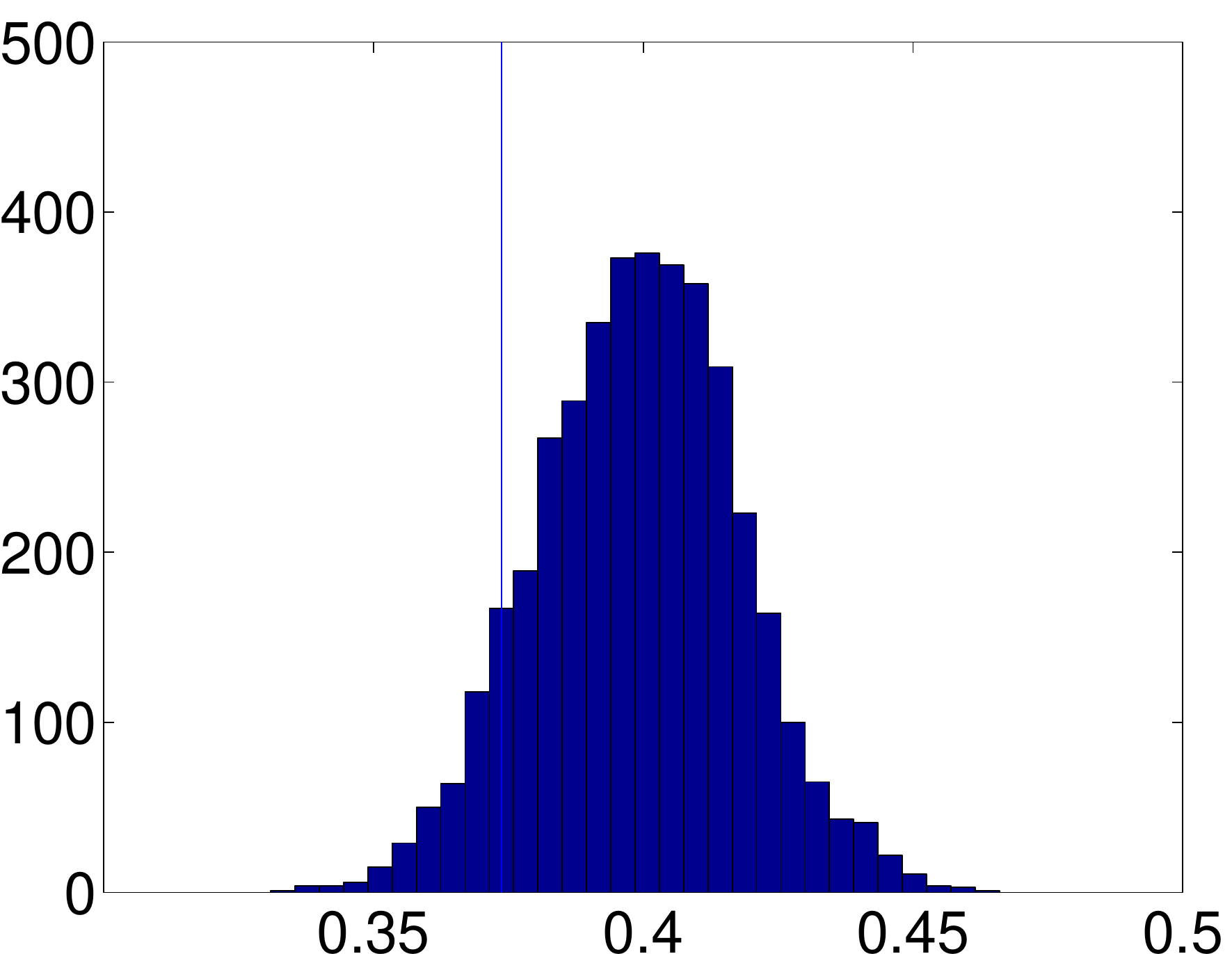}}
\caption{\rev{Estimated marginal posterior distributions (histograms) of the noise variance $\sigma_n^2$ (a)-(c), the hyperparameters $\xi$ (d)-(f) and $\gamma$ (g)-(j). 
The vertical lines represent the ground truths of the corresponding parameters. Each column corresponds to a given image.}}
\label{fig:hists}
\end{figure}

\begin{table}[htpt]
\begin{center}
\caption{\rev{Parameter Estimations for the synthetic data}}
\label{tab1:estimate_paras}
\begin{tabular}{|c|c|c|c|c|}
\hline
Group & Parameters& True values & MMSE  & Standard \\
      &           &             &       & deviation \\
\hline 
 \multirow{3}{*}{Group 1}&  $\sigma_{n}^2$ ($\times 10^{-5}$)& 3.72   & 3.65 & 0.35 \\
                         &  $\xi$          & 2          & 1.98     & 0.04  \\
                         &  $\gamma$       & 2          & 2.00     &  0.05     \\
\hline
 \multirow{3}{*}{Group 2}&   $\sigma_{n}^2$ ($\times 10^{-5}$) & 3.22 & 3.63 &  0.61\\
                         &   $\xi$          & 1.50     &  1.41    &  0.09     \\
                         &   $\gamma$       & 1.26     &   1.16   &  0.09    \\
\hline
 \multirow{3}{*}{Group 3}&    $\sigma_{n}^2$ ($\times 10^{-5}$)& 3.13 & 4.15& 0.60 \\
                         &    $\xi$          & 0.60     &   0.59  &   0.03         \\
                         &    $\gamma$       & 0.37     &   0.37 &    0.02   \\
\hline
\end{tabular}
\end{center}
\end{table}
The typical deconvolution performance for one column of each of the three observed images is depicted in Fig. \ref{fig:rflines}. These results show a good performance of the proposed image deconvolution algorithm. 
Fig. \ref{fig:hists} shows the histograms of the generated samples from one single Markov chain for the noise variance, the GGD parameters and the hyperparameters of three synthetic images. These histograms are clearly in good agreement with the true values of the parameters indicated by the vertical lines. More quantitative results of the parameter estimation are reported in Table \ref{tab1:estimate_paras}. 

\subsubsection{\rev{Segmentation}}
\begin{figure}[htpt]
\centering
\subfigure{\includegraphics[width=0.3\linewidth]{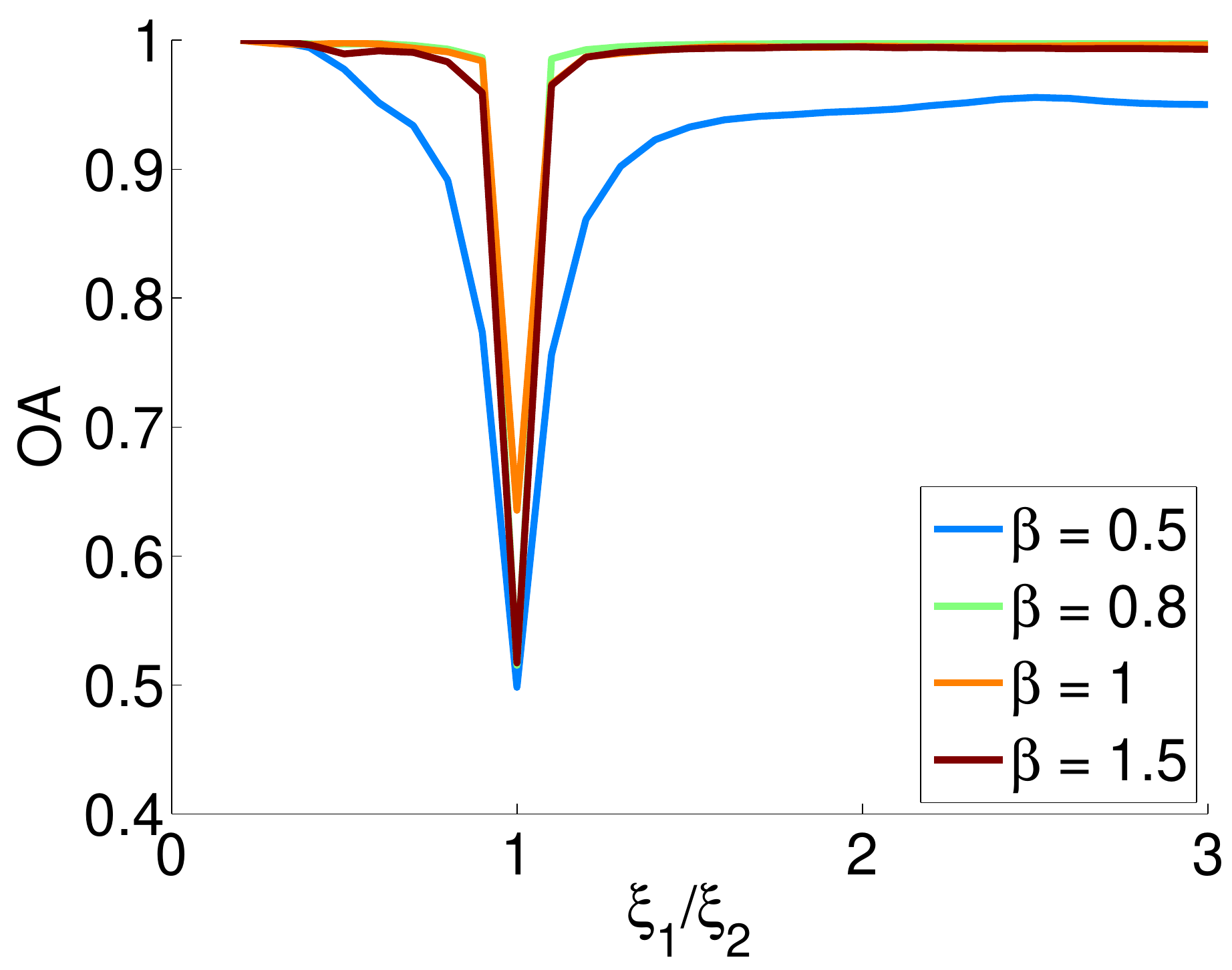}}
\subfigure{\includegraphics[width=0.3\linewidth]{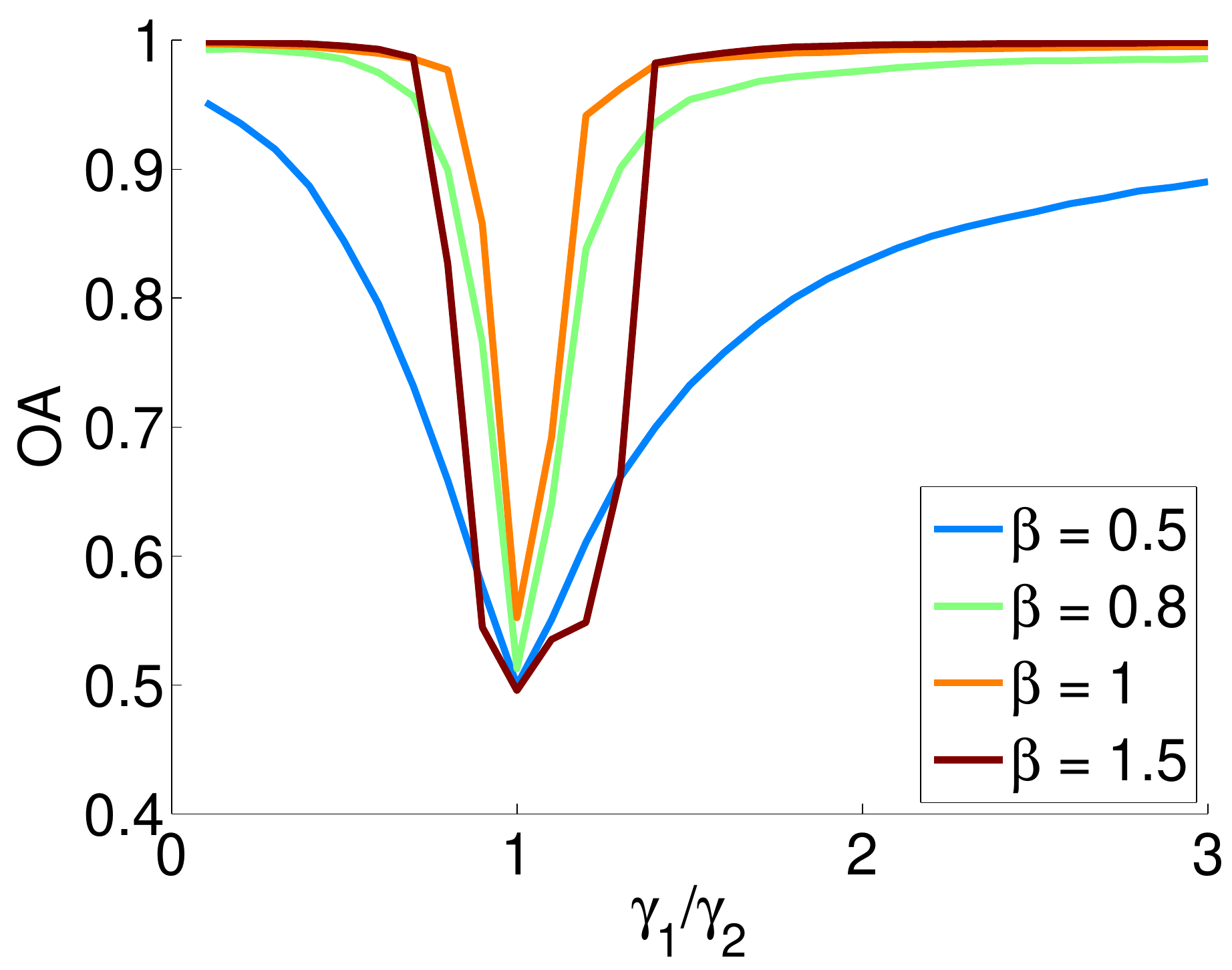}}
\caption{\rev{OA versus the ratios of the GGD parameters (left: scale parameters $\gamma_1=\gamma_2=20$, right: shape parameters $\xi_1=\xi_2 = 1$).} }
\label{fig:OAvsGGDpara}
\end{figure}

\rev{This section evaluates the performance of our method for the segmentation of two regions of the same size ($128\times 64$) using the overall accuracy (OA). 
Given that pixels in both regions have a zero-mean GGD, the difference between the two regions is controlled by the ratios of the shape or scale parameters in the two regions. The values of OAs obtained for different ratios of GGD parameters are displayed in Fig. \ref{fig:OAvsGGDpara}.  
Comparing the two graphs in Fig. \ref{fig:OAvsGGDpara}, the variations of OA are clearly sharper for the left figure, showing that the segmentation accuracy is more sensitive to the shape parameters.} 

\subsection{\rev{Joint Deconvolution and Segmentation}}
\subsubsection{\rev{Comparison with existing methods}}
\paragraph{Simulated US images}
The proposed joint deconvolution and segmentation algorithm (denoted as ``$\rm{Joint_{MCMC}}$'') was compared to the technique proposed in \cite{MartinoAlessandrini2011_AI} (that performs US deconvolution with GGD priors using the EM algorithm, denoted here by ``$\rm{Deconv_{EM}}$'') on simulated data. Since ``$\rm{Deconv_{EM}}$'' was proposed for statistical homogeneous regions, we assumed that the labels associated with the statistically homogeneous regions were known for ``$\rm{Deconv_{EM}}$''. In order to test the robustness of our method to label estimation errors, we also implemented the proposed algorithm using the true labels (denoted as ``$\rm{Deconv_{MCMC}}$''). In this case, similar to ``$\rm{Deconv_{EM}}$'', only the deconvolution process was performed, without label estimation. \rev{Finally, we compared our results with the $\ell_2$ and $\ell_1$ norm constrained optimization solutions.} For the $\ell_2$-norm optimization problem, a numerical solution is given by\begin{equation}
\hat{\bfx} = (\bfH^T\bfH + \lambda \bfI)^{-1}\bfH^T\bfy
\label{AS_l2}
\end{equation} 
where $\lambda$ is the regularization parameter. Concerning the $\ell_1$ norm optimization problem, numerous dedicated algorithms, e.g., ISTA \cite{BeckTeboulle2009}, FISTA \cite{BeckTeboulle2009}, TwIST \cite{TwIST_Bioucas-Dias07} or GEM \cite{GEM_Bioucas-Dias2006} are available in the literature. \rev{The conjugate gradient (CG) method was considered in this work. Note that the regularization parameters were fixed manually by cross validation for the $\ell_2$ and $\ell_1$ norm constraint optimization problems.}
\paragraph{\rev{\textit{In vivo} US images}}
\rev{Due to the fact that the ground truth for the label map is not available for \textit{in vivo} US data, we were not able to test the methods ``$\rm{Deconv_{EM}}$'' and ``$\rm{Deconv_{MCMC}}$'' for these images. Instead,} we considered Gaussian and Laplacian priors that have been extensively used in the US image deconvolution literature \cite{Michailovich2007}, \cite{Jirik2008}, \cite{Yu2012}. 
\rev{The analytical solution for the $\ell_2$-norm optimization problem is given by \eqref{AS_l2}.} 
\rev{The GPSR (gradient projection for sparse reconstruction) \cite{Figueredo2007GPSR} algorithm is implemented for the $\ell_1$ norm constrained optimization problem for the real data, where the regularization parameter is chosen as $0.1 \|\bfH^T\bfy \|_{\infty}$, as suggested in \cite{Figueredo2007GPSR}.}

\subsubsection{Joint deconvolution and segmentation for simulated US images}
\begin{figure}[htpt]
\begin{center}
\subfigure[histograms of $\bsxi_{\textrm{in}}$]{\includegraphics[width=0.3\linewidth]{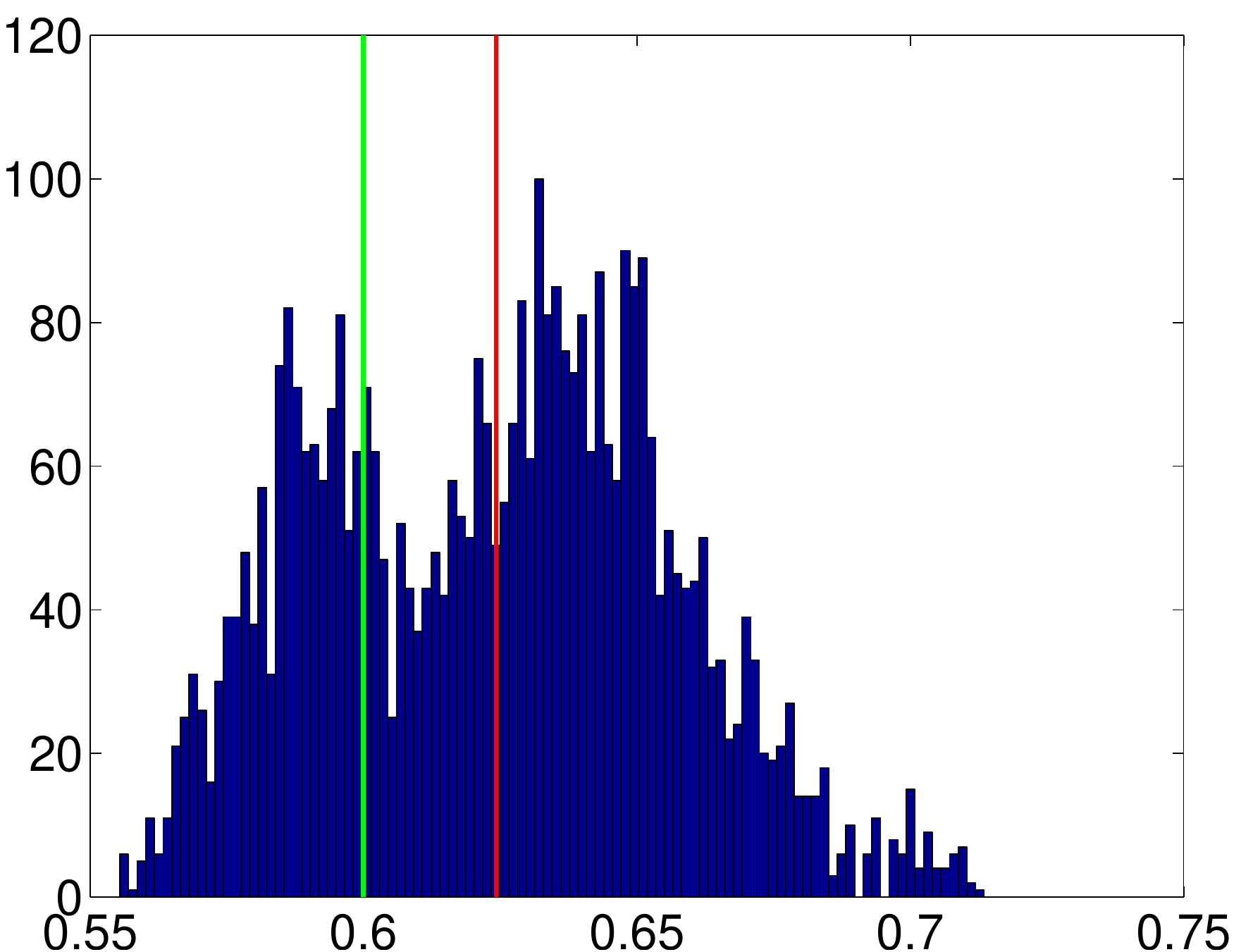}}
\subfigure[histograms of $\bsxi_{\textrm{out}}$]{\includegraphics[width=0.3\linewidth]{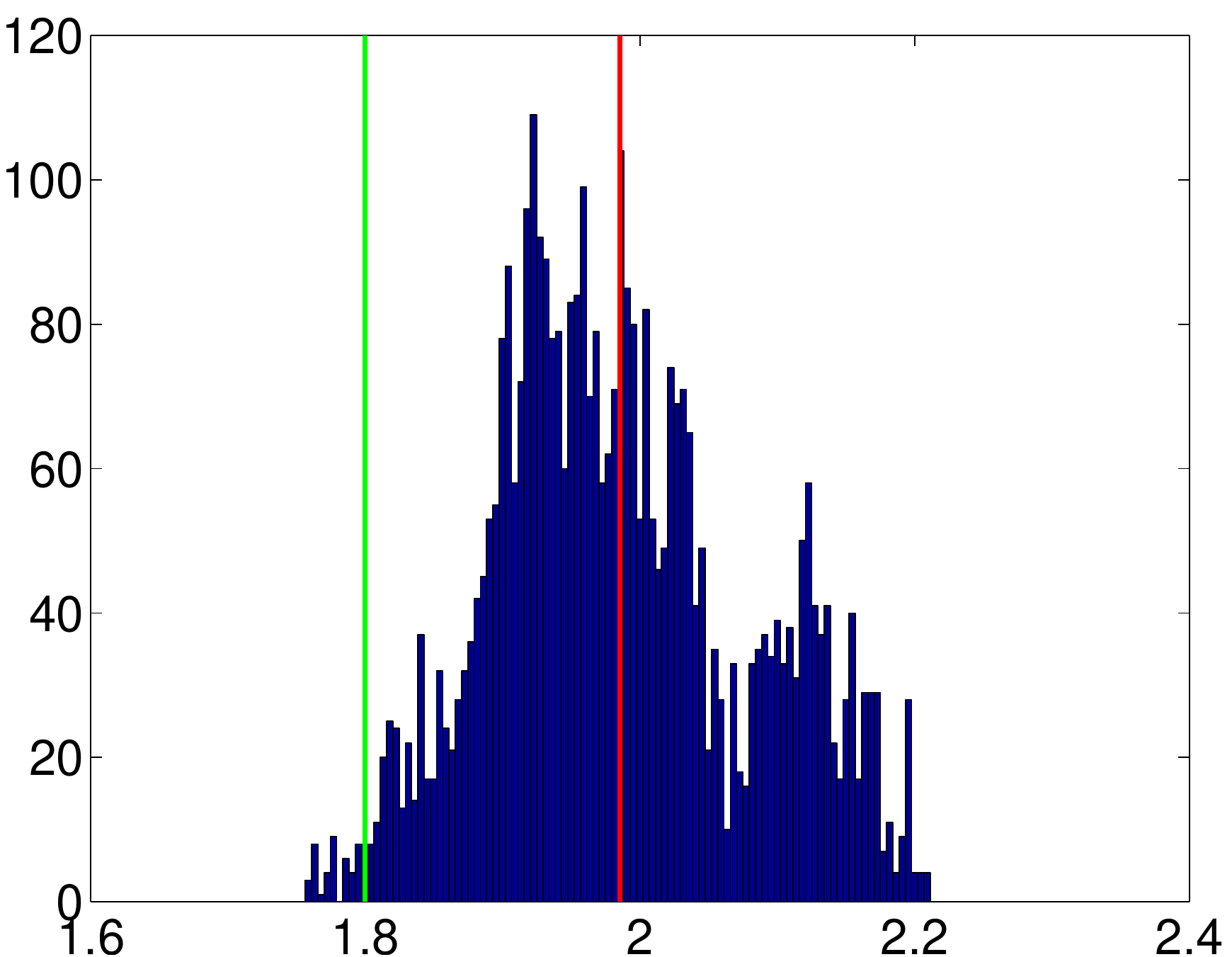}}\\
\subfigure[histograms of $\bsgamma_{\textrm{in}}$]{\includegraphics[width=0.3\linewidth]{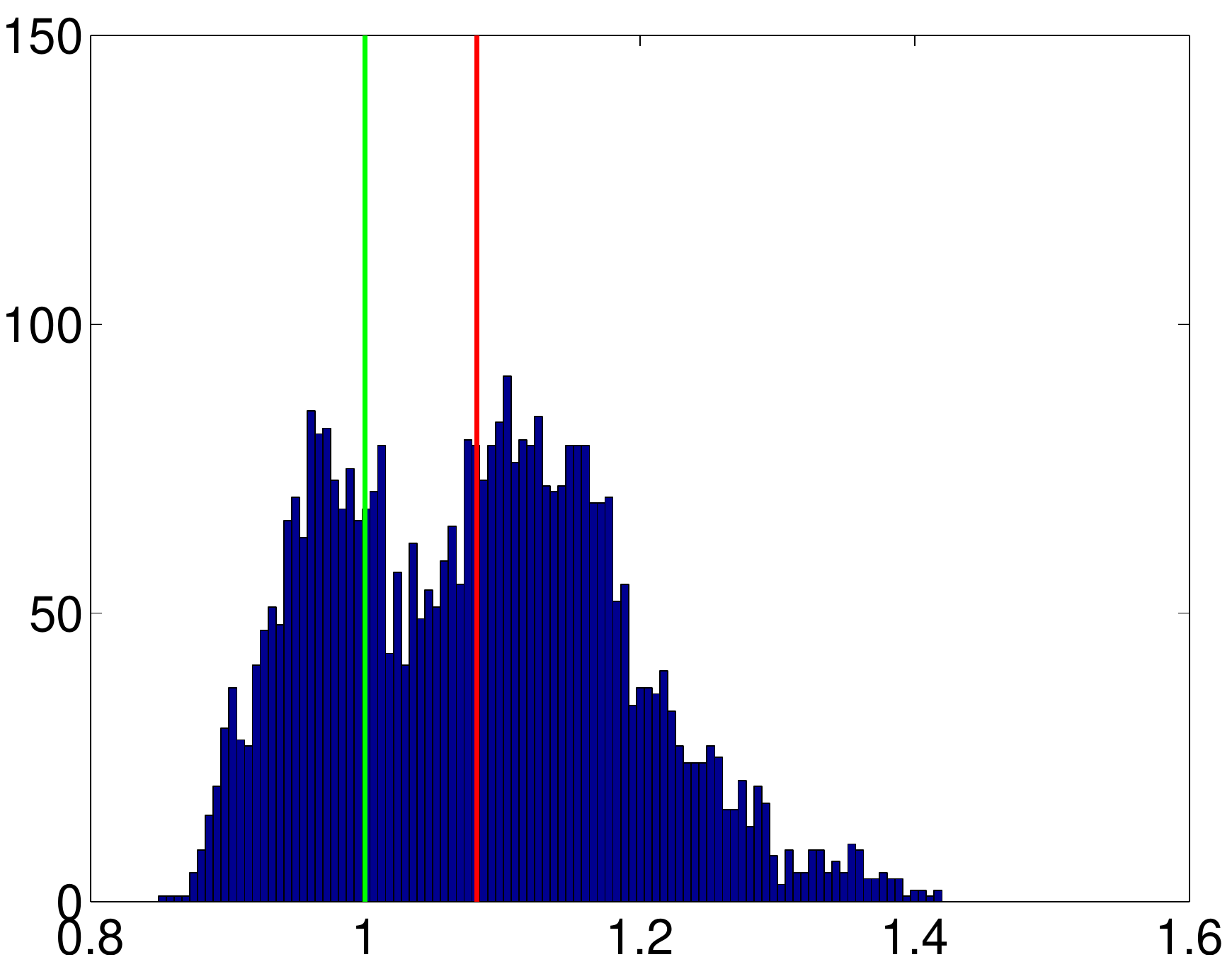} \label{circle_xi_sample}}         
\subfigure[histograms of $\bsgamma_{\textrm{out}}$]{\includegraphics[width=0.3\linewidth]{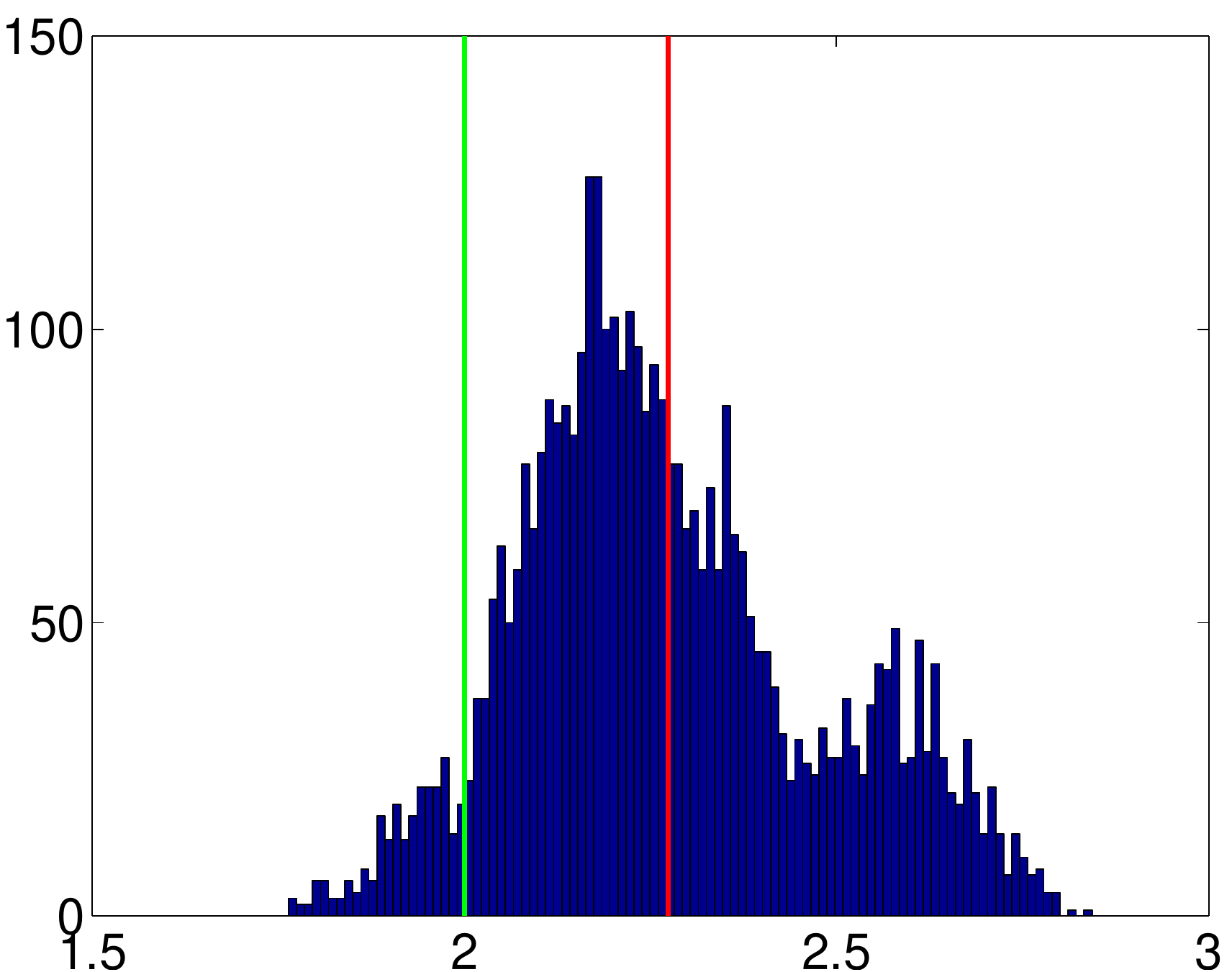} \label{circle_scale_sample}}
\caption{Group 1:  
(a) and (b) are the histograms of shape parameters $\bsxi$ for the pixels inside and outside the inclusion;
(c) and (d) are the histograms of scale parameters $\bsgamma$ for the pixels inside and outside the inclusion; 
The red and green vertical lines are the MMSE estimates and the true values of the parameters $\bsxi$, $\bsgamma$, respectively. 
}
\label{fig:1circle_samples}
\end{center}
\end{figure}

\begin{figure*}
\centering
\subfigure[PSF]{\includegraphics[width=0.18\linewidth]{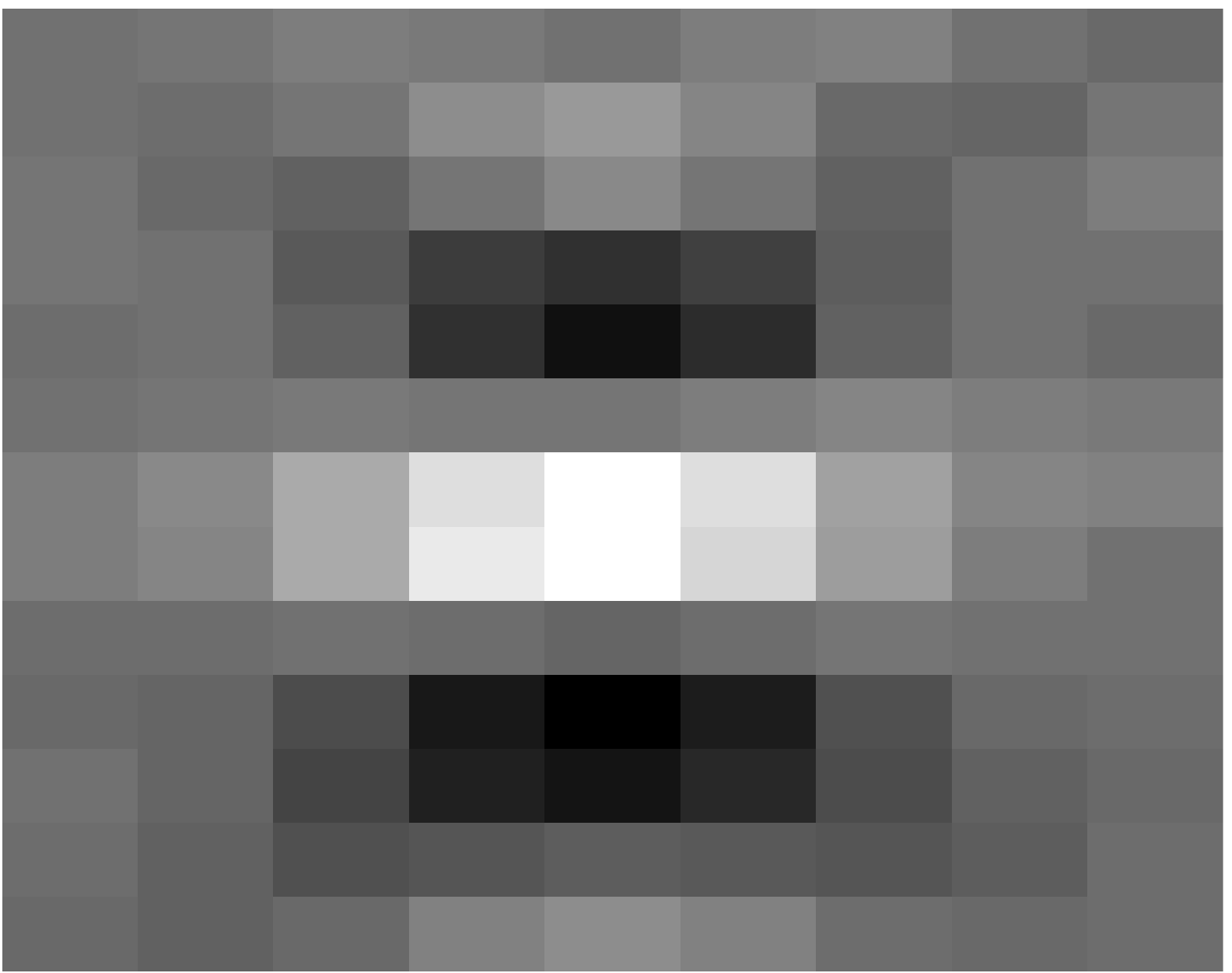} \label{fig:psf0}}
\subfigure[\rev{TRF}]{\includegraphics[width=0.18\linewidth]{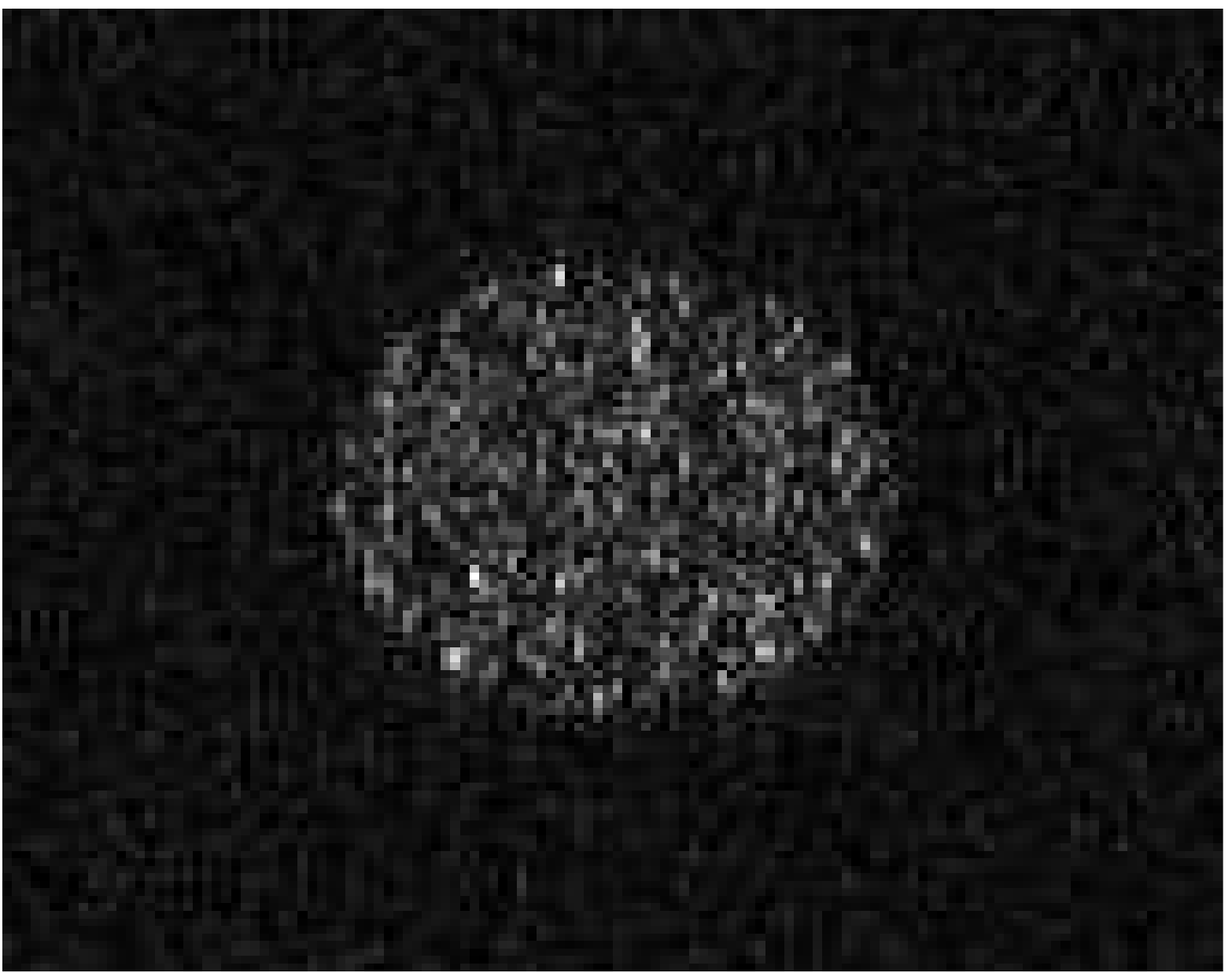} \label{fig:refl0}}
\subfigure[Label]{\includegraphics[width=0.18\linewidth]{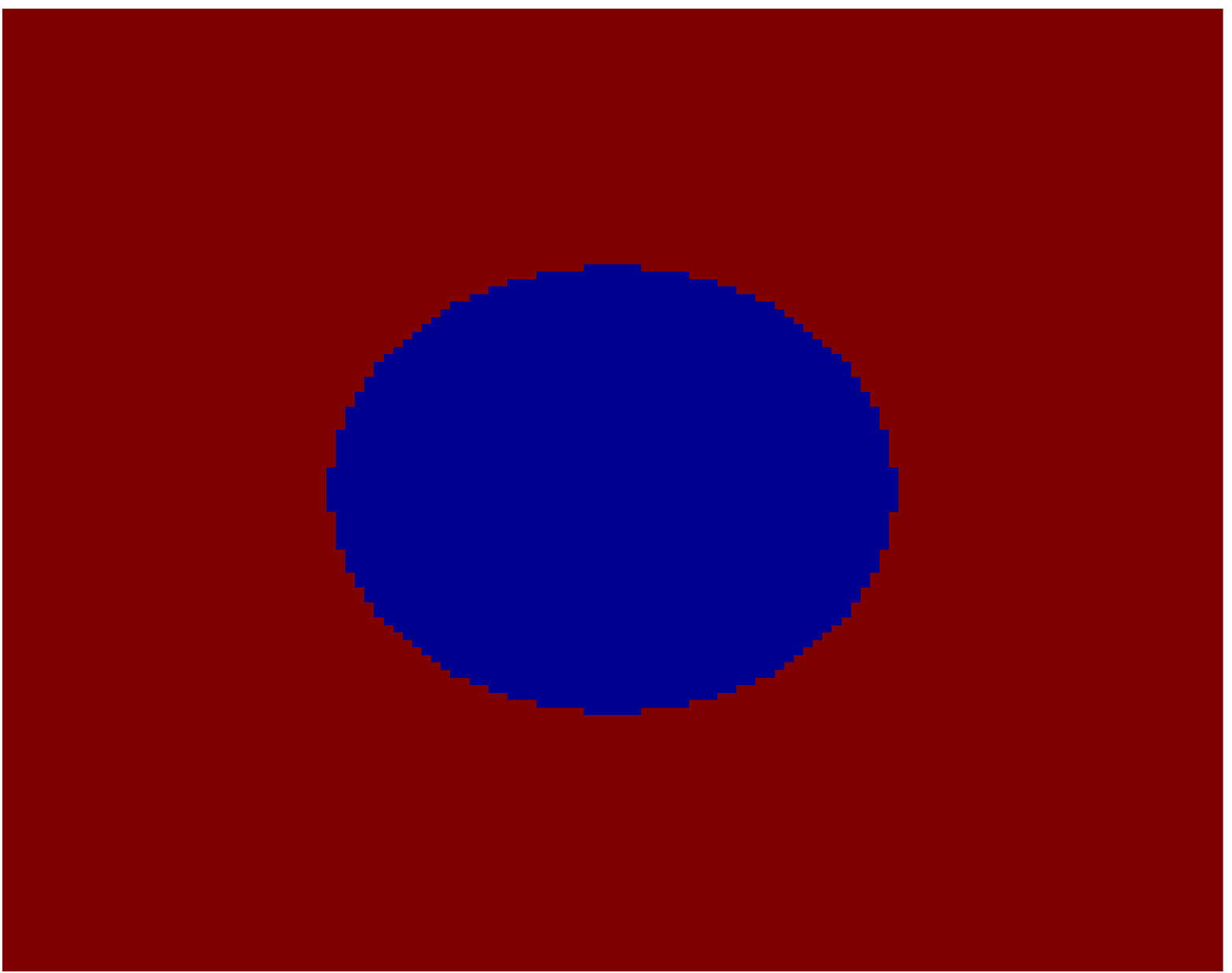} \label{fig:mask0}}
\subfigure[B-mode]{\includegraphics[width=0.18\linewidth]{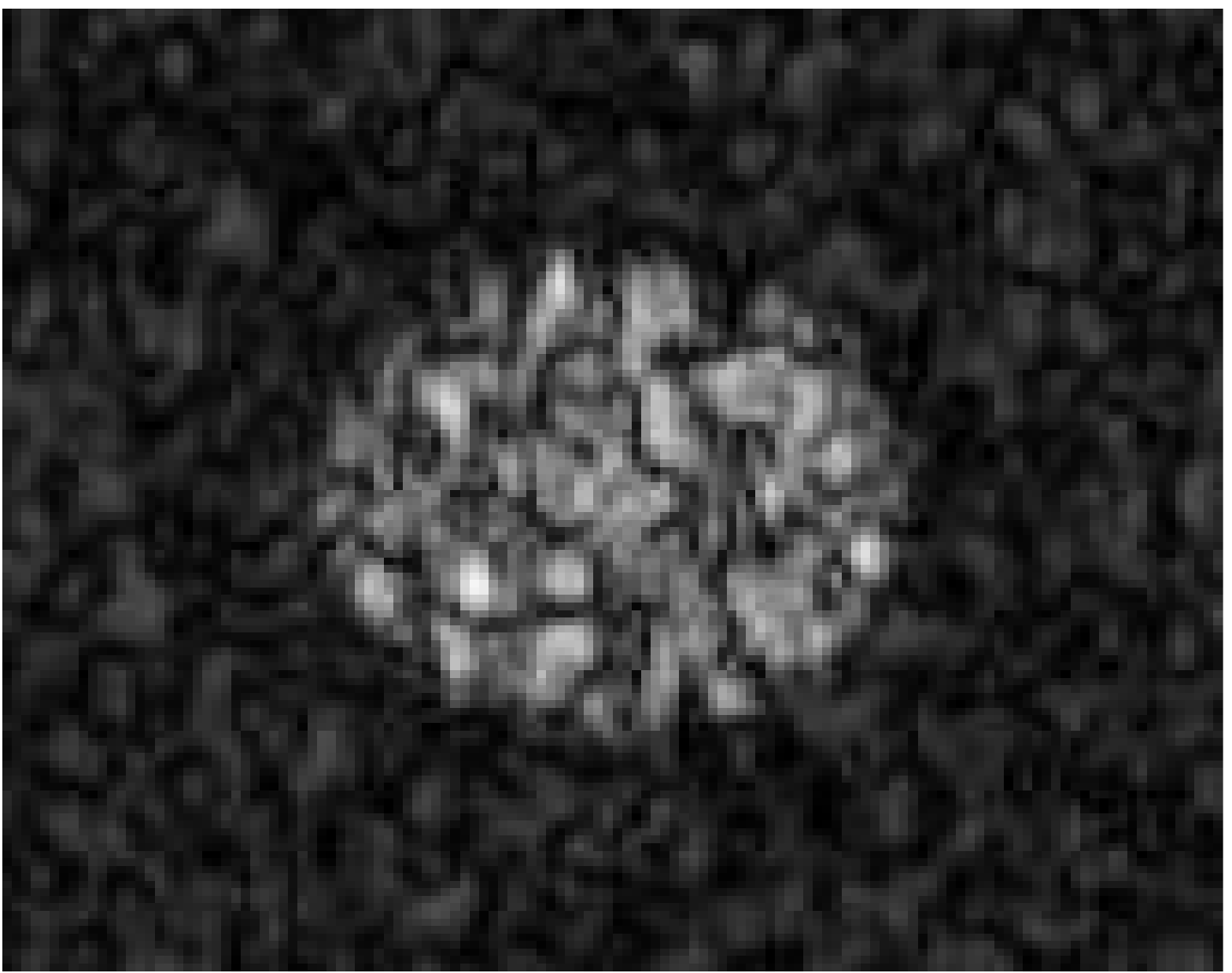} \label{fig:bmode_y0}}
\subfigure[$\ell_2$]{\includegraphics[width=0.18\linewidth]{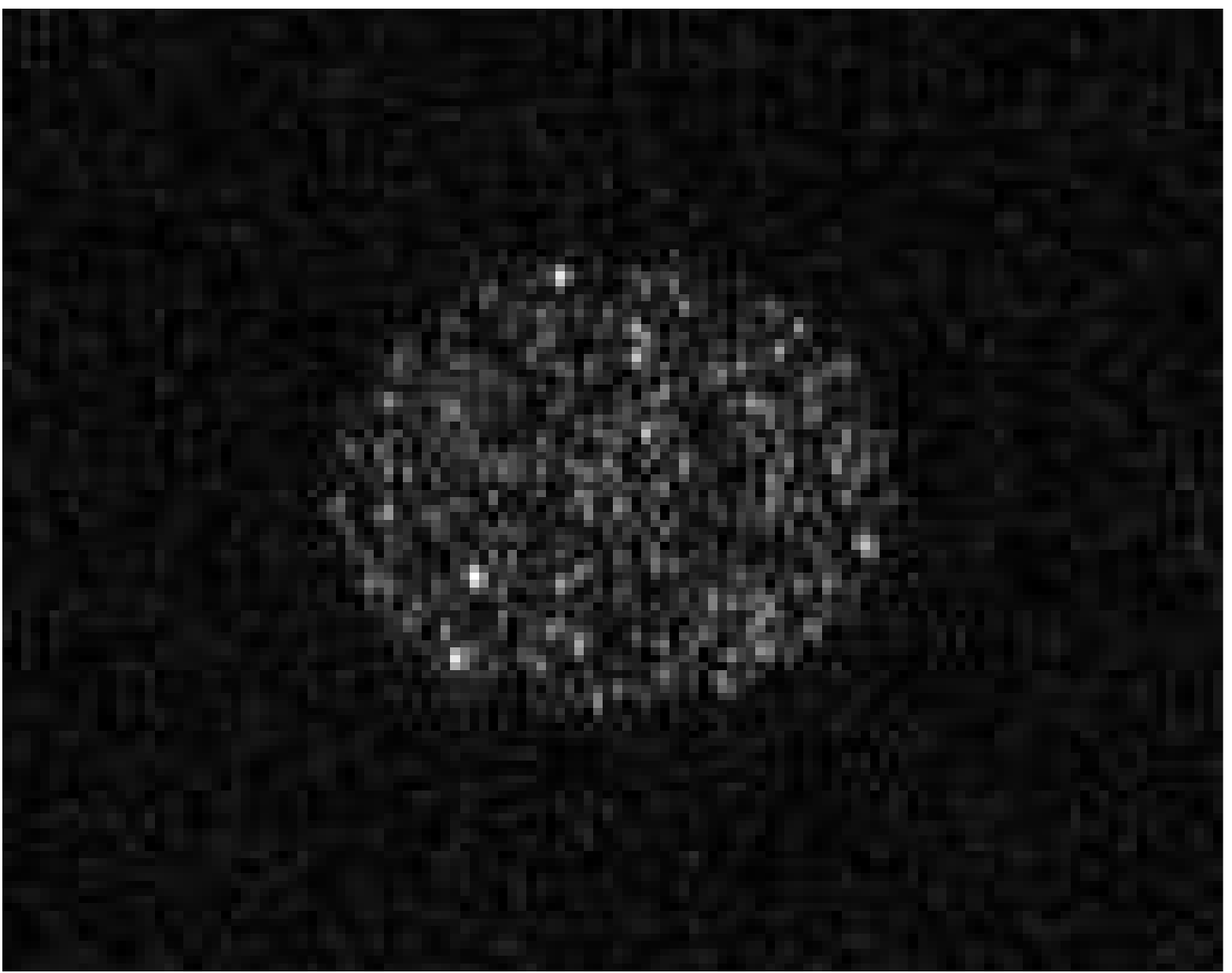} \label{fig:bmode_Deconvl20}}\\
\subfigure[$\ell_1$]{\includegraphics[width=0.18\linewidth]{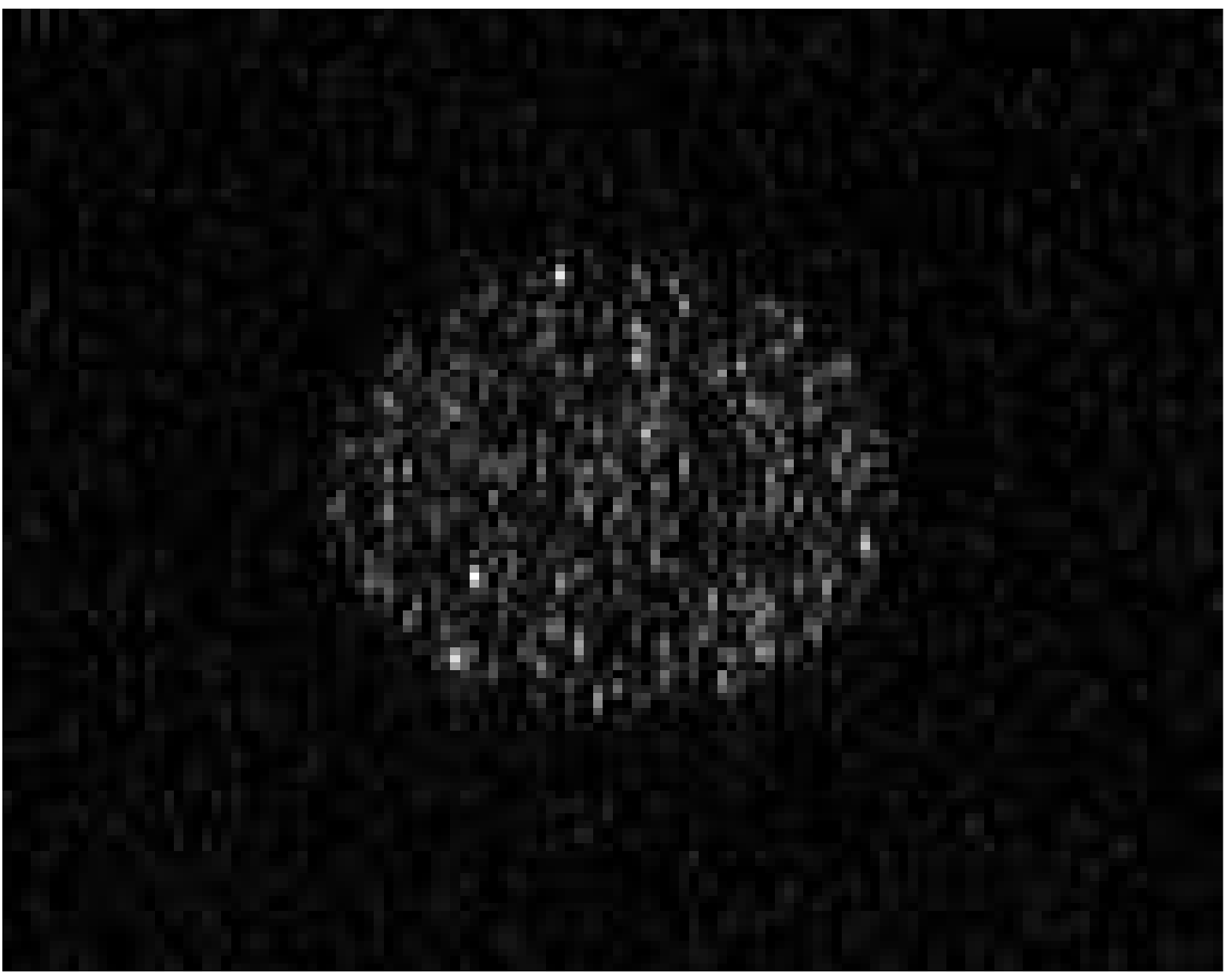} \label{fig:bmode_Deconvl10}}
\subfigure[$\rm{Deconv_{EM}}$]{\includegraphics[width=0.18\linewidth]{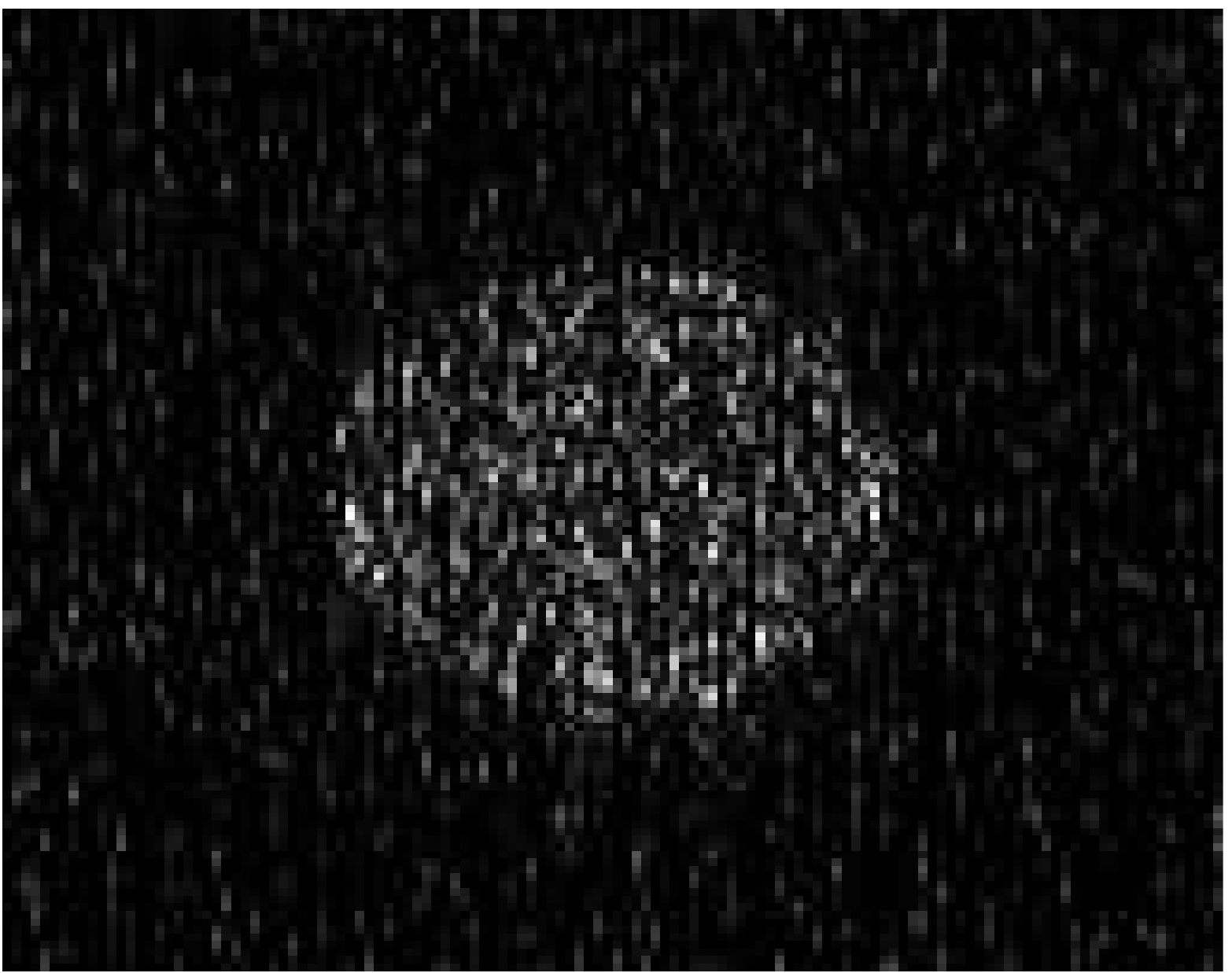} \label{fig:bmode_Deconvem0}}
\subfigure[$\rm{Deconv_{MCMC}}$]{\includegraphics[width=0.18\linewidth]{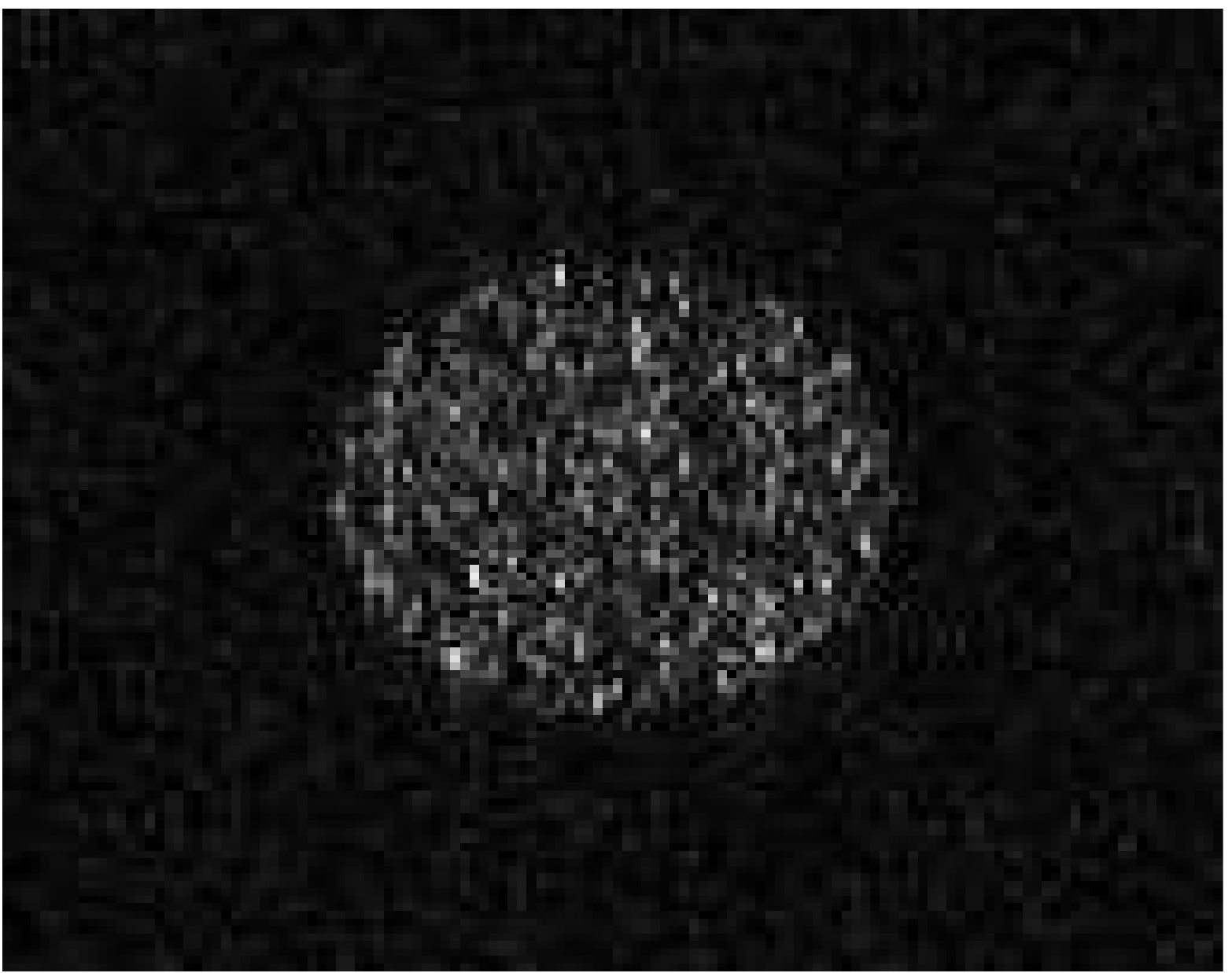} \label{fig:bmode_Deconvmcmc0}}
\subfigure[$\rm{Joint_{MCMC}}$]{\includegraphics[width=0.18\linewidth]{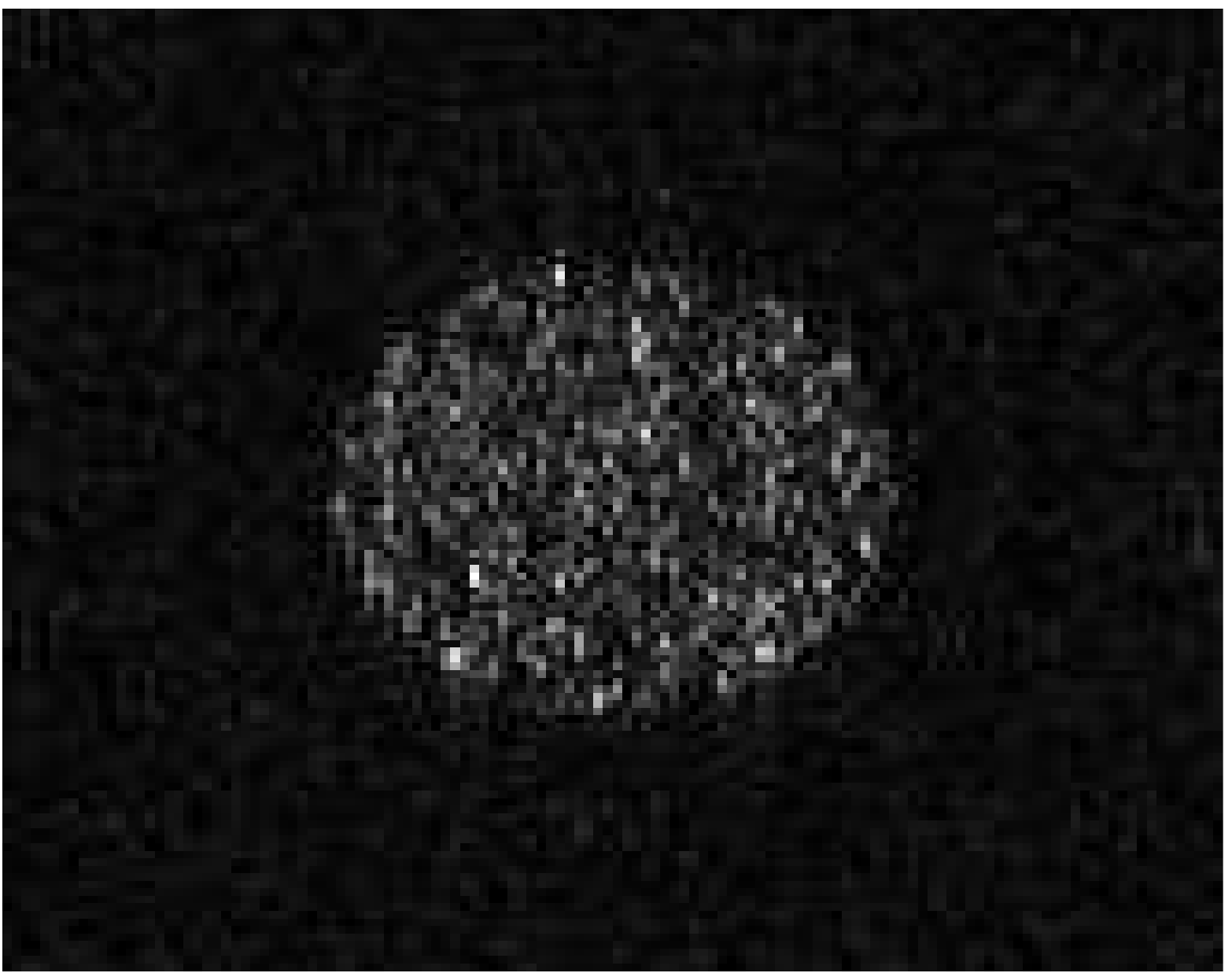}\label{fig:bmode_Jointmcmc0}}
\subfigure[$\rm{Joint_{MCMC}}$]{\includegraphics[width=0.18\linewidth]{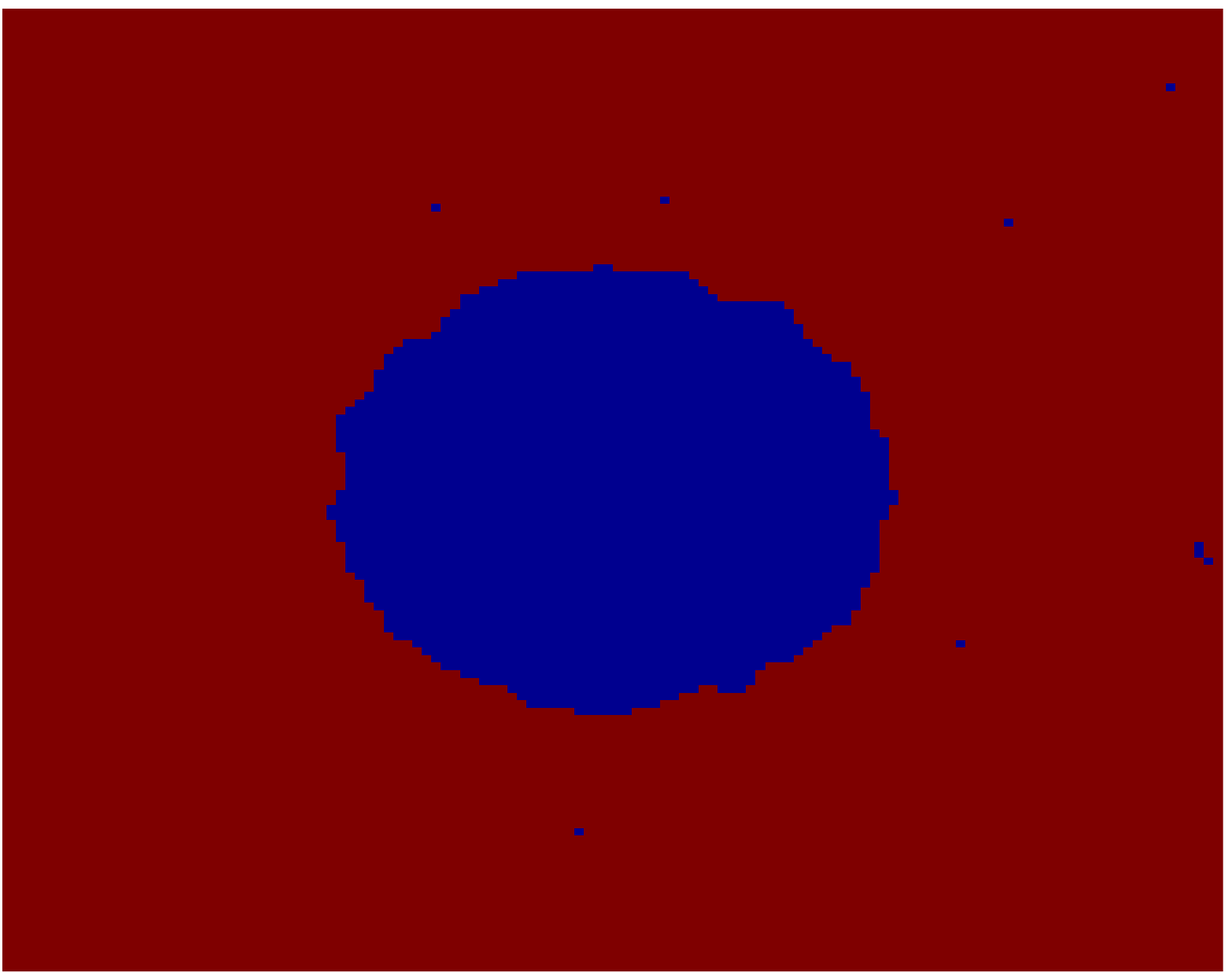} \label{fig:zmap_Jointmcmc0}} 
\caption{Group 1: (a) Simulated PSF; (b) Ground truth of the TRF; (c) Ground truth for label map; (d) Observed B-mode image; (e)-(i) Estimated TRFs in B-mode form obtained with methods $\ell_2$, $\ell_1$, $\rm{Deconv_{EM}}$, $\rm{Deconv_{MCMC}}$ and the proposed $\rm{Joint_{MCMC}}$; (j) Estimated label map obtained with the proposed method (regularization parameters for the $\ell_2$ and $\ell_1$ methods set to $0.01$ and $0.1$).}
\label{fig:1circle}
\end{figure*}

\rev{Experiments were first conducted on three groups of simulated US images with a simulation scenario inspired by \cite{JamesNg2007}.} The PSF was simulated with a realistic state-of-the-art ultrasound simulator Field II \cite{Field1996} corresponding to a 3.5 MHz linear probe as shown in Fig. \ref{fig:psf0}. All images were simulated with  the same PSF. \rev{All the simulation results presented hereinafter were obtained using $6000$ Monte Carlo iterations, including a burn-in period of $2000$ iterations.}
\paragraph{Group 1}
The TRF $\bfx$ mimicking a hyperechoic (bright) round inclusion into an homogeneous medium was blurred by the simulated PSF and contaminated by an AWGN with BSNR $=30$ dB. \rev{The simulated images are of size $128\times128$.} 
The pixels located inside and outside the inclusion, indicated by the label map in Fig. \ref{fig:mask0}, are distributed according to GGDs with parameters ($\xi,\gamma$) = (0.6,1) (inside) and ($\xi,\gamma$) = (1.8,2) (outside) as highlighted in Fig. \ref{fig:refl0}.
The simulated observed B-mode image (log-compressed envelop image \rev{of the corresponding beamformed RF data} which is commonly used for visualization purpose in US imaging) is shown in Fig. \ref{fig:bmode_y0}. \rev{The quality of the deconvolution can be appreciated by comparing the estimated TRFs shown in Figs. \ref{fig:bmode_Deconvl20}-\ref{fig:bmode_Jointmcmc0} obtained with the methods $\ell_2$, $\ell_1$, $\rm{Deconv_{EM}}$, $\rm{Deconv_{MCMC}}$ and the proposed $\rm{Joint_{MCMC}}$. The quality of the segmentation can be observed in Fig. \ref{fig:zmap_Jointmcmc0}, which shows the estimated label map obtained with the method $\rm{Joint_{MCMC}}$.} Finally, the performance of the GGD parameter estimators is illustrated by the histograms of the generated GGD parameters ($\bsxi,\bsgamma$) displayed in Fig. \ref{fig:1circle_samples}, where the red and green vertical lines indicate the MMSE estimates and the true values of the parameters, respectively. 

\begin{figure*}
\centering
\subfigure[\rev{TRF}]{\includegraphics[width=0.18\linewidth]{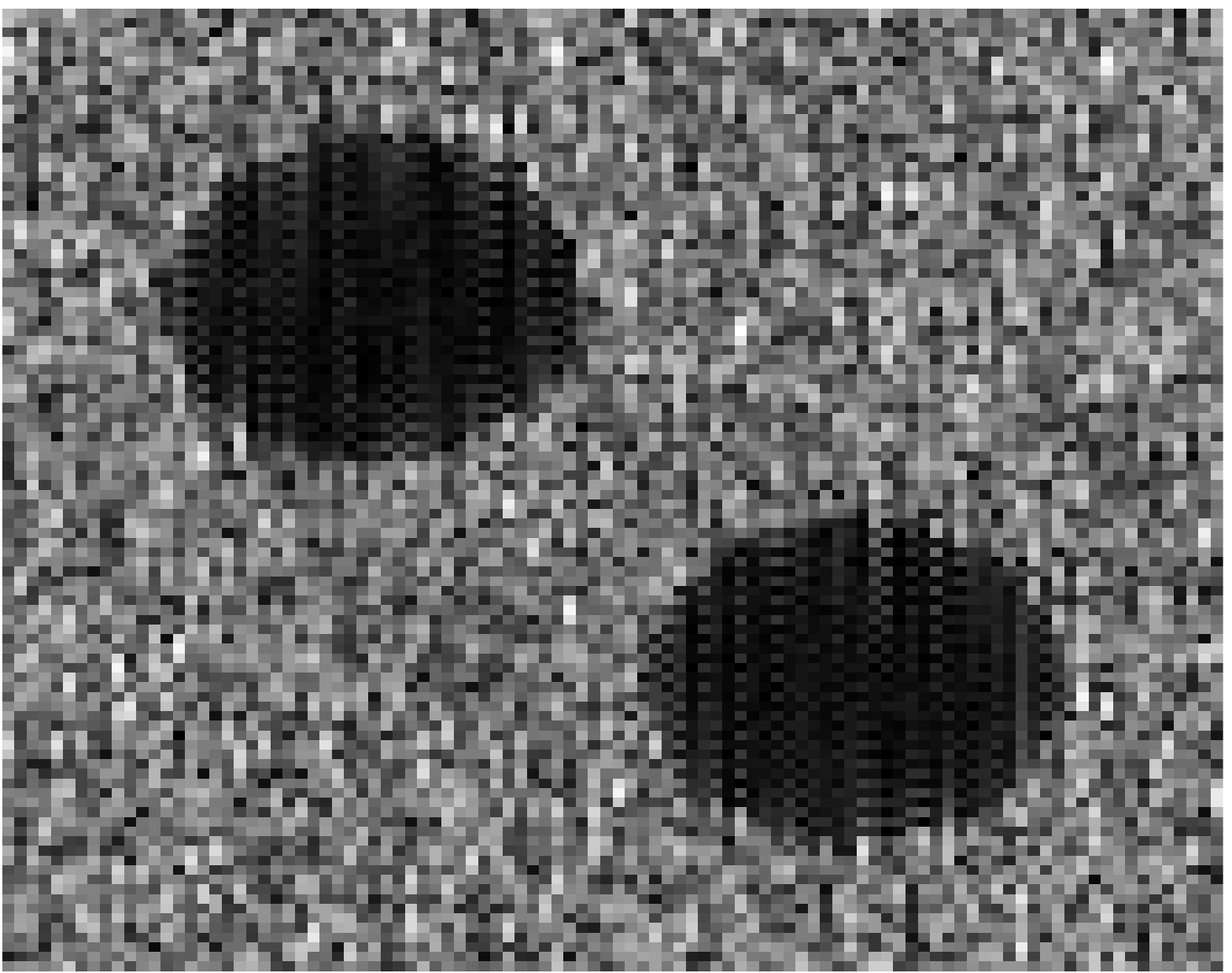}\label{fig:refl}}
\subfigure[Label]{\includegraphics[width=0.18\linewidth]{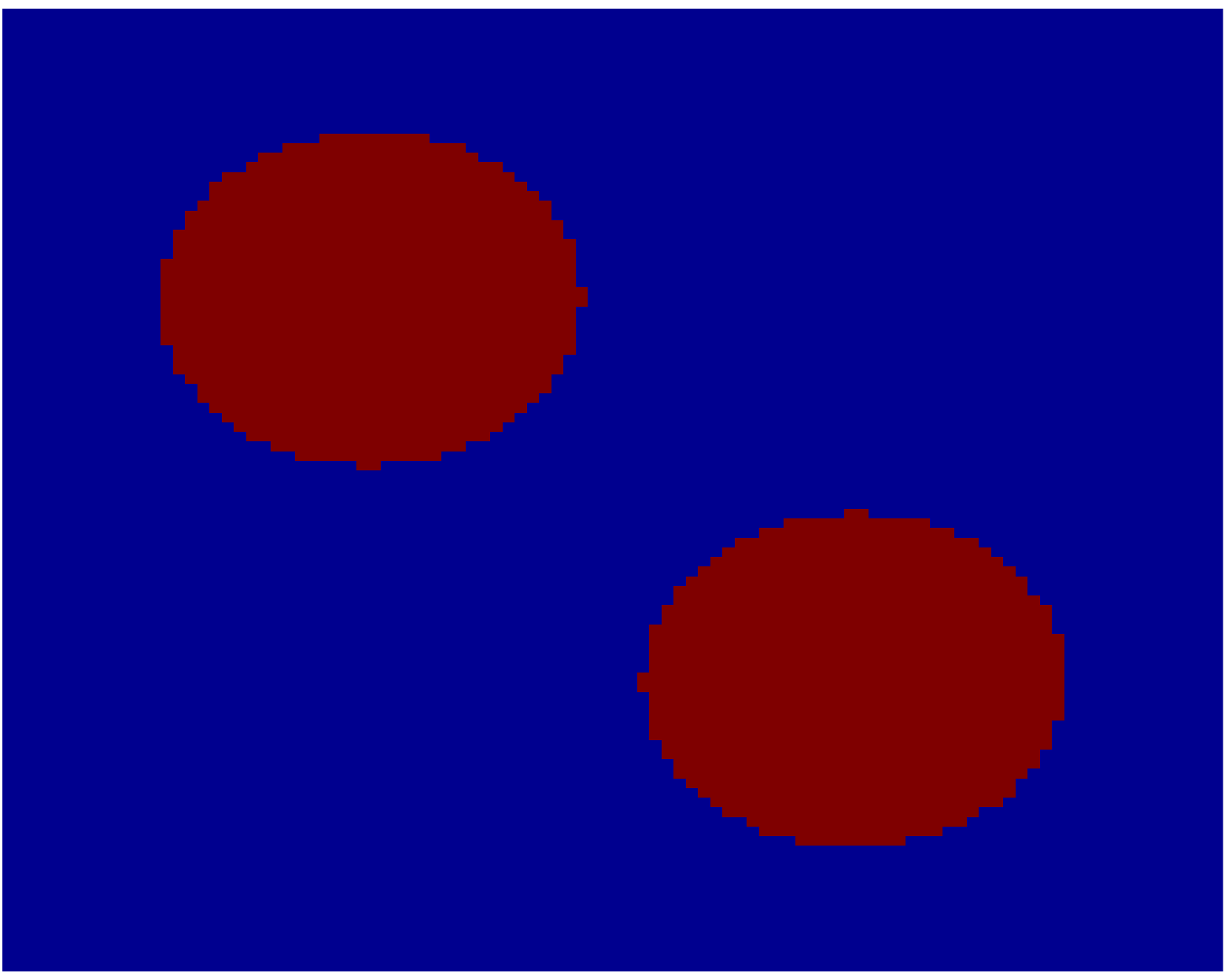} \label{fig:mask}}
\subfigure[B-mode]{\includegraphics[width=0.18\linewidth]{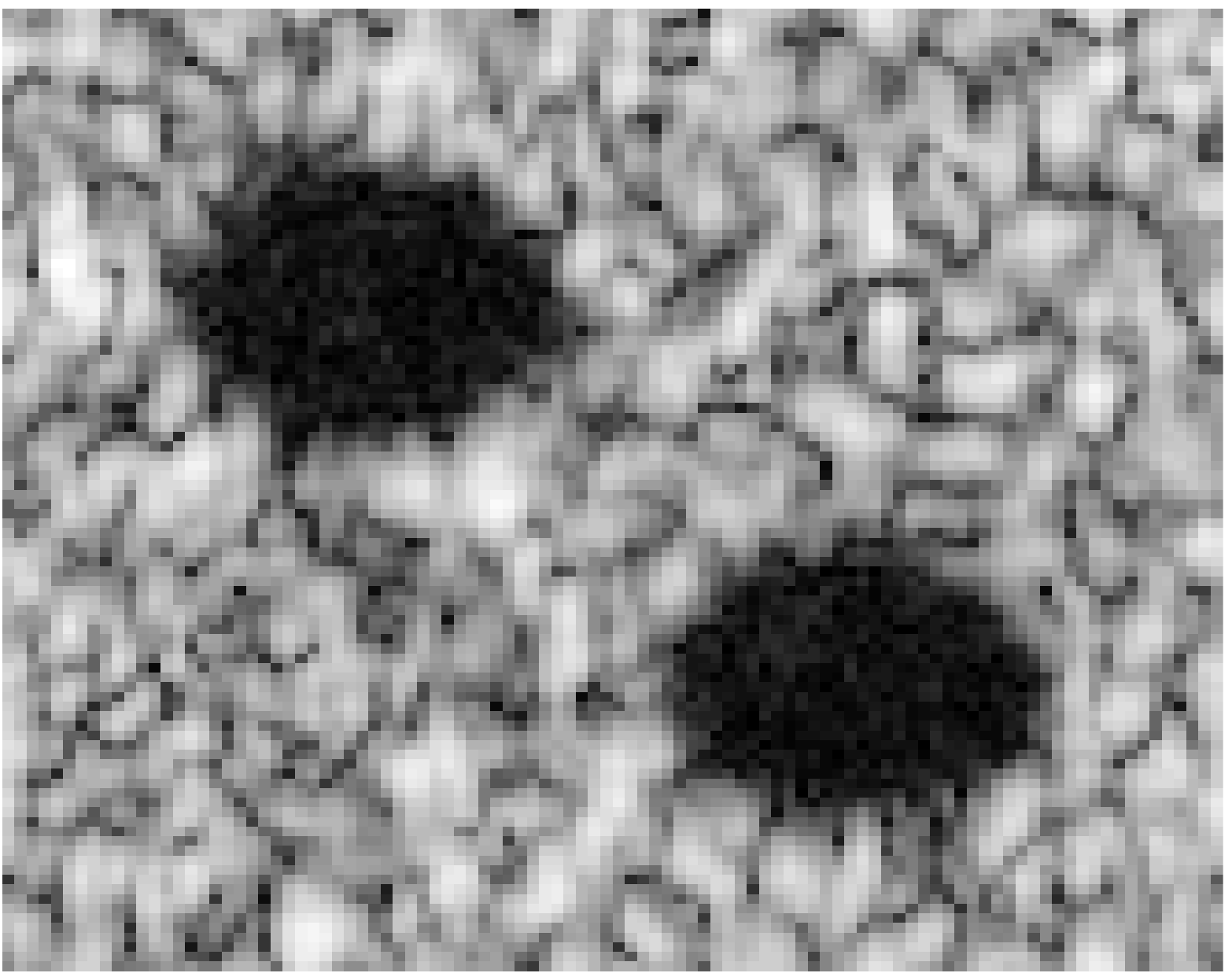} \label{fig:bmode_y}}
\subfigure[$\ell_2$]{\includegraphics[width=0.18\linewidth]{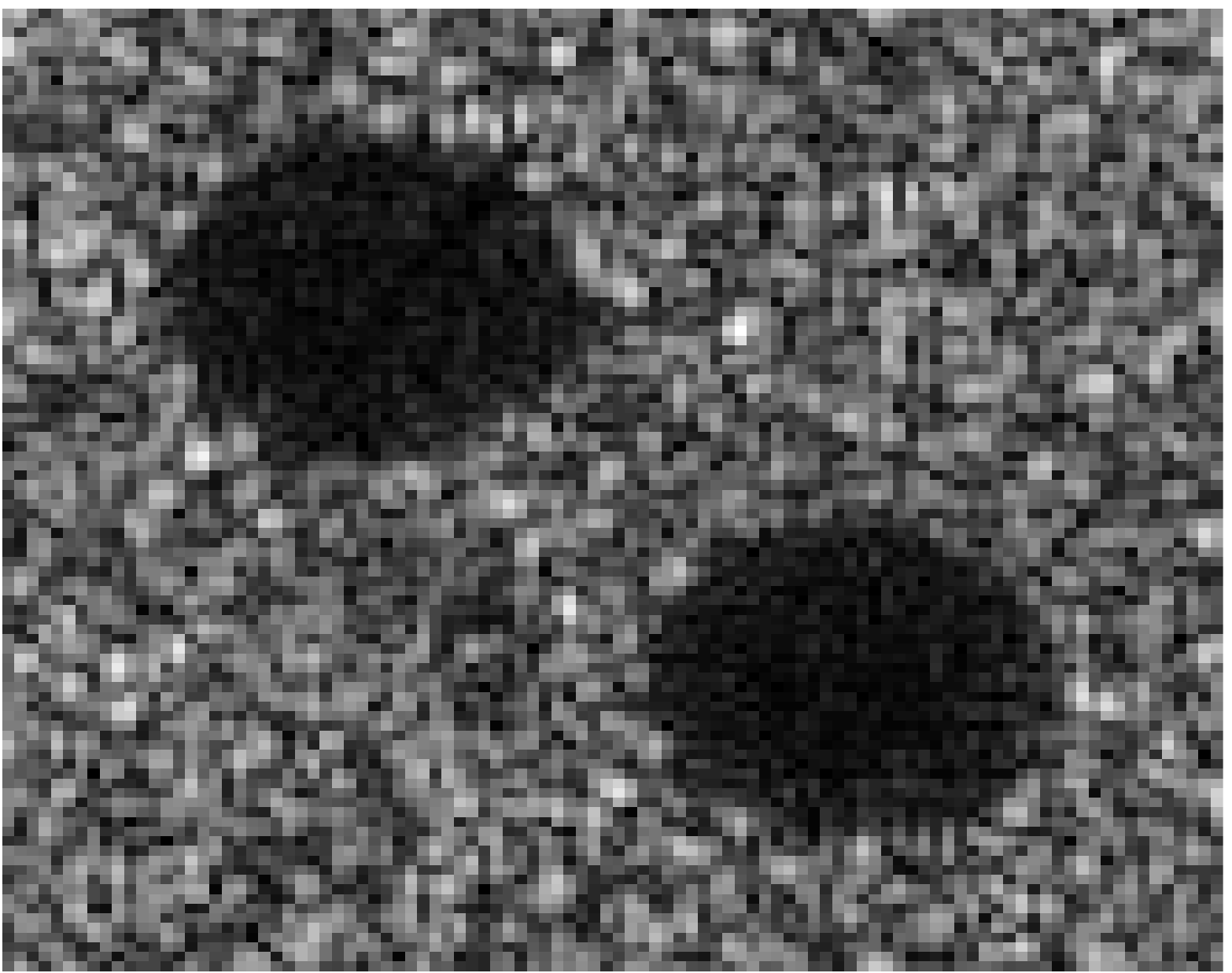} \label{fig:bmode_Deconvl2}}\\
\subfigure[$\ell_1$]{\includegraphics[width=0.18\linewidth]{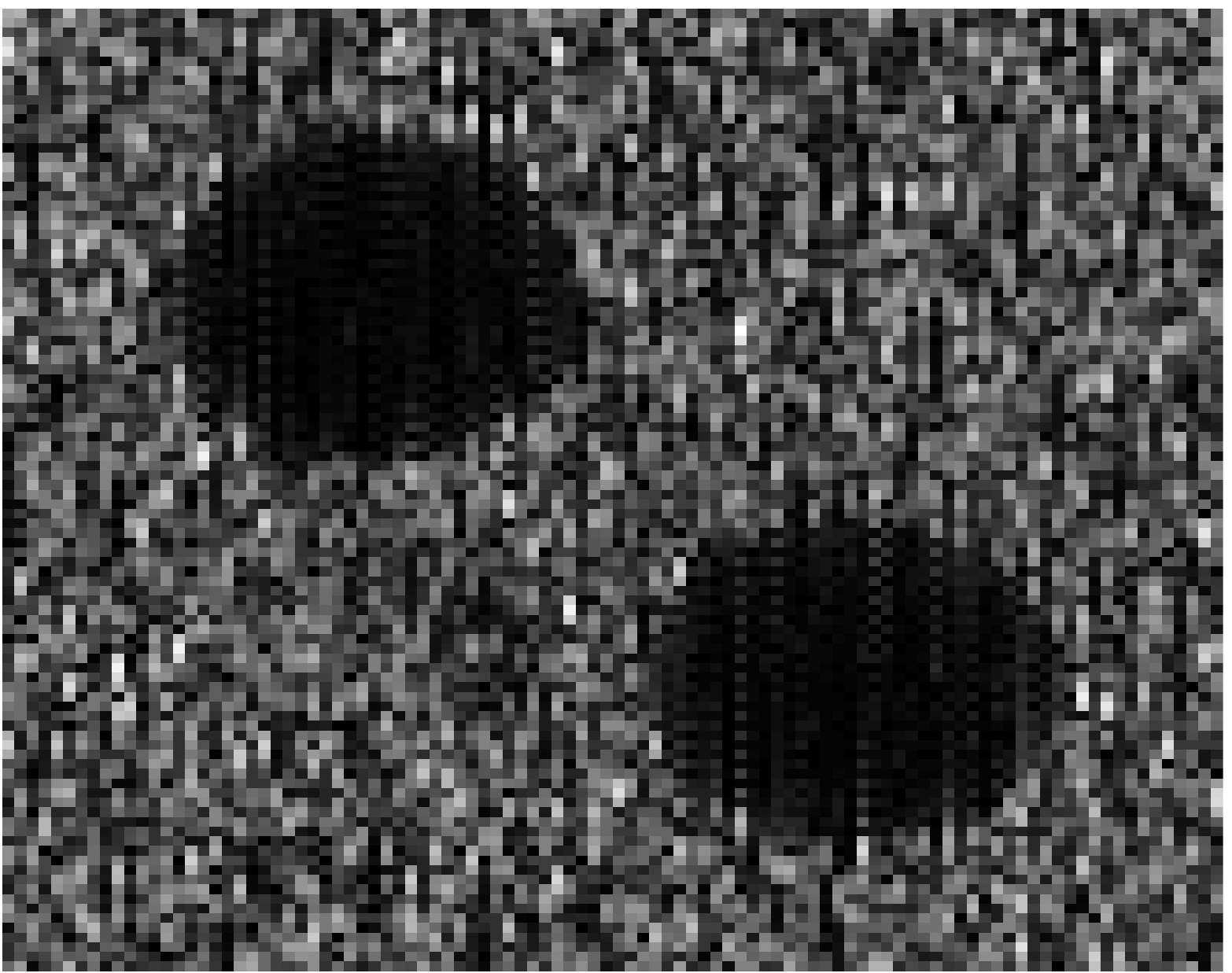} \label{fig:bmode_Deconvl1}}
\subfigure[$\rm{Deconv_{EM}}$]{\includegraphics[width=0.18\linewidth]{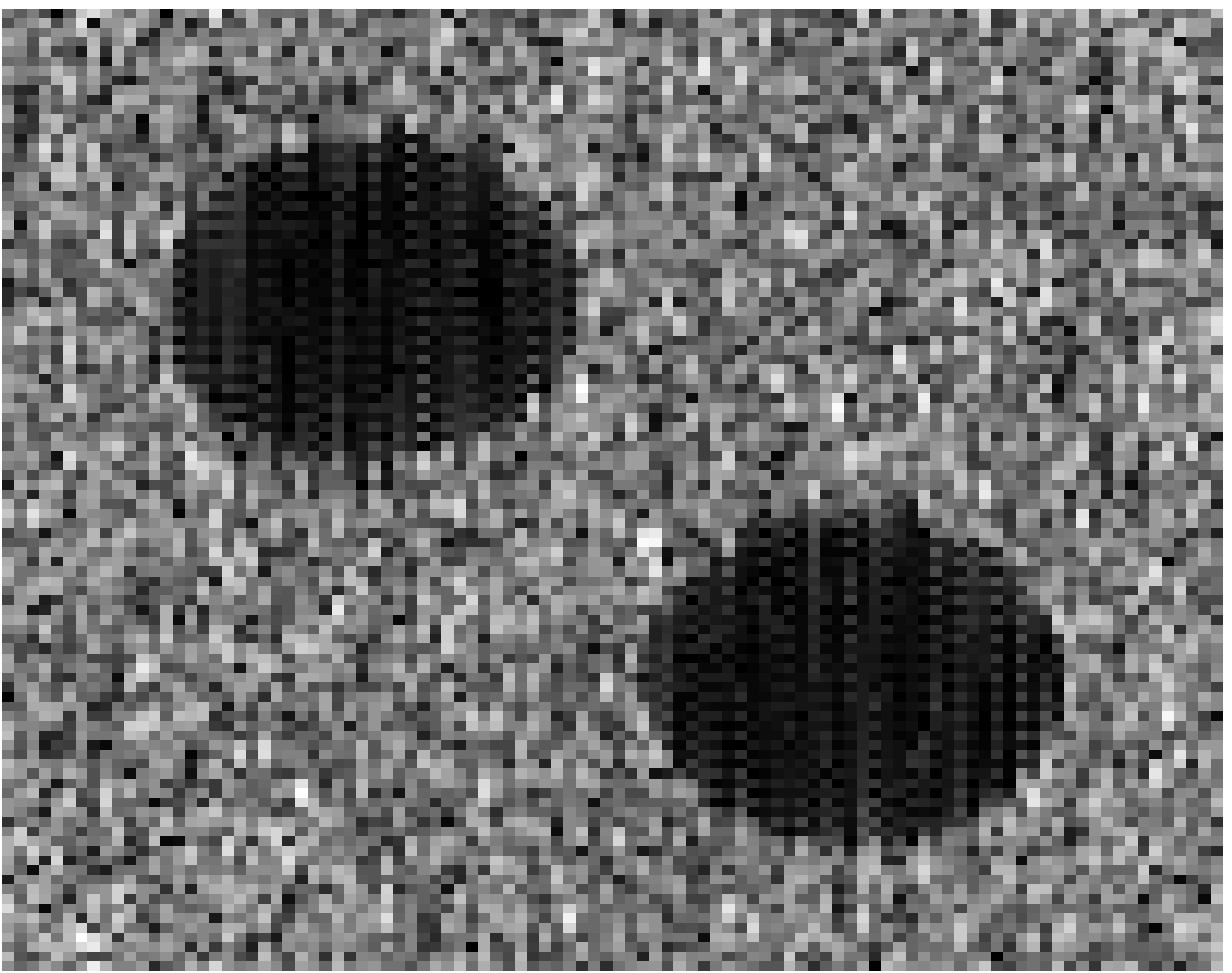} \label{fig:bmode_Deconvem}}
\subfigure[$\rm{Deconv_{MCMC}}$]{\includegraphics[width=0.18\linewidth]{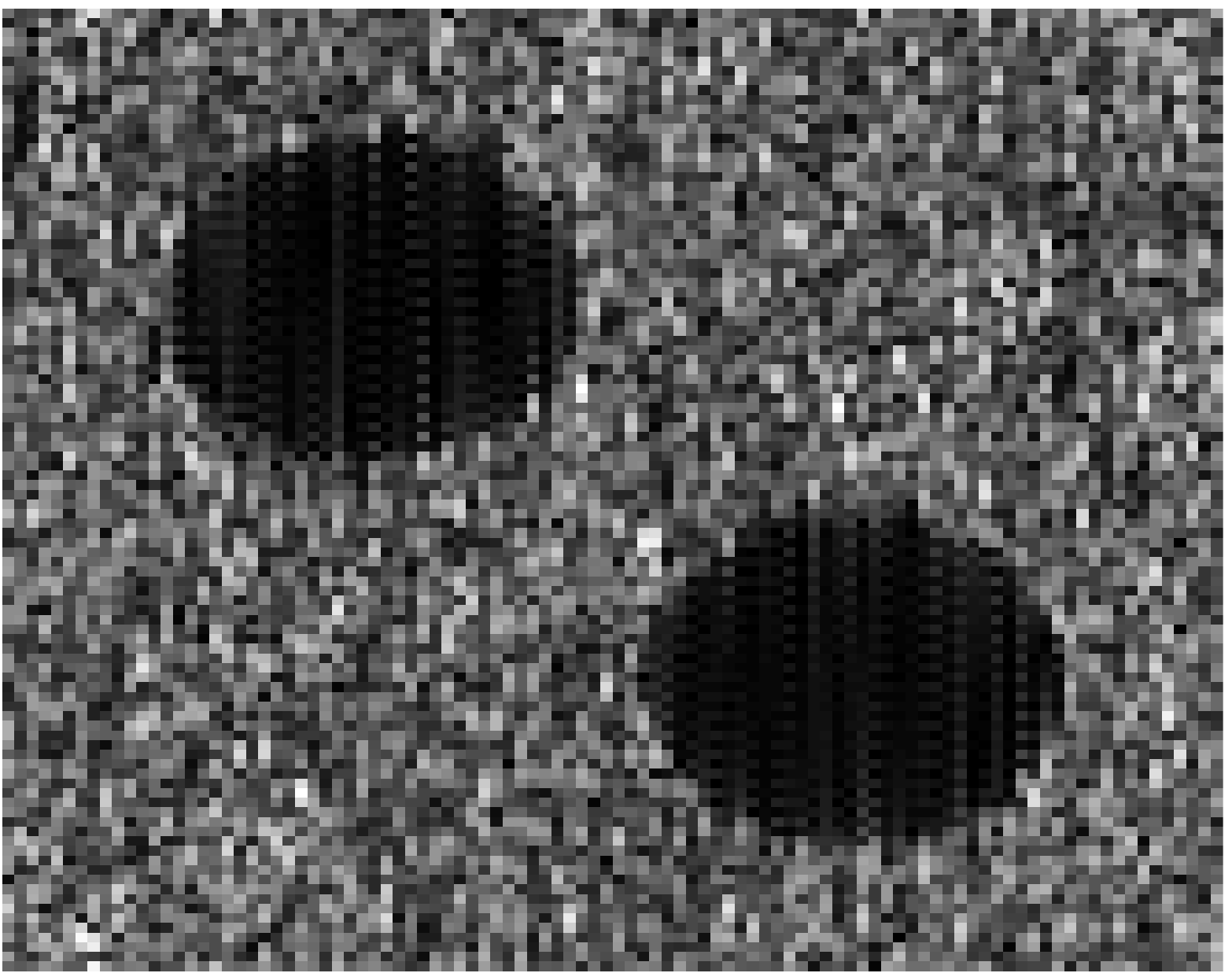} \label{fig:bmode_Deconvmcmc}}
\subfigure[$\rm{Joint_{MCMC}}$]{\includegraphics[width=0.18\linewidth]{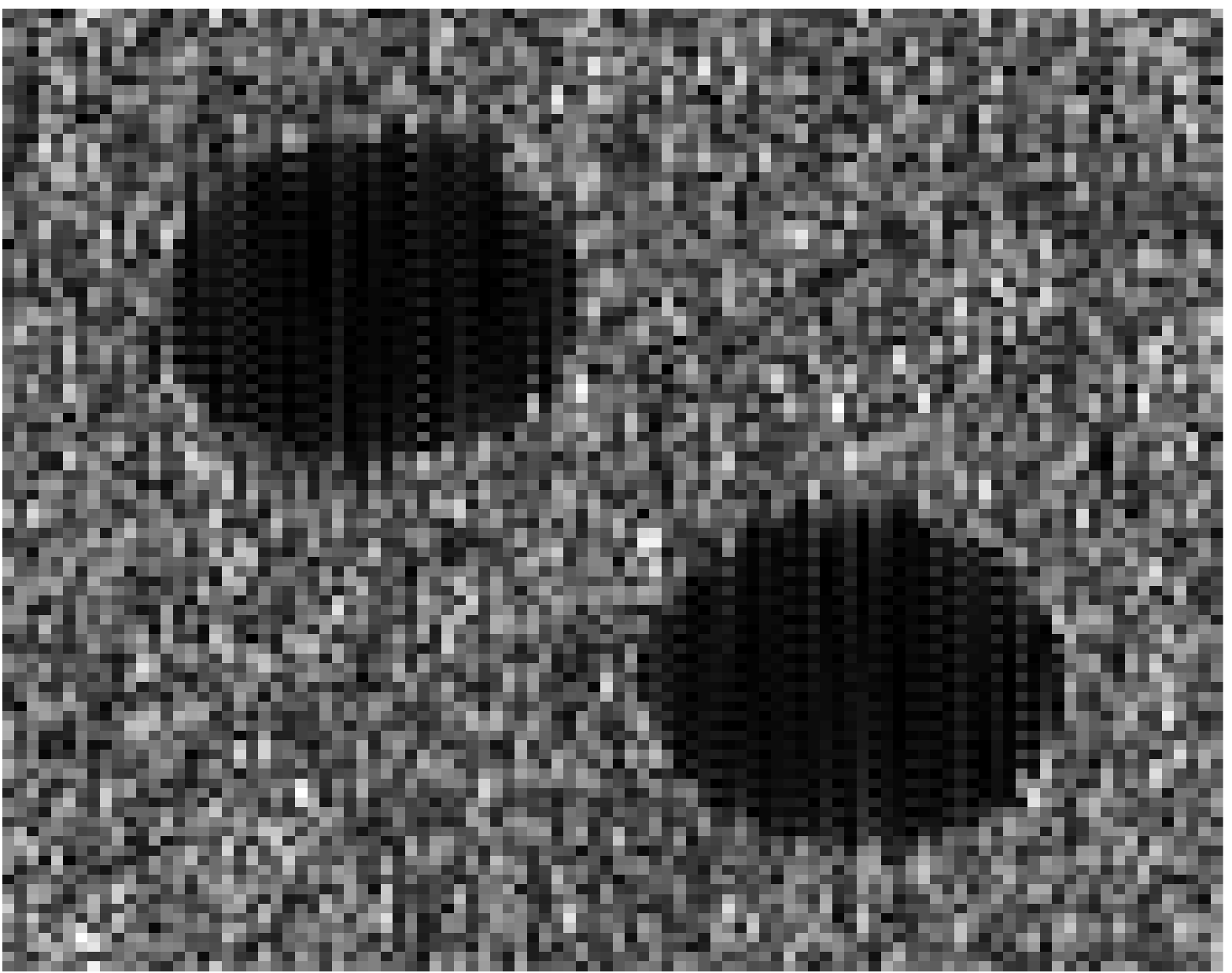}\label{fig:bmode_Jointmcmc}} 
\subfigure[$\rm{Joint_{MCMC}}$]{\includegraphics[width=0.18\linewidth]{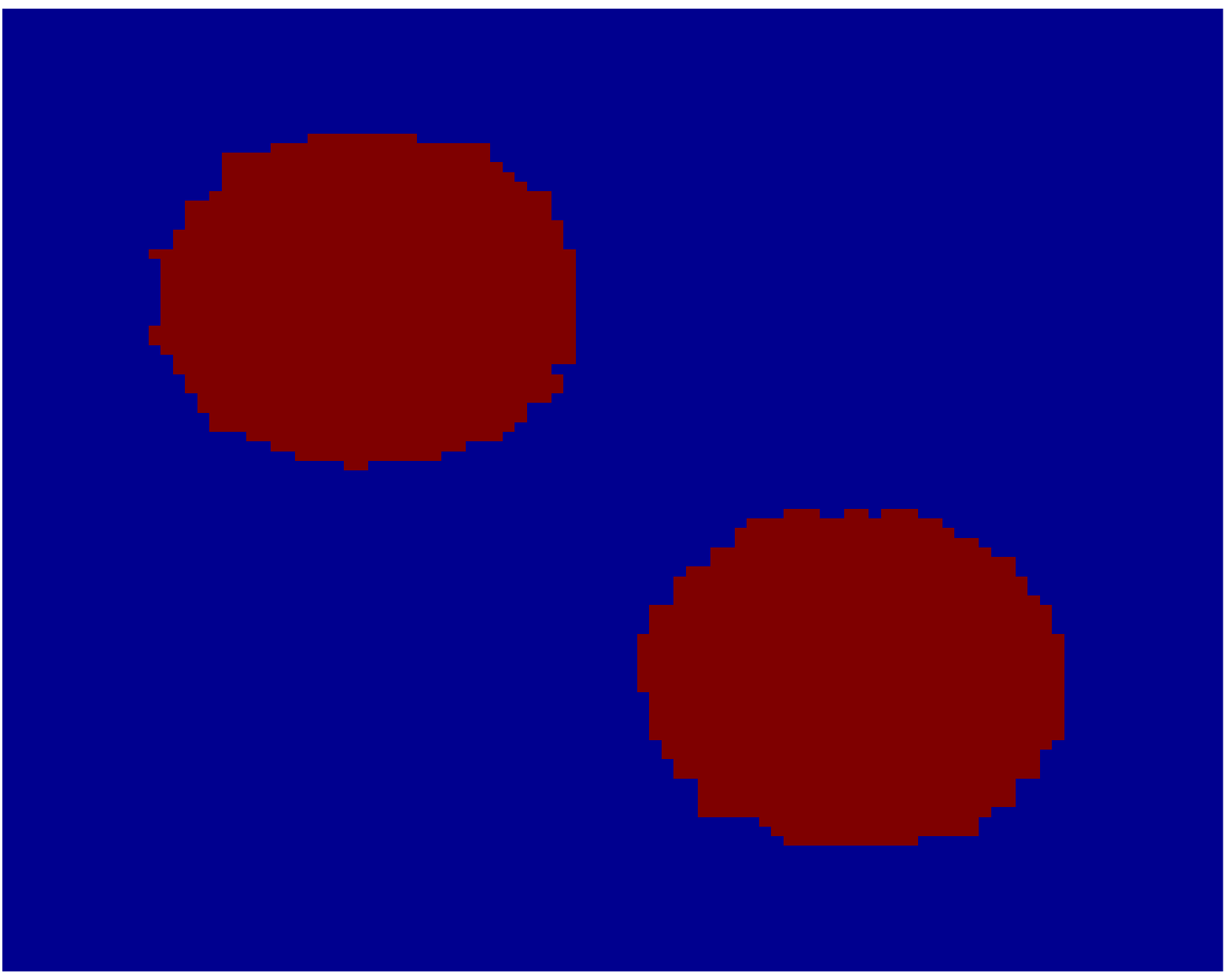} \label{fig:zmap_Jointmcmc}}
\caption{Group 2: (a) Ground truth of the TRF; (b) Ground truth for label map; (c) Observed B-mode image; 
(d)-(h) Estimated TRFs in B-mode form obtained with the methods $\ell_2$, $\ell_1$, $\rm{Deconv_{EM}}$, $\rm{Deconv_{MCMC}}$ and the proposed $\rm{Joint_{MCMC}}$; (i) Estimated label map obtained with the proposed method (regularization parameters for the $\ell_2$ and $\ell_1$ methods set to $0.1$ and $1$).}
\end{figure*}

\paragraph{Group 2}
The TRF $\bfx$ \zhao{is an homogeneous medium with two hypoechoic (dark) round inclusions} (see Fig. \ref{fig:refl}) that was blurred by the same simulated PSF and contaminated by an AWGN. The size of the US reflectivity image is $100\times100$ and BSNR $=30$ dB.
The pixels located inside and outside the inclusions are distributed according to GGDs with parameter vectors ($\xi,\gamma$) = (0.8,10) (inside) and ($\xi,\gamma$) = (1.5,1) (outside) as highlighted in Fig. \ref{fig:refl}.
The simulated observed B-mode image is shown in Fig. \ref{fig:bmode_y} whereas the ground truth of the label map is given in Fig. \ref{fig:mask}. 
\rev{Figs. \ref{fig:bmode_Deconvl2}-\ref{fig:bmode_Jointmcmc} show the estimated TRFs obtained with the methods $\ell_2$, $\ell_1$, $\rm{Deconv_{EM}}$, $\rm{Deconv_{MCMC}}$ and the proposed $\rm{Joint_{MCMC}}$, confirming the good performance of  $\rm{Joint_{MCMC}}$ for the deconvolution of US images. The estimated label map obtained with the method $\rm{Joint_{MCMC}}$ is shown in Fig. \ref{fig:zmap_Jointmcmc}, confirming its good segmentation performance.} Finally, the hyperparameter estimates of Group 2 are shown in Table \ref{Hyper}, confirming the good estimation performance. 

\begin{table}[htpt]
\caption{Hyperparameter Estimations for Simulated data (Group 2)}
\label{Hyper}
\centering
\begin{tabular}{|c|c|c|c|c|}
\hline
 Method                 & $\xi_1$  &$\xi_2$  & $\gamma_1$ & $\gamma_2$ \\
 \hline
Ground truth            &  0.8     &  1.5     &  10       & 1\\
\hline
$\rm{Deconv_{EM}}$      & 0.60   & 0.96     &  21.10 & 0.42 \\
\hline
$\rm{Deconv_{MCMC}}$    & 0.80   &  2.15 &  10.05   & 1.50\\
\hline 
$\rm{Joint_{MCMC}}$     &  0.82  &  1.37 &  11.24   & 0.82\\
\hline
\end{tabular}
\end{table}

\begin{figure*}
\centering
\subfigure[\rev{TRF}]{\includegraphics[width=0.18\linewidth]{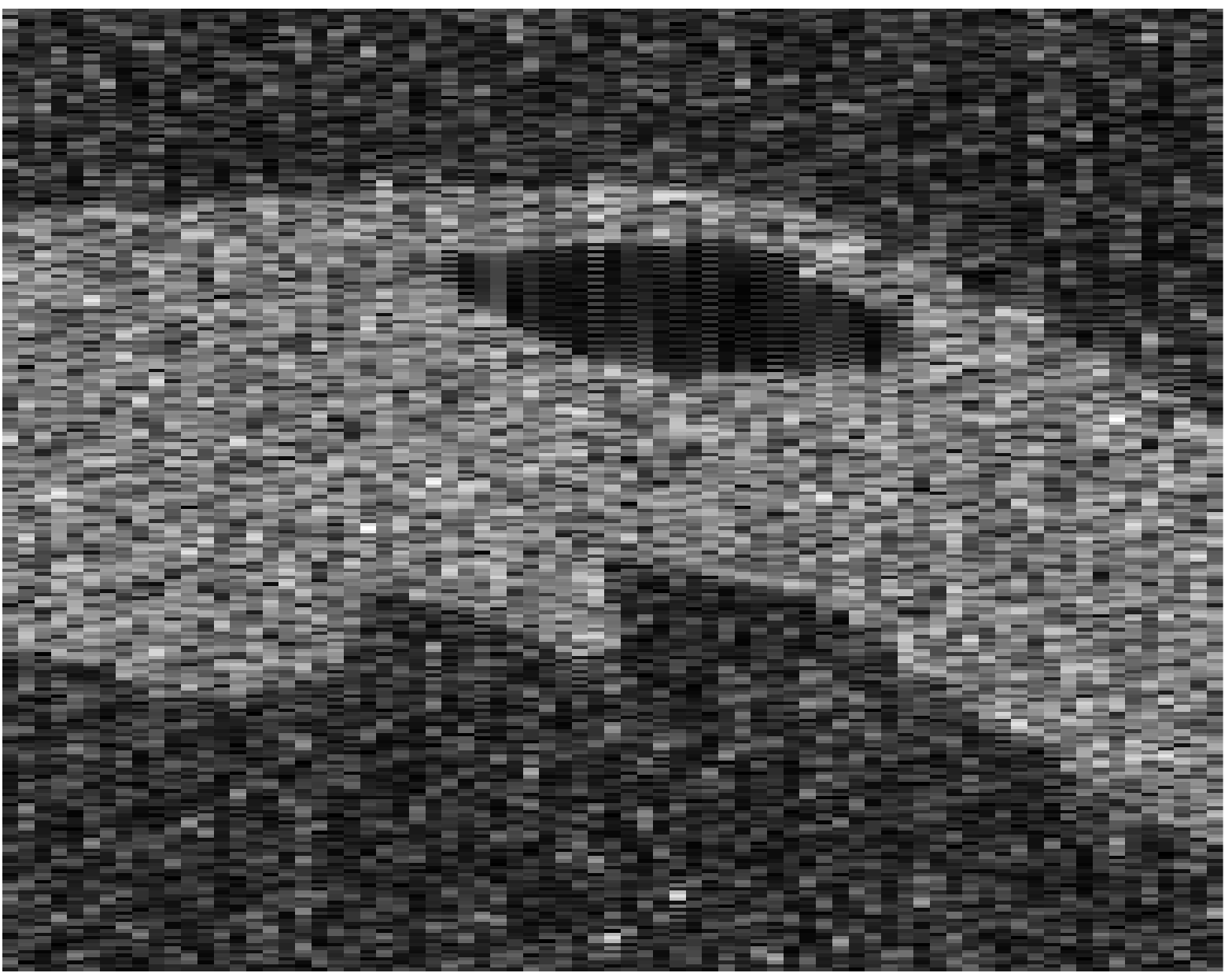}\label{fig:refl2}}
\subfigure[Label]{\includegraphics[width=0.18\linewidth]{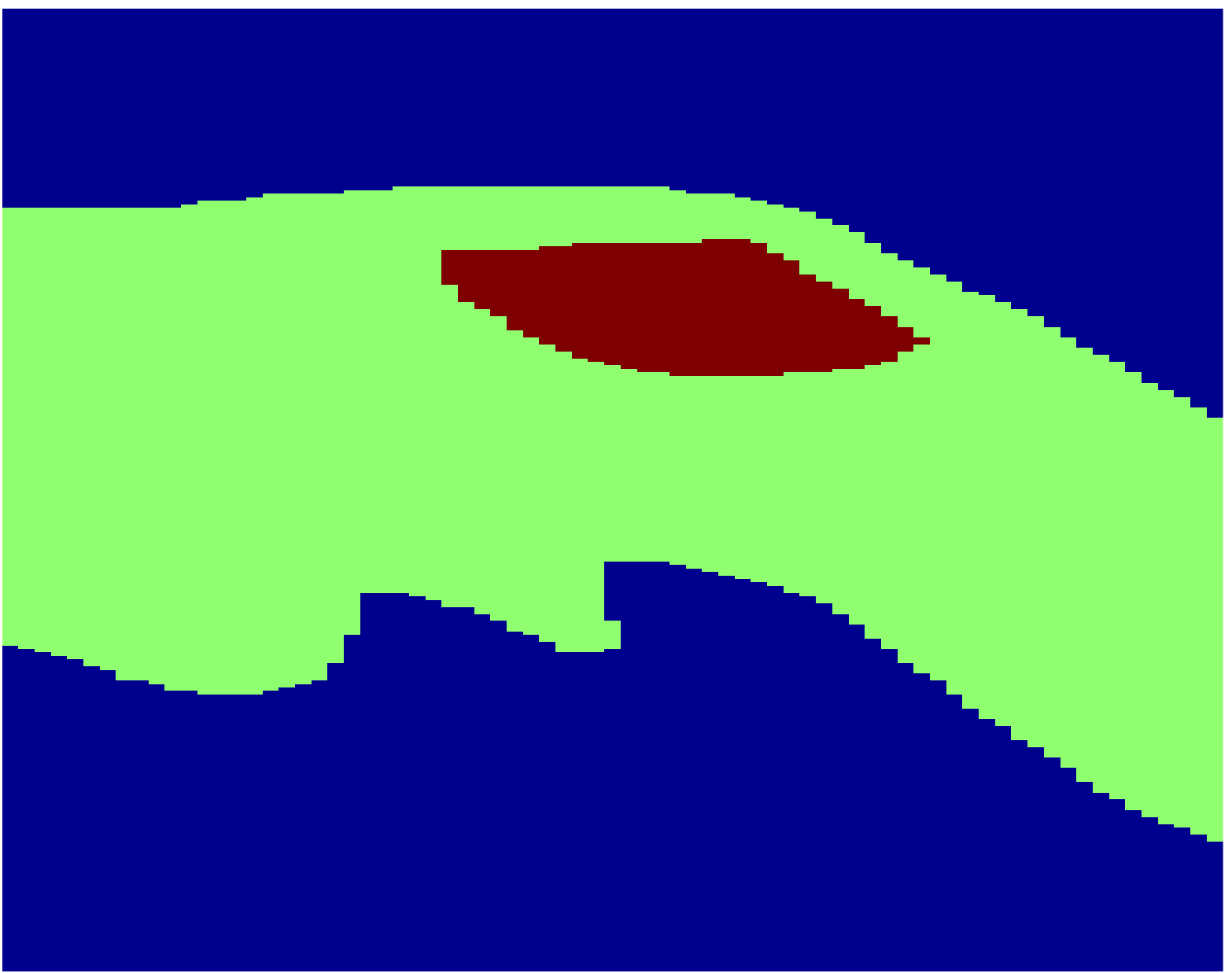} \label{fig:mask2}}
\subfigure[B-mode]{\includegraphics[width=0.18\linewidth]{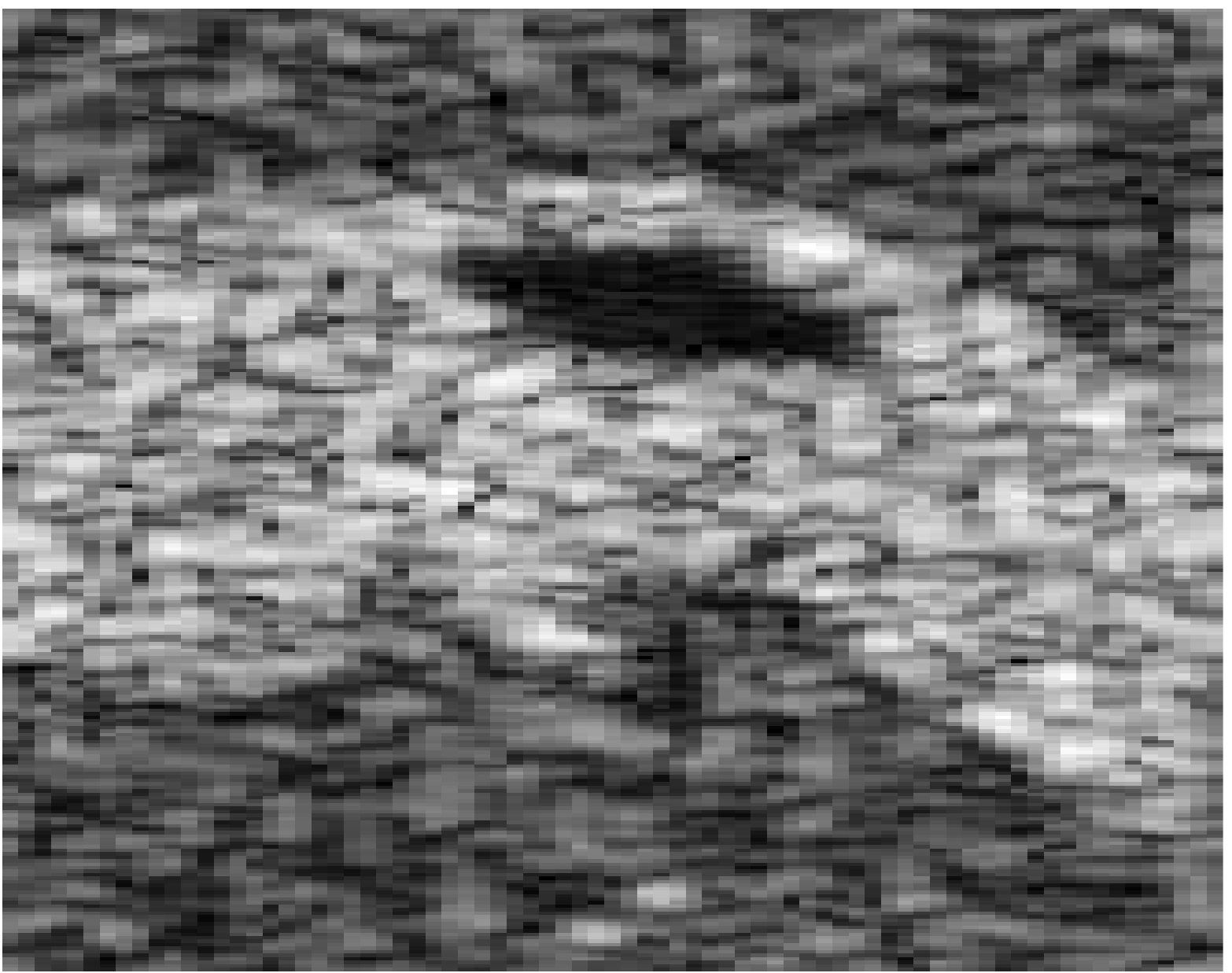} \label{fig:bmode_y2}}
\subfigure[$\ell_2$]{\includegraphics[width=0.18\linewidth]{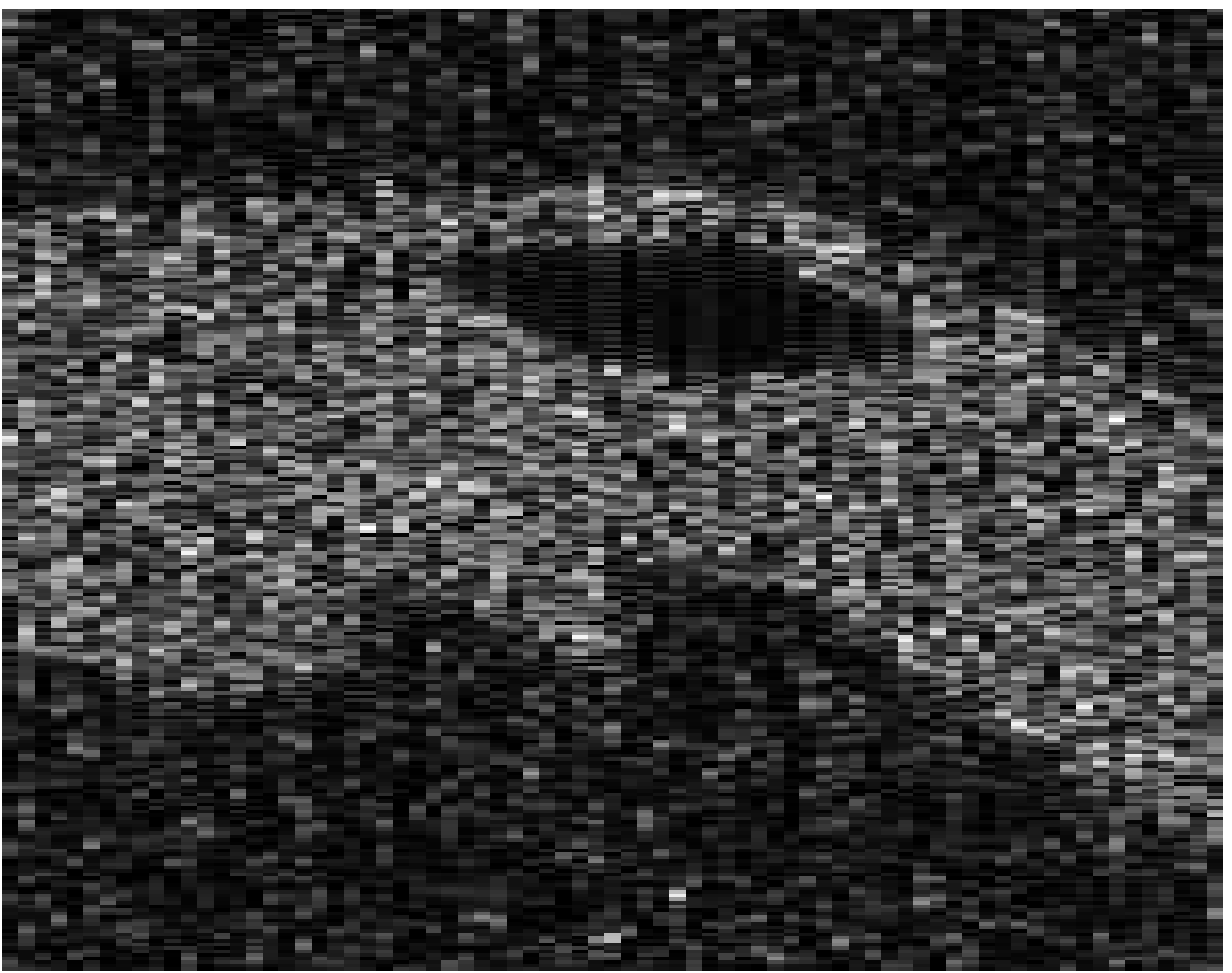} \label{fig:bmode_Deconvl22}}\\
\subfigure[$\ell_1$]{\includegraphics[width=0.18\linewidth]{figures/part2/bmode_new_l1} \label{fig:bmode_Deconvl12}}
\subfigure[$\rm{Deconv_{EM}}$]{\includegraphics[width=0.18\linewidth]{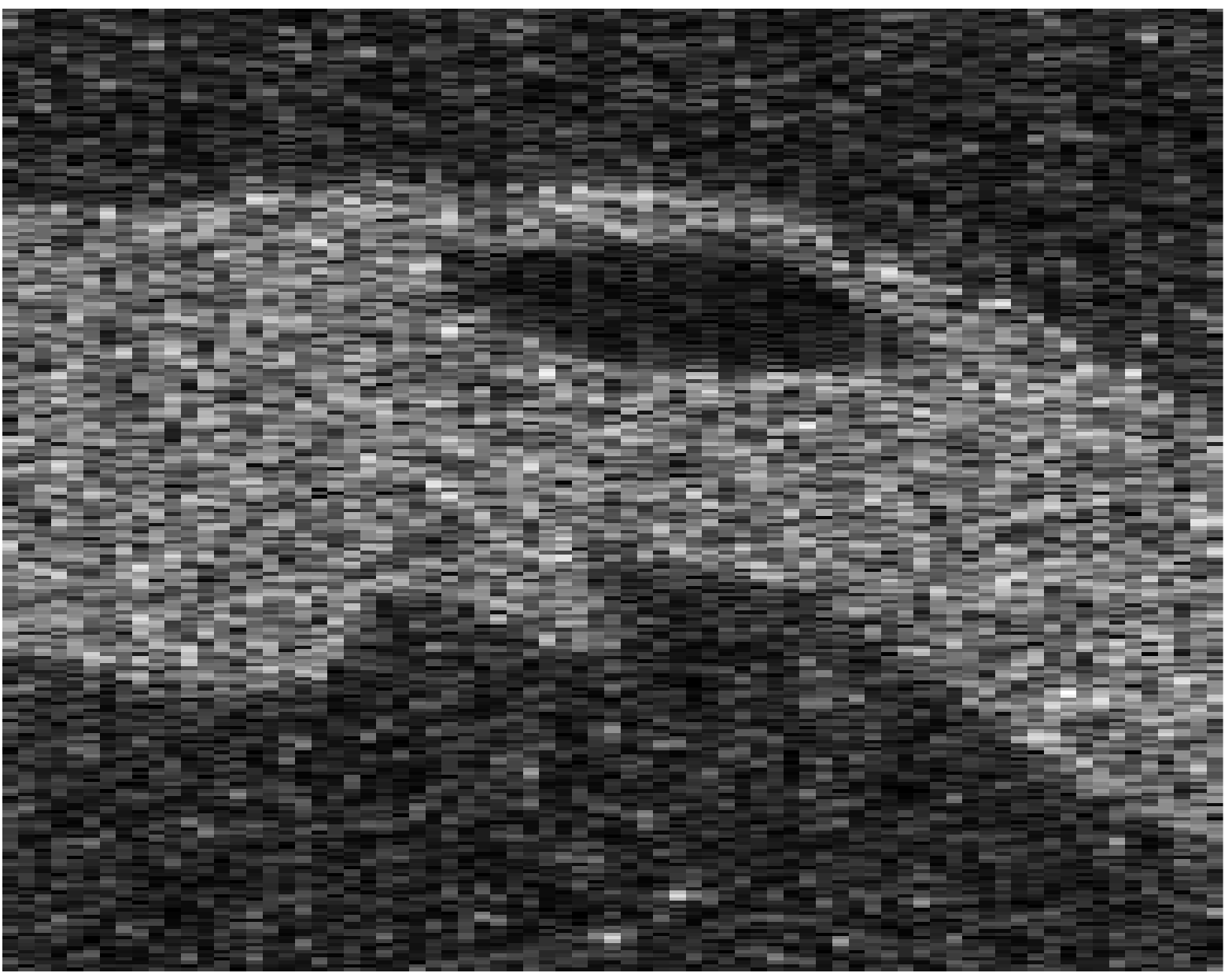} \label{fig:bmode_Deconvem2}}
\subfigure[$\rm{Deconv_{MCMC}}$]{\includegraphics[width=0.18\linewidth]{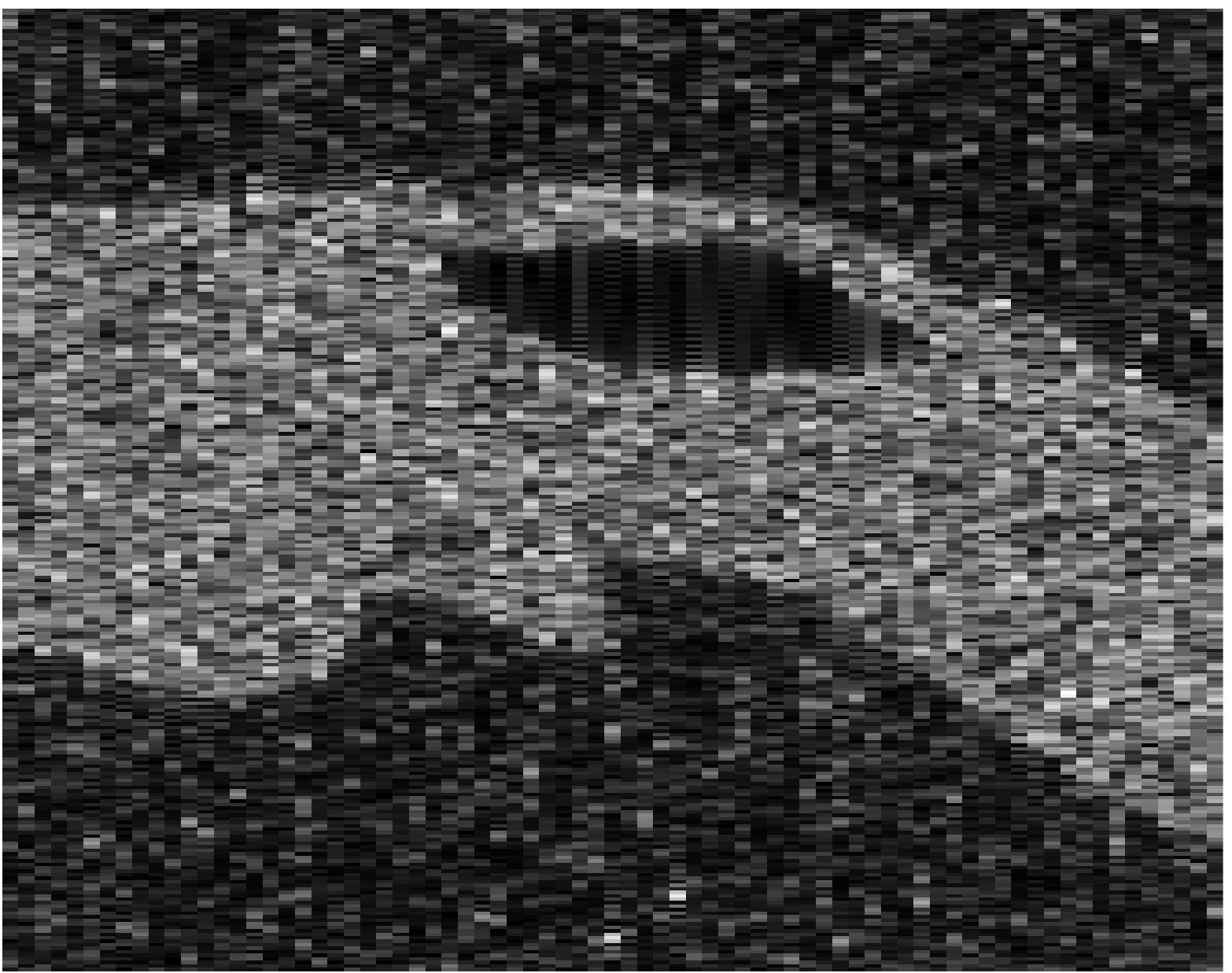} \label{fig:bmode_Deconvmcmc2}}
\subfigure[$\rm{Joint_{MCMC}}$]{\includegraphics[width=0.18\linewidth]{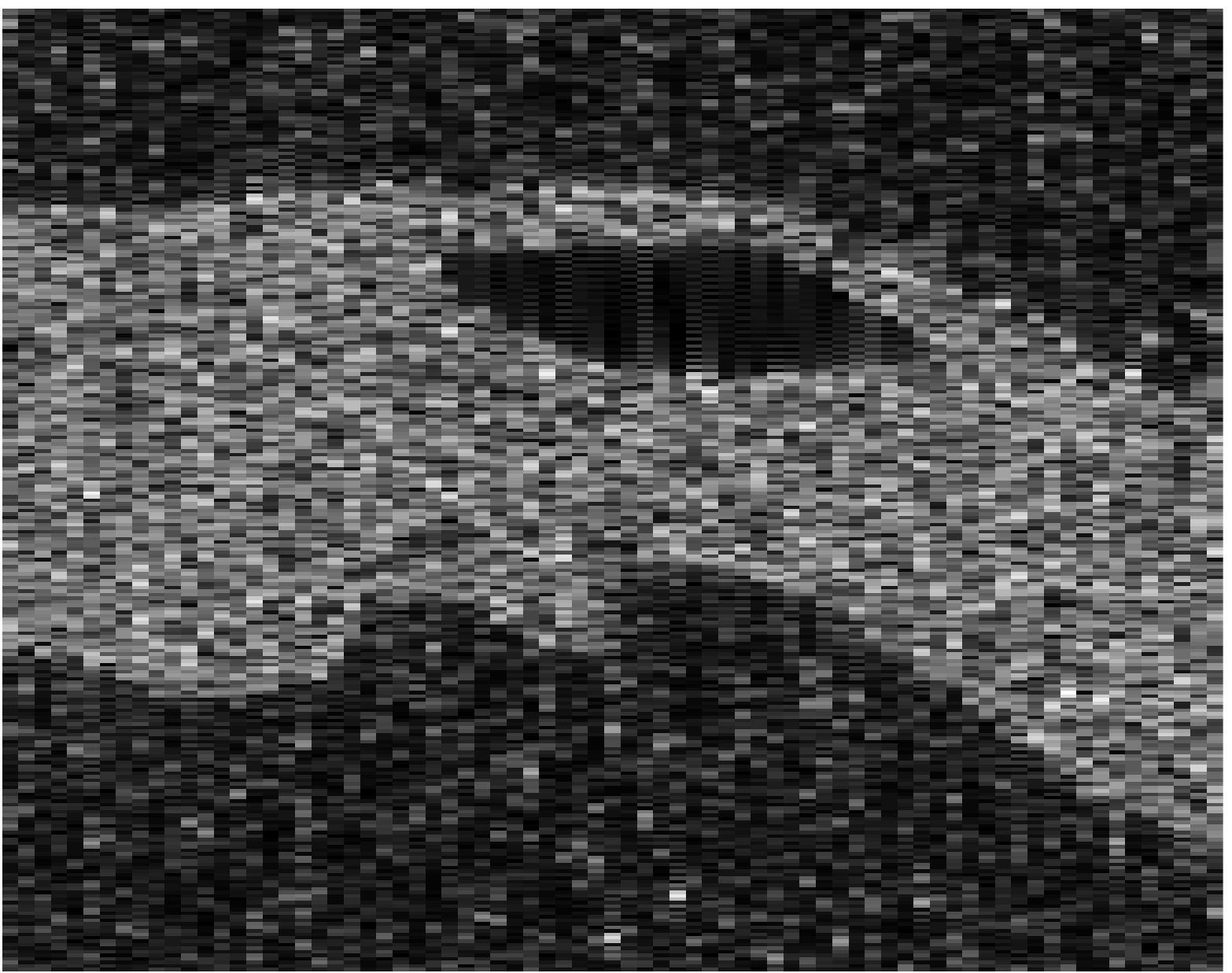}\label{fig:bmode_Jointmcmc2}} 
\subfigure[$\rm{Joint_{MCMC}}$]{\includegraphics[width=0.18\linewidth]{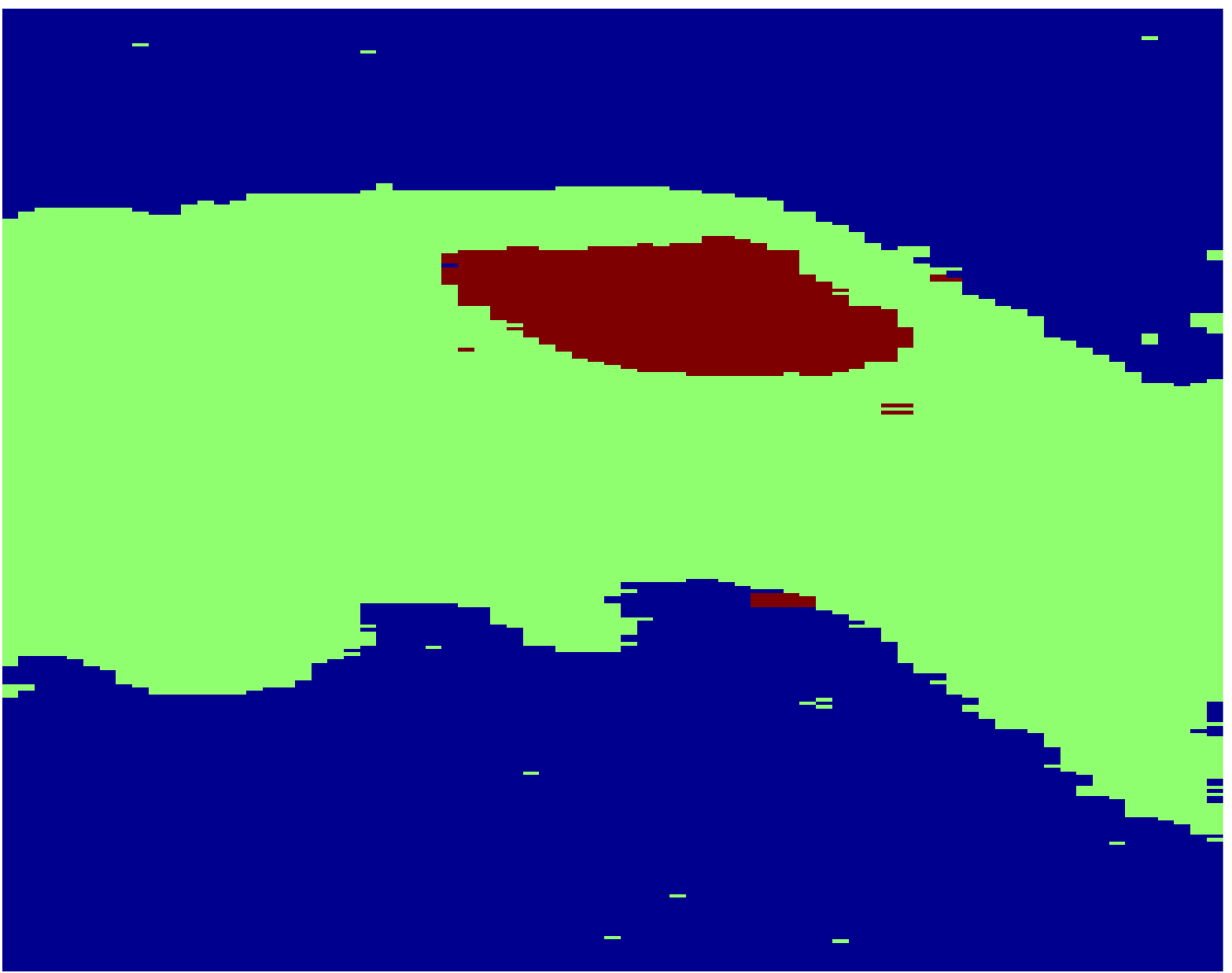} \label{fig:zmap_Jointmcmc2}} 
\caption{Group 3: (a) Ground truth of the TRF; (b) Ground truth for label map; (c) Observed B-mode image; 
(d)-(h) Estimated TRFs in B-mode form obtained with methods $\ell_2$, $\ell_1$, $\rm{Deconv_{EM}}$, $\rm{Deconv_{MCMC}}$ and the proposed $\rm{Joint_{MCMC}}$; (i) Estimated label map obtained with the proposed method (regularization parameters for the $\ell_2$ and $\ell_1$ methods set to $0.1$ and $1$).}
\end{figure*}

\paragraph{Group 3}
The third simulated image was obtained by using a clean TRF $\bfx$ of size $275 \times 75$ (see Fig. \ref{fig:refl2}) blurred by the same simulated PSF and contaminated by an AWGN such that BSNR $=30$ dB. A more realistic geometry of the simulated tissues was considered, inspired by one of the \textit{in vivo} results provided in the next section (see Fig. \ref{fig:peau2_bmodey}). Three different structures were generated mimicking the skin, the tumor and the surrounding tissue (green, red and blue regions in Fig. \ref{fig:mask2}). 
The pixels in the different regions are distributed according to GGDs with different parameters: $(\xi,\gamma) = (0.5,1)$ for the blue region, $(\xi,\gamma) =(1,30)$ for the green region and $(\xi,\gamma) =(1.8,2)$ for the red region. 
\rev{ Figs. \ref{fig:bmode_Deconvl22}-\ref{fig:bmode_Jointmcmc2} show the estimated TRFs obtained with the methods $\ell_2$, $\ell_1$, $\rm{Deconv_{EM}}$, $\rm{Deconv_{MCMC}}$ and $\rm{Joint_{MCMC}}$. The estimated label map obtained with the method $\rm{Joint_{MCMC}}$ is also shown in Fig. \ref{fig:zmap_Jointmcmc2}.} Visually, we remark that all the three methods provide images with better object boundary definition (better spatial resolution) than the observed B-mode images. The quantitative results reported in Table \ref{tab:1circle_metrics} confirm that given the same conditions (knowledge of the true label map), our approach ``$\rm{Deconv_{MCMC}}$'' is more accurate than the existing ``$\rm{Deconv_{EM}}$''. 
Moreover, we can note that the proposed technique ``$\rm{Joint_{MCMC}}$'' is able to estimate the label map with a precision of more than $98\%$ and with a small quality loss for the estimated TRF.

\begin{table}
\begin{center}
\caption{Deconvolution Quality Assessment for \rev{Simulated} data}
\label{tab:1circle_metrics}
\begin{tabular}{|c|c|c|c|c|c|c|}
\hline
 G$^{\flat}$&Method & ISNR  & NRMSE &PSNR & MSSIM & OA \\
  &       & (dB)  &       &(dB) &  &   \\
\hline
\hline
\multirow{5}{*}{1}
&$\ell_2$   &  12.83   & 0.52  & 33.19 &  0.98 & N/A\\
&$\ell_1$   &  12.83   & 0.52  & 33.19 &  0.98 & N/A\\
&$\rm{Deconv_{EM}}$   &  13.04   & 0.46  & 33.74 &  0.98 & N/A\\
&$\rm{Deconv_{MCMC}}$ &  16.21 & 0.35 &  36.57 & 0.99 & N/A\\
&$\rm{Joint_{MCMC}}$&   16.01 &  0.36 &  36.37 &  0.99 &0.99\\
\hline\hline
\multirow{5}{*}{2}
&$\ell_2$   &  10.63   & 0.69  & 21.02 &  0.61 & N/A\\
&$\ell_1$   &  12.75   & 0.54  & 23.30 &  0.79 & N/A\\
&$\rm{Deconv_{EM}}$   &  14.31 & 0.45  & 24.70 &  0.82 & N/A  \\
&$\rm{Deconv_{MCMC}}$ &  15.09 & 0.41 &  25.39 &  0.88 & N/A \\
&$\rm{Joint_{MCMC}}$  &  15.00 & 0.42 &  25.26 &  0.88 & 0.99 \\
\hline\hline
\multirow{5}{*}{3}
&$\ell_2$   &  9.96   & 0.70  & 21.92 &  0.64 & N/A\\
&$\ell_1$   &  11.49  & 0.59  & 23.45 &  0.76 & N/A\\
&$\rm{Deconv_{EM}}$   &  12.21 & 0.54 & 24.16 &  0.78 &N/A  \\
&$\rm{Deconv_{MCMC}}$ &  12.40 & 0.52 & 24.40 &  0.80 & N/A \\
&$\rm{Joint_{MCMC}}$&   12.38  & 0.53 & 24.37 &  0.79 & 0.98 \\
\hline
\end{tabular}
\end{center}
{\footnotesize \rule{0in}{1.2em}$^\flat$\scriptsize Represents Group.}
\end{table} 

\begin{figure}[!h]
\centering
\subfigure{\includegraphics[width=0.3\linewidth]{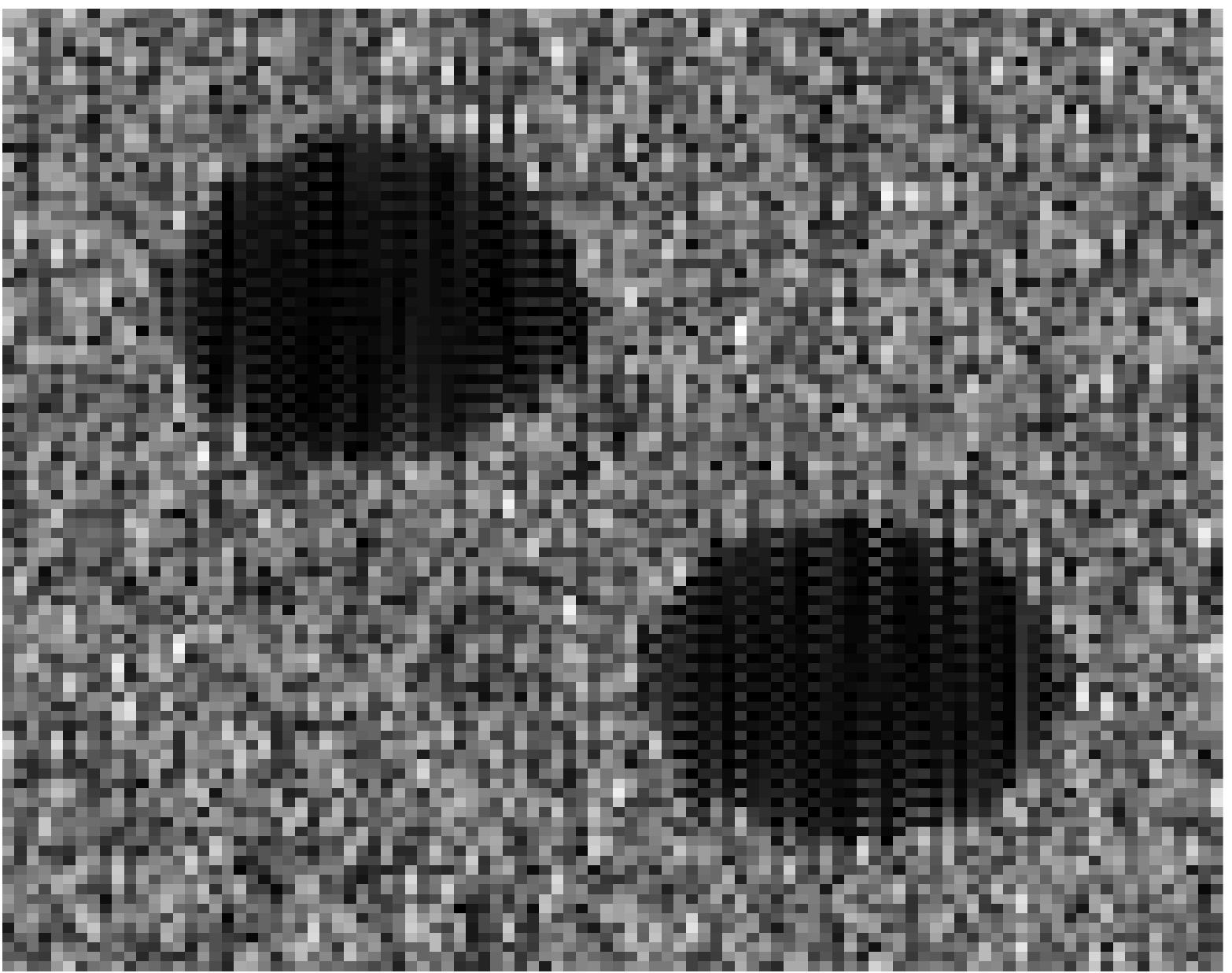} \label{fig:K_bmode}}
\subfigure{\includegraphics[width=0.3\linewidth]{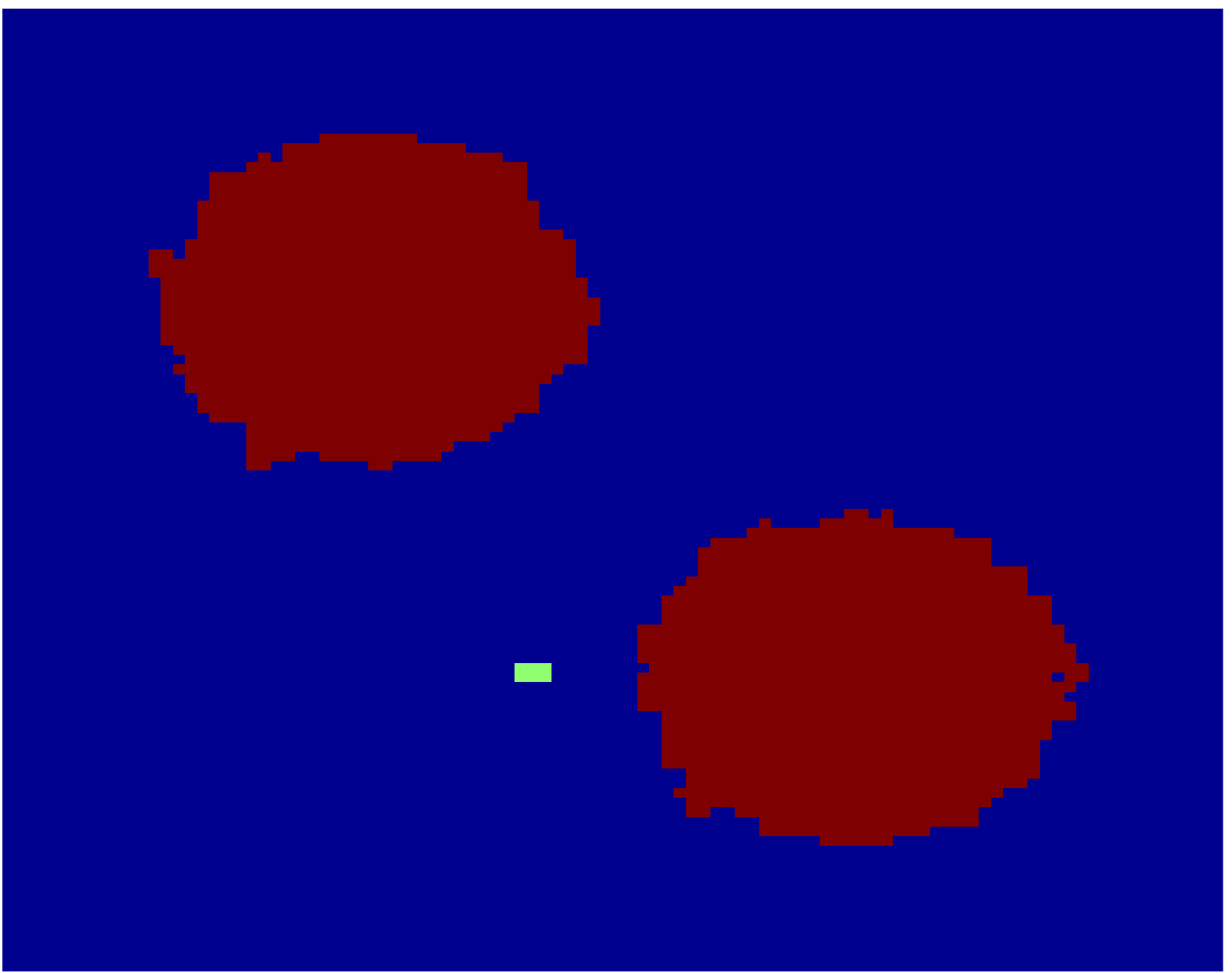} \label{fig:K_label}}
\caption{\rev{Estimated TRF (left) and label map (right) for a two-class image with $K=3$ (ISNR $= 14.46$ and OA $= 0.8$).}}
\label{fig:K_est}
\end{figure}

\paragraph{\rev{Influence of the number of classes}}
\rev{While most of the hyperparameters are automatically estimated in our Bayesian method, the number of classes $K$ has to be tuned manually. This section studies the influence of the parameter $K$ on the segmentation and deconvolution. For this purpose, we reconsider the simulated image of Group 2 by setting $K=3$, while the TRF only contains two classes of pixels. The corresponding estimated TRFs and label maps are shown in Fig. \ref{fig:K_est}. A visual inspection as well as the obtained ISNR show that the restored TRF in Fig. \ref{fig:K_est} (left) is similar to the result in Fig. \ref{fig:bmode_Jointmcmc} that was obtained by setting $K=2$. A slight degradation of the estimated label field can be observed, as highlighted by the OA that decreases from $0.99$ to $0.8$.} 

\subsubsection{Joint deconvolution and segmentation for \textit{in vivo} US images}
\begin{figure*}
\centering
\subfigure[Observation]{\includegraphics[width=0.22\linewidth]{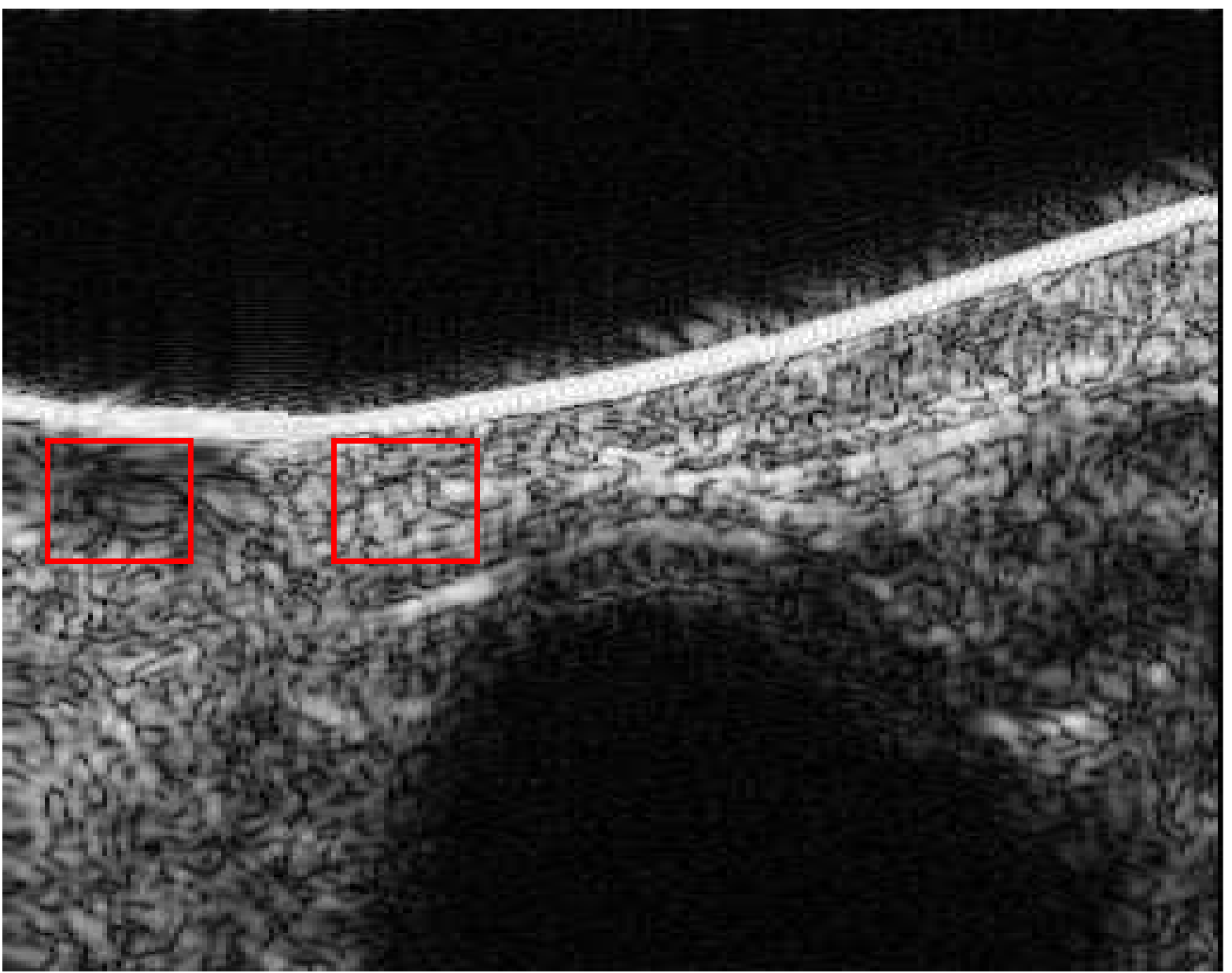}  
\label{fig:eg2_bmode_y}}
\subfigure[$\ell_2$]{\includegraphics[width=0.22\linewidth]{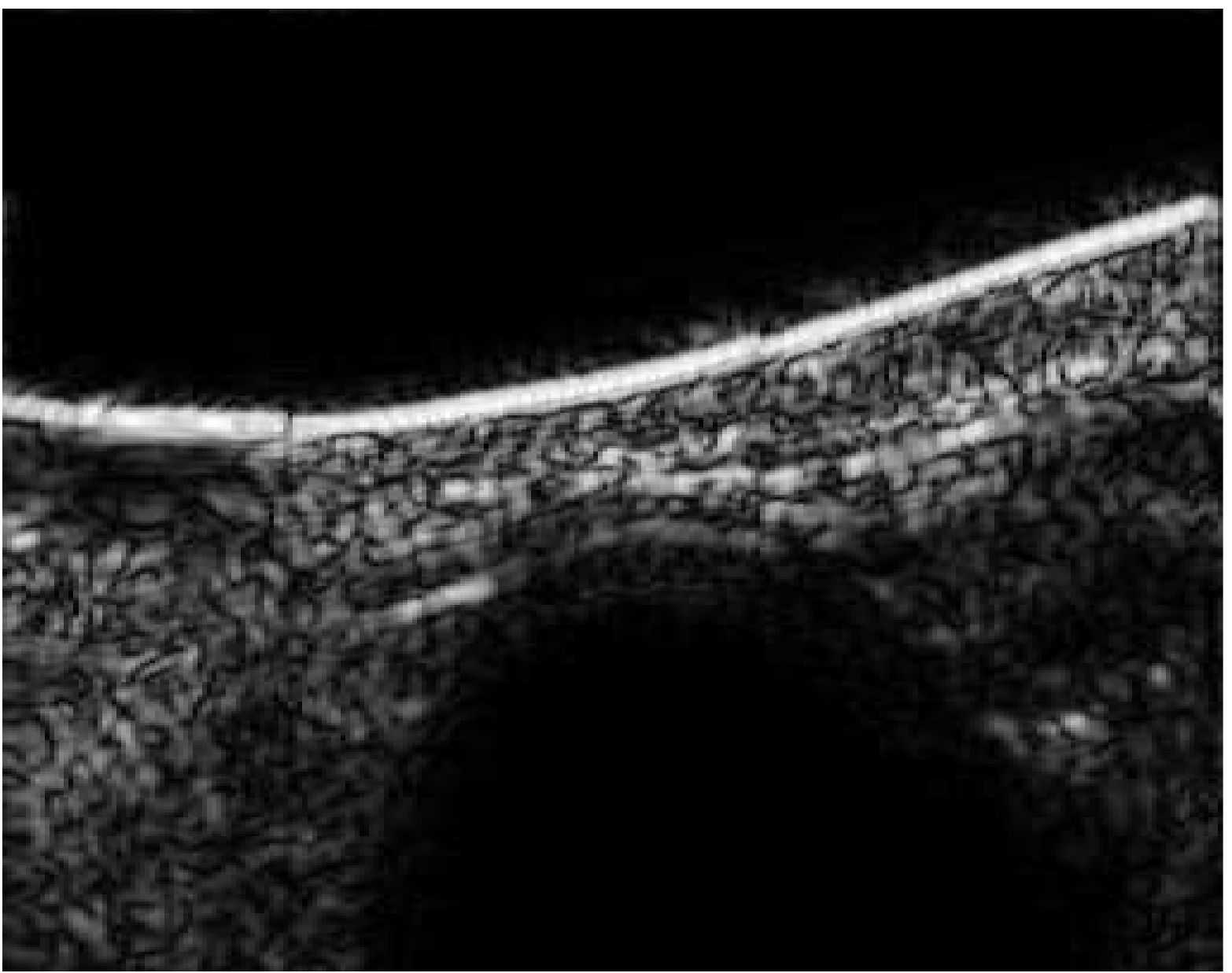}
\label{fig:eg2_bmode_new_Gauss}}
\subfigure[$\ell_1$]{\includegraphics[width=0.22\linewidth]{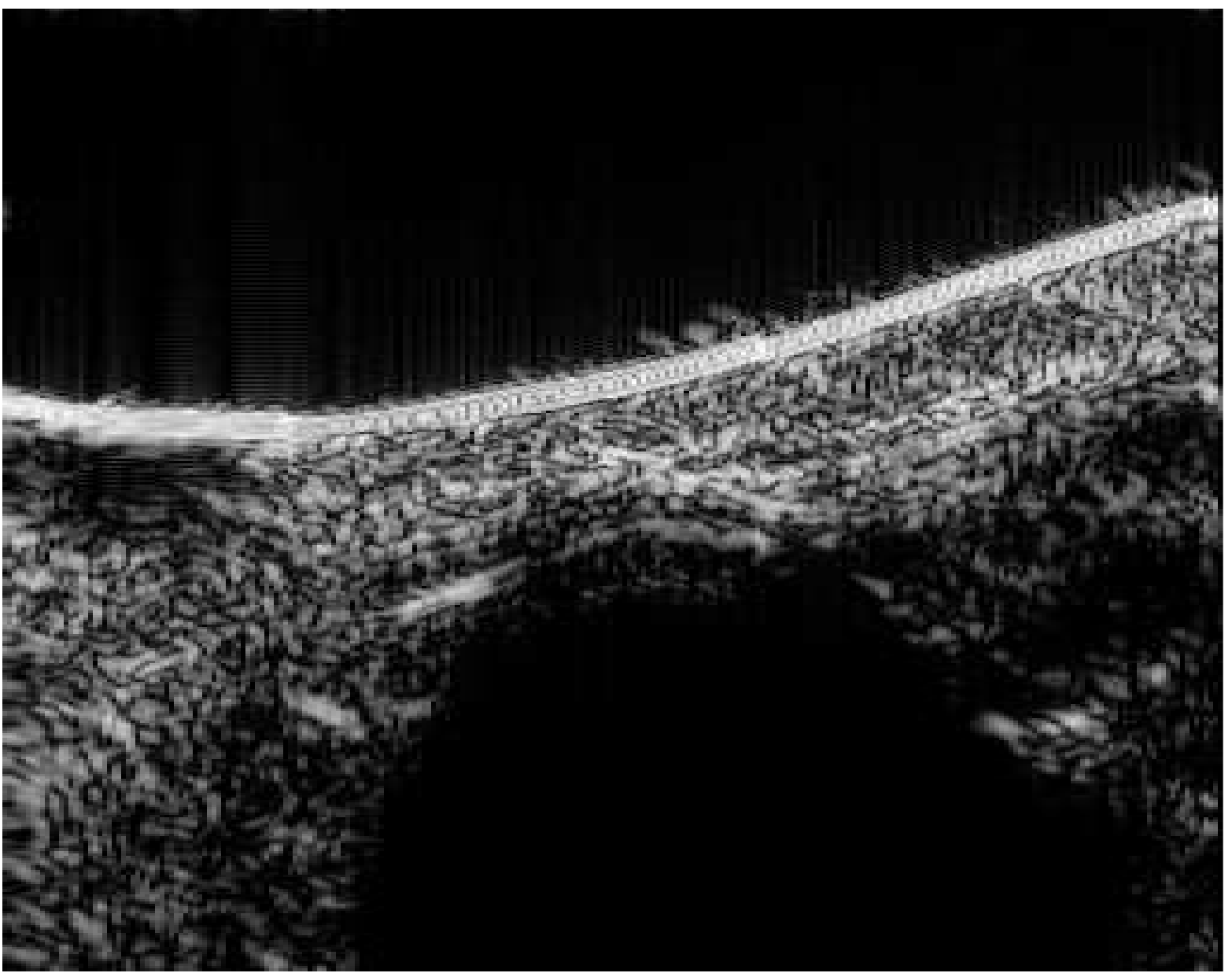}
\label{fig:eg2_bmode_new_Laplace}}
\subfigure[Proposed]{\includegraphics[width=0.22\linewidth]{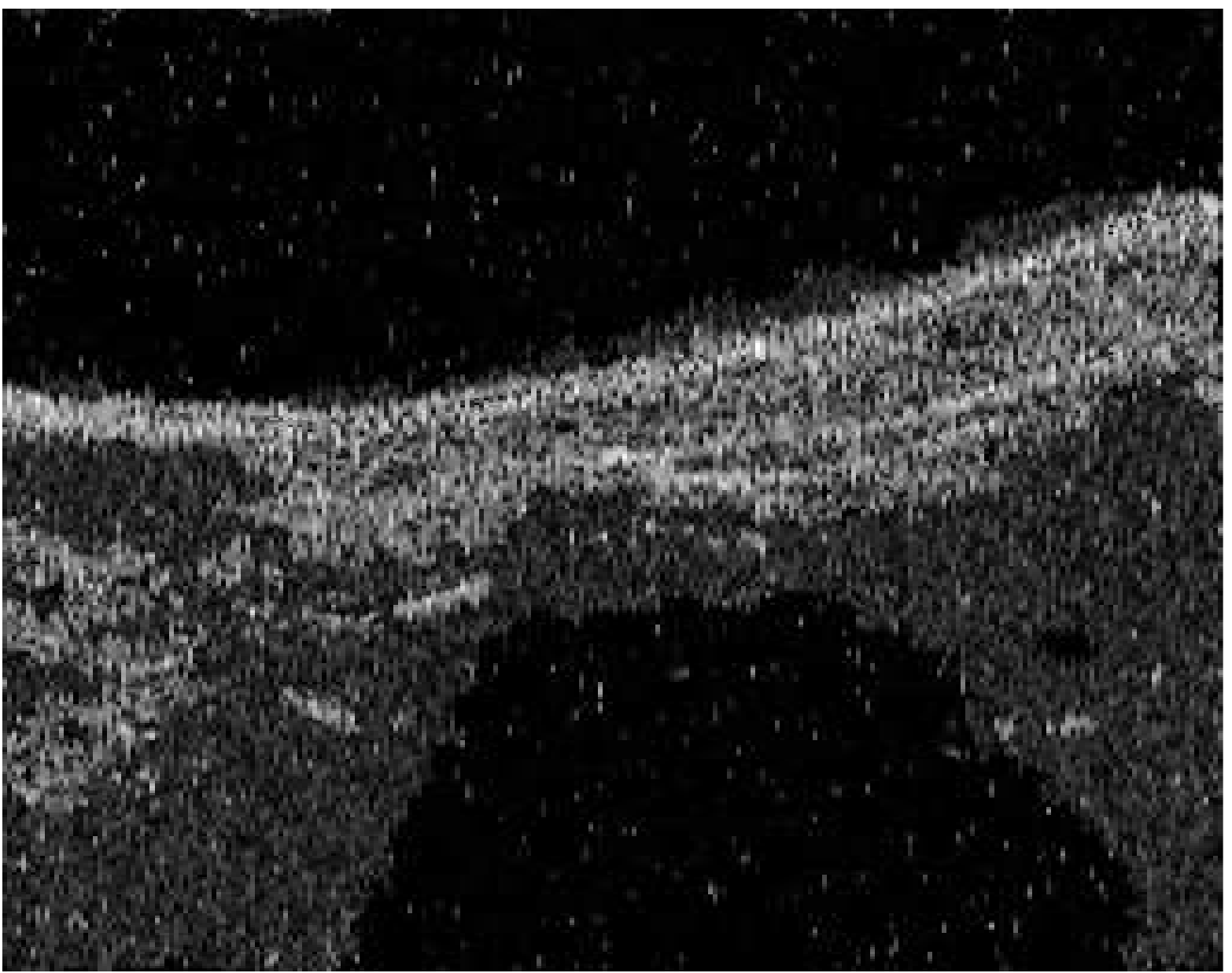}
\label{fig:eg2_bmode_new_GGD}}\\

\subfigure[Observation]{\includegraphics[width=0.22\linewidth]{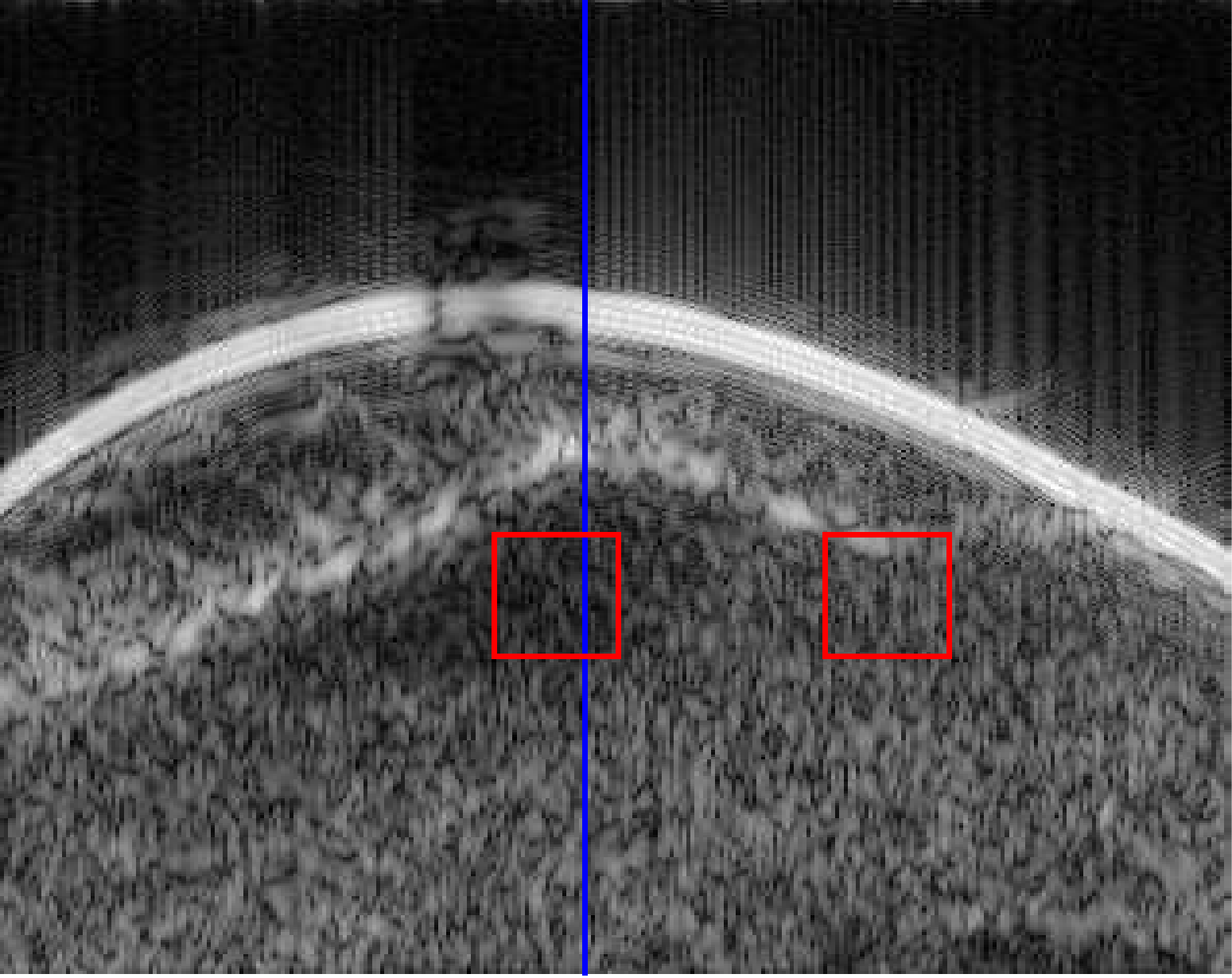} \label{fig:eg5_bmode_y}}
\subfigure[$\ell_2$]{\includegraphics[width=0.22\linewidth]{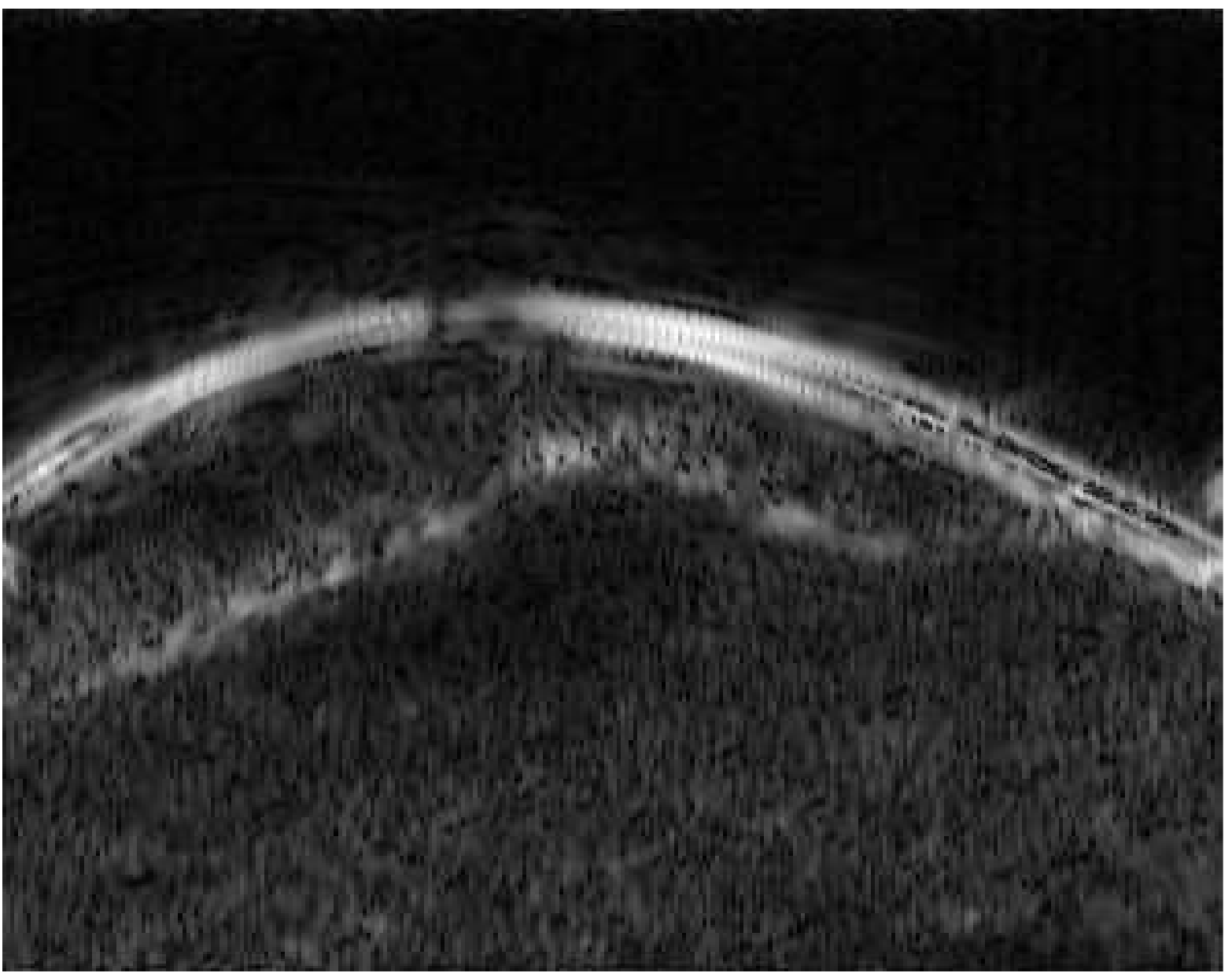}\label{fig:eg5_bmode_new_Gauss} }
\subfigure[$\ell_1$]{\includegraphics[width=0.22\linewidth]{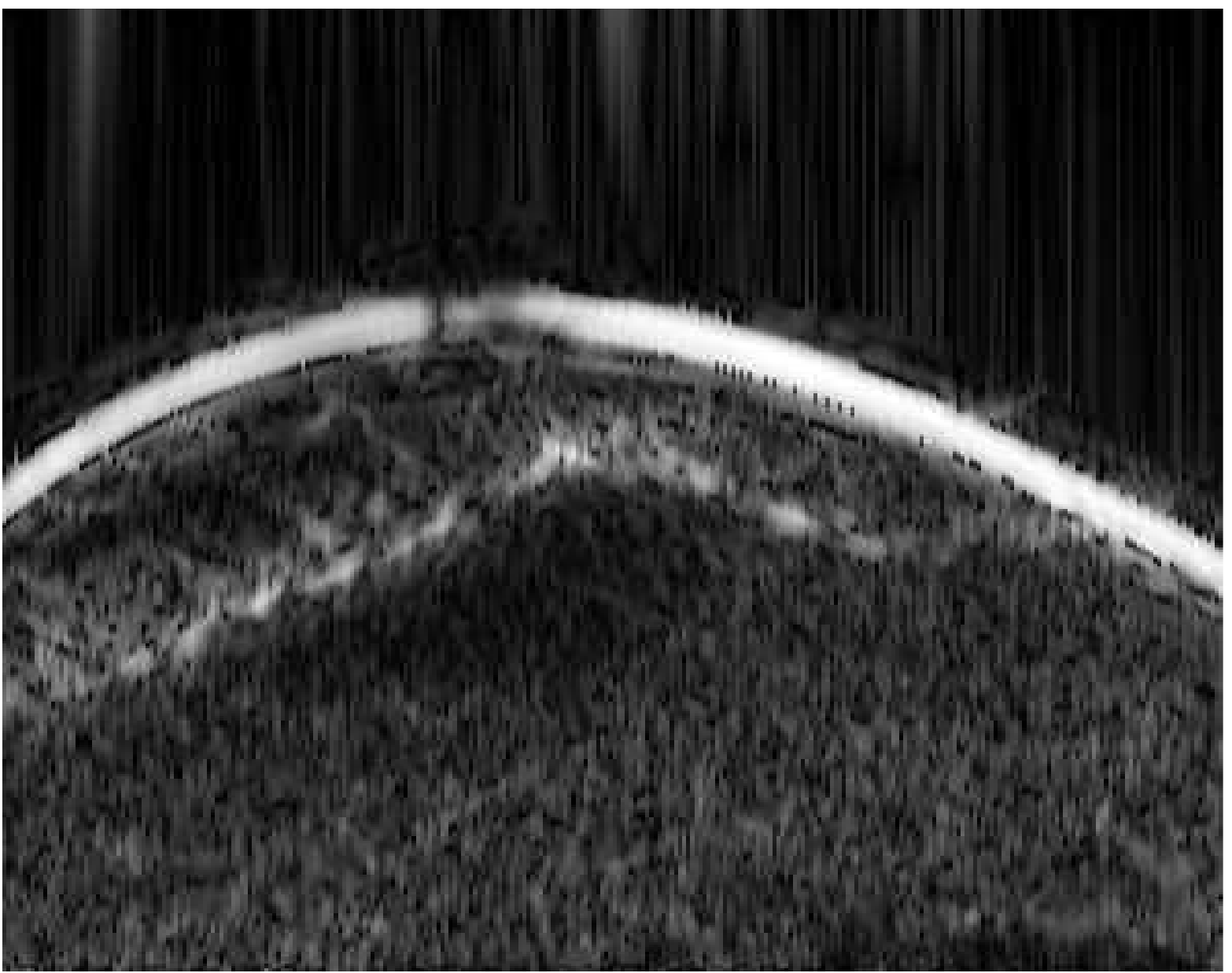}\label{fig:eg5_bmode_new_Laplace}}
\subfigure[Proposed]{\includegraphics[width=0.22\linewidth]{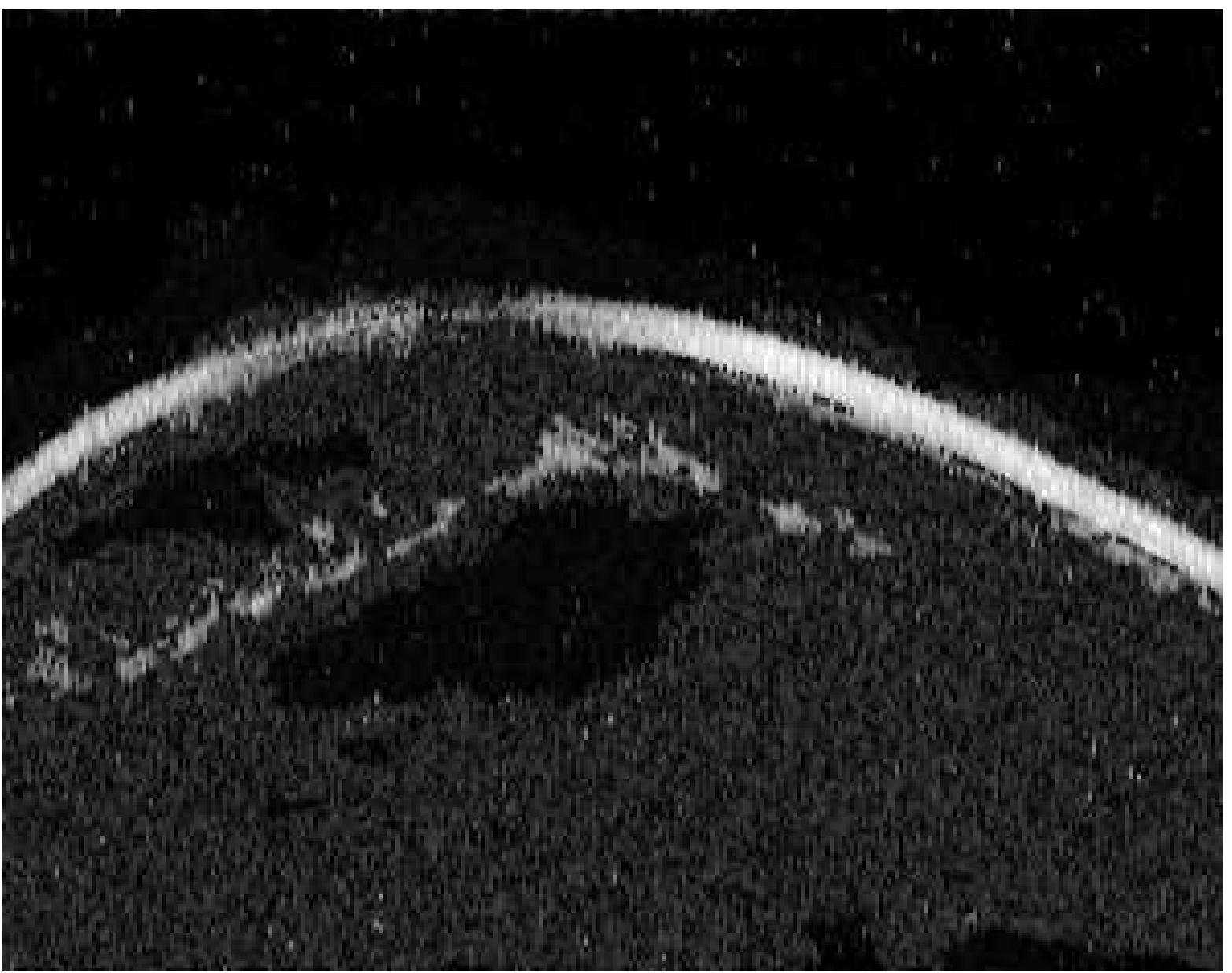}\label{fig:eg5_bmode_new_GGD}}\\

\subfigure[Observation]{\includegraphics[width=0.22\linewidth]{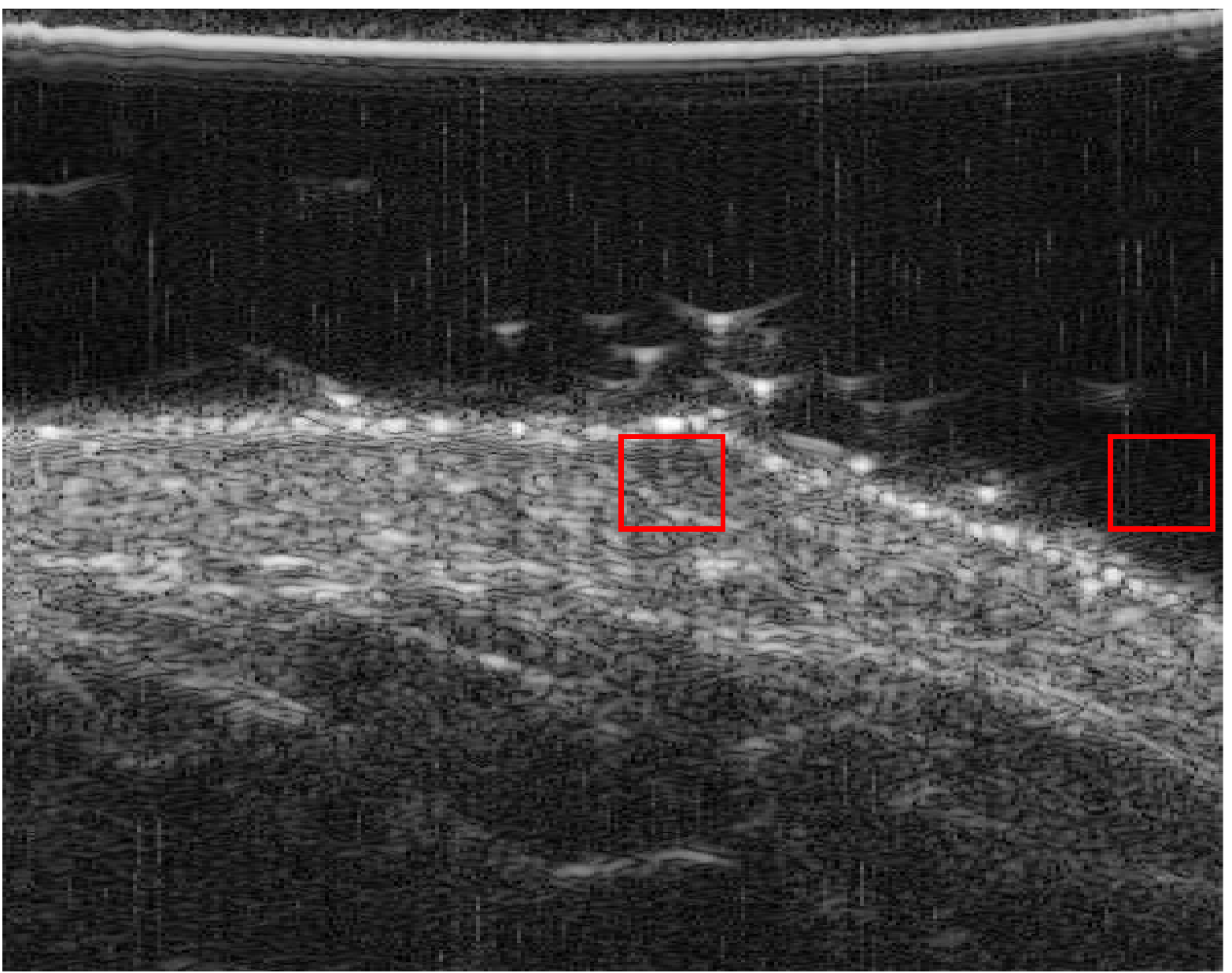} \label{fig:peau2_bmodey} }
\subfigure[$\ell_2$]{\includegraphics[width=0.22\linewidth]{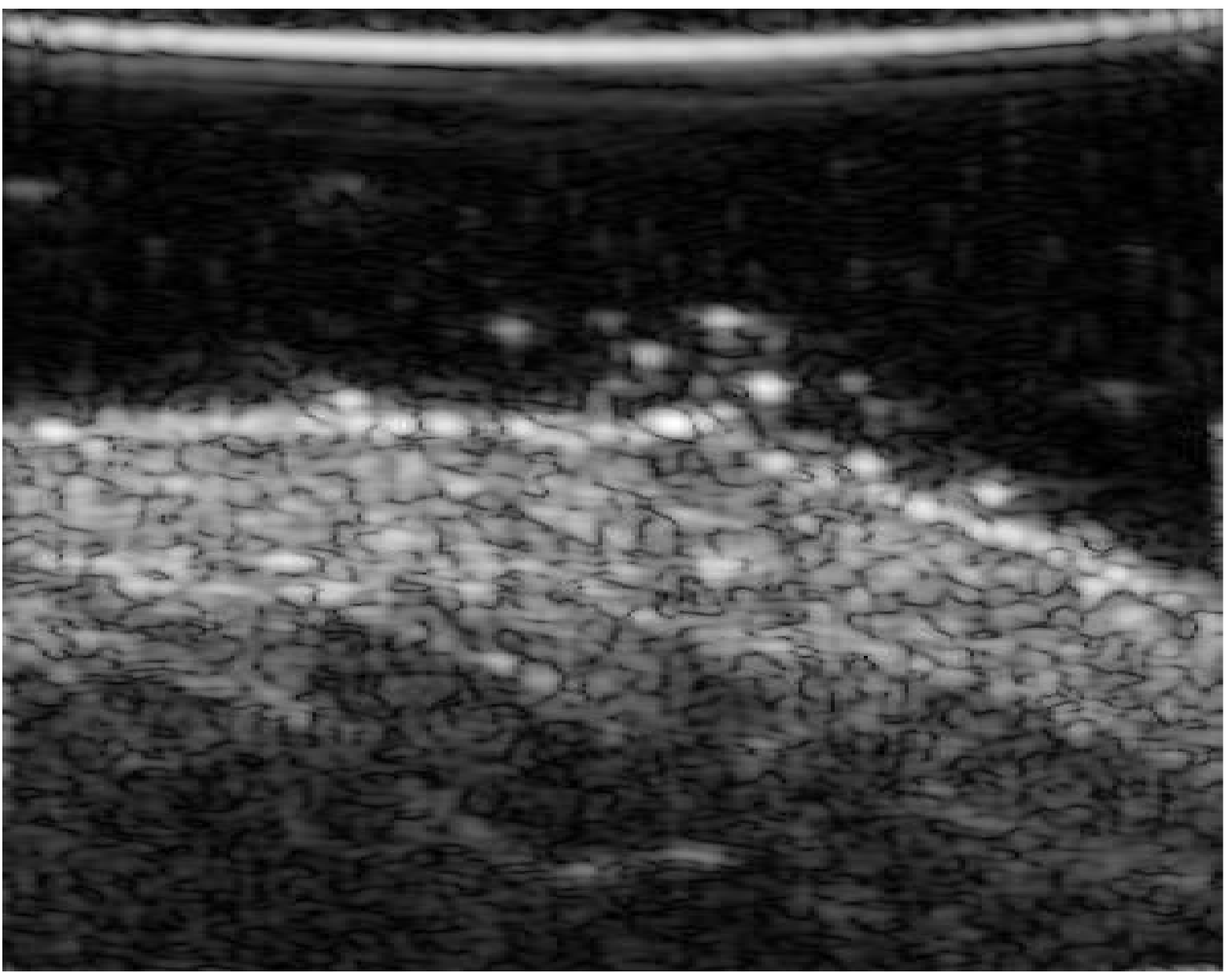}\label{fig:peau2_bmodenew_l2}} 
\subfigure[$\ell_1$]{\includegraphics[width=0.22\linewidth]{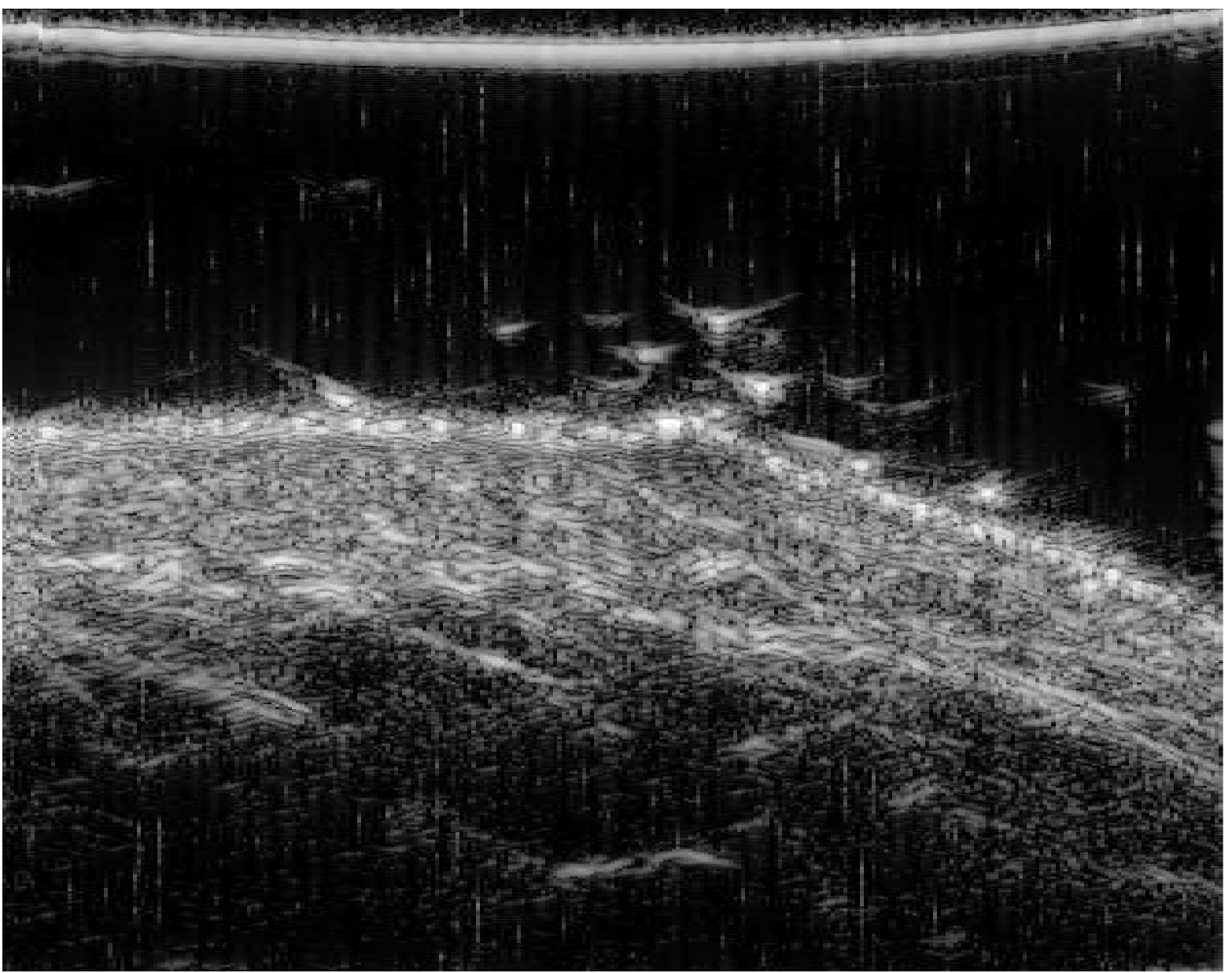}\label{fig:peau2_bmodenew_l1}} 
\subfigure[Proposed]{\includegraphics[width=0.22\linewidth]{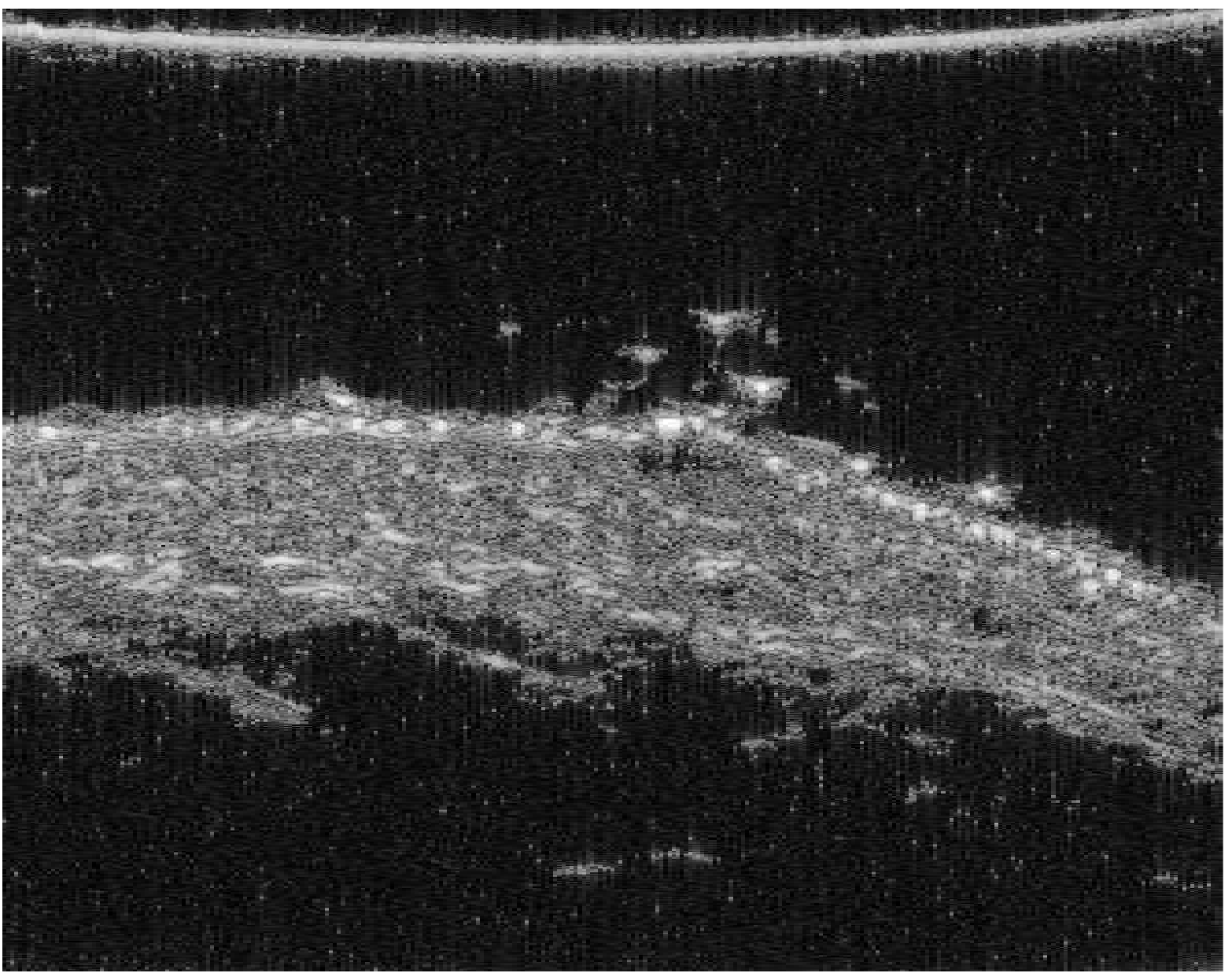}\label{fig:peau2_bmodenew_lp} }\\
\caption{\rev{From up to down: 1st row corresponds to the mouse bladder; 2nd row is for the skin melanoma; 3rd row is for the healthy skin tissue.} From left to right: Observed B-mode image, Restored B-mode images with $\ell_2$-norm, $\ell_1$-norm and the proposed method. The regions selected for computing CNR are shown in the red boxes in the observed B-mode images, i.e., (a), (e), (i).}
\label{images_TV}
\end{figure*}

Three groups of experiments have been conducted to evaluate the performance of the proposed method for \textit{in vivo} US images. The images were acquired with a 20 MHz single-element US probe. In contrast to the simulation scenarios studied previously, the PSF and the TRF were not available for \textit{in vivo} experiments. For this reason, the PSF was estimated from the RF image 
using the method of \cite{Michailovich2005}. The regions selected for the computation of CNR are shown in the red rectangles in Figs. \ref{fig:eg2_bmode_y}, \ref{fig:eg5_bmode_y}, \ref{fig:peau2_bmodey}.
\rev{All the estimated TRFs are shown in B-mode like form, after envelope detection and log-compression. The envelope detection is generally performed by considering the magnitude of the analytic signal in US imaging. While it is adapted to bandlimited modulated RF signals, this envelope detector may generate artifacts on TRFs. To avoid this phenomenon, we have used a different envelope detection method for the restored TRF, \textit{i.e.}, the method of \cite{Flandrin2004} based on the detection and interpolation of local maxima.}

\begin{figure*}[htbp]
\centering
\subfigure[]{\includegraphics[width=0.22\linewidth]{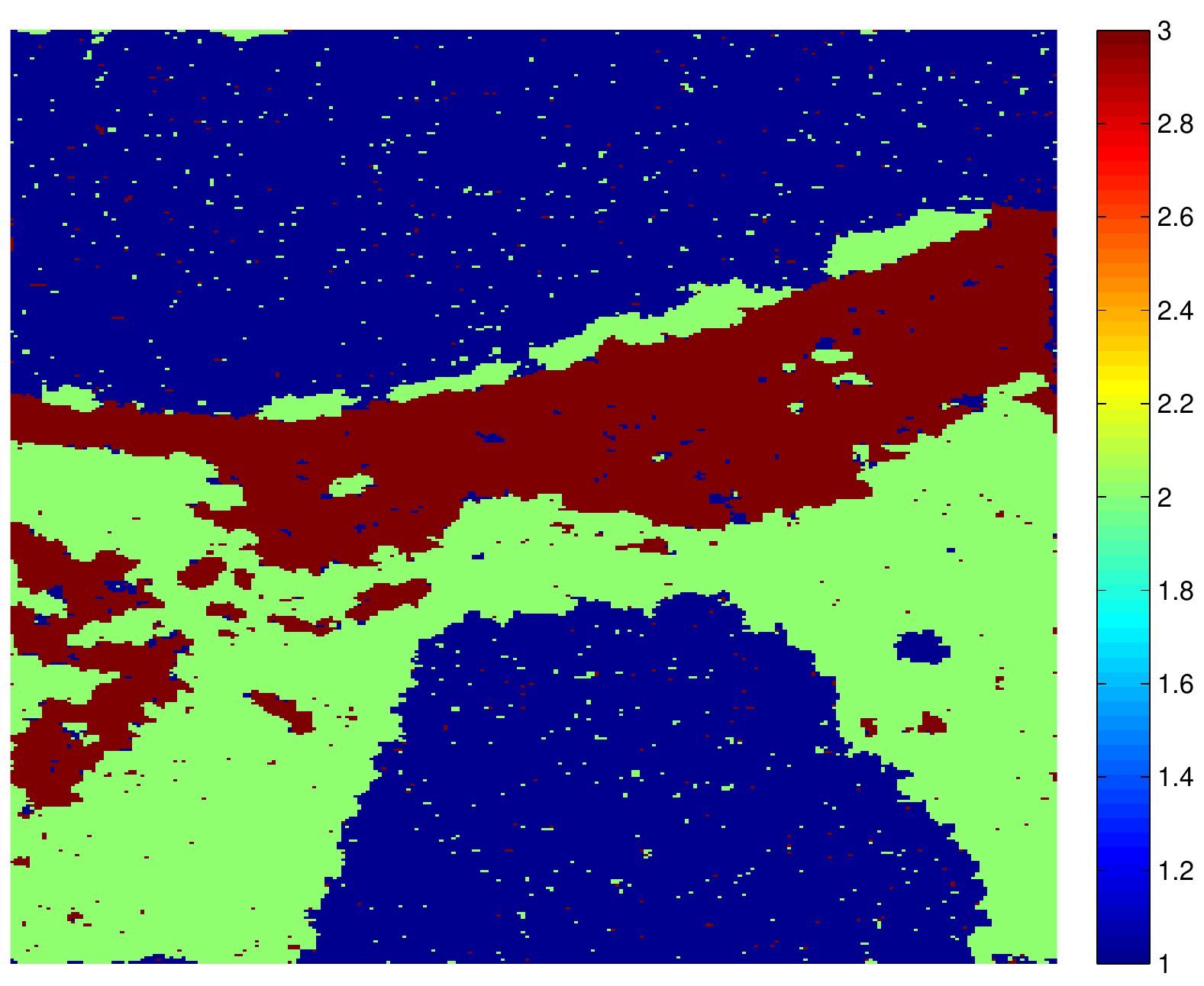} \label{fig:zmap_eg6}}
\subfigure[]{\includegraphics[width=0.22\linewidth]{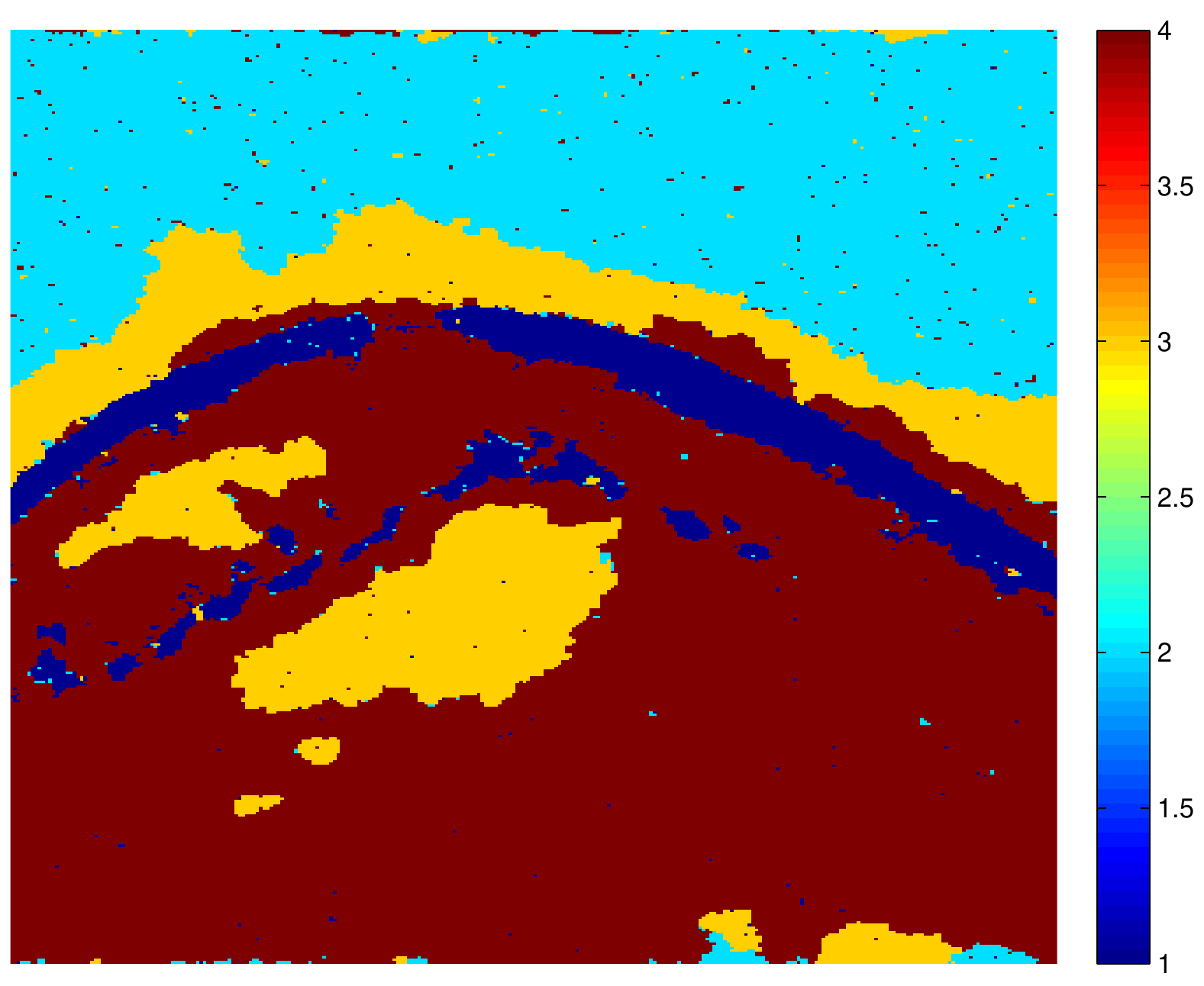} \label{fig:zmap_peau}}
\subfigure[]{\includegraphics[width=0.22\linewidth]{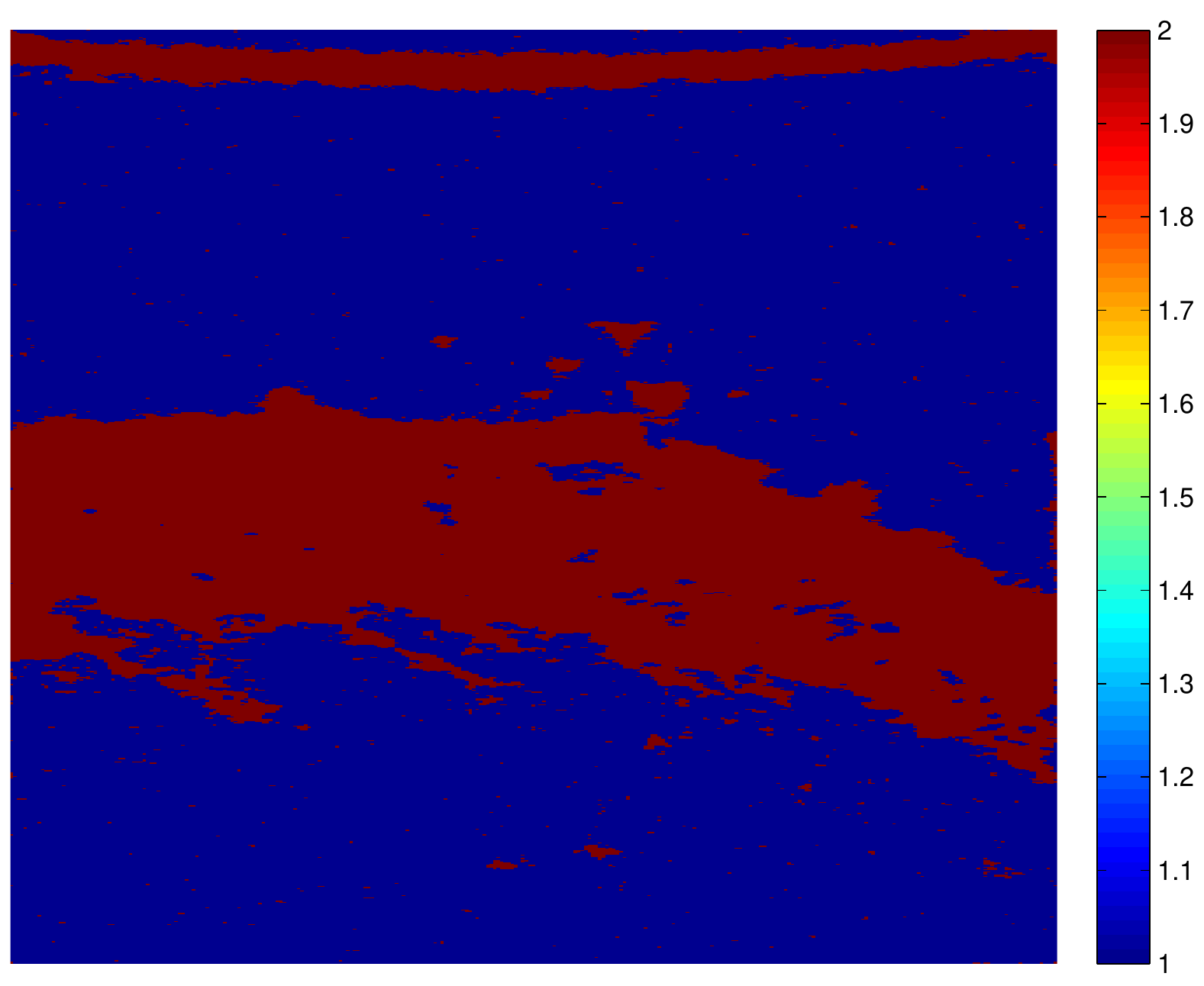}\label{fig:zmap_peau2}}
\caption{Marginal MAP estimates of labels. (a) \rev{is the label map for the mouse bladder.} The labels in red correspond to liquid regions whereas the other labels represent tissue regions with different statistical properties. (b) \rev{is the label map for the skin melanoma.} The yellow region shows the water-based gel ensuring an efficient contact between the US probe and the skin, the red pixels correspond to the tumor and the healthy skin tissues appear in blue. (c) \rev{is the label map for the healthy skin tissue.} The skin tissue appears in \rev{red}.}
\end{figure*}

\paragraph{Group 1 \rev{- Mouse bladder}} 
The observed B-mode image \rev{of size $400\times256$} is shown in Fig. \ref{fig:eg2_bmode_y} and displays a mouse bladder. The US transducer was placed into a small water container to ensure an efficient transmission of the US waves into the tissues. As there is no US scatterer in the water, the region located in the upper part of the image in Fig. \ref{fig:eg2_bmode_y} appears dark (no signal). It is also the case for the region located inside the bladder that also contains a fluid with poor reflection for the US waves. The number of homogeneous regions was set to $K=3$ in this experiment, which is sufficient to represent the anatomical structures of the image. The number of Monte Carlo iterations was fixed to 10 000 (including 5 000 burn-in iterations).  The parameters of the HMC method for the \textit{in vivo} data were adjusted to the same values as in the previous experiments. \rev{The regularization parameters for the $\ell_2$-norm and $\ell_1$-norm constraint optimization problems were set to $10$ and $54.39$ by cross-validation.} 
Figs. \ref{fig:eg2_bmode_new_Gauss}-\ref{fig:eg2_bmode_new_GGD}
display the restored TRFs obtained with the $\ell_2$, $\ell_1$ optimization algorithms and the proposed method. 
The proposed method provides good restoration results, especially with clearer boundaries. Fig. \ref{fig:zmap_eg6} shows the marginal MAP estimates of the labels, which segment the estimated image into several statistically homogeneous regions. The different anatomical structures of the image can be clearly recovered. Note that the two regions corresponding to fluids are identified with the same estimated label.

\begin{figure}
\centering
\includegraphics[scale=0.3]{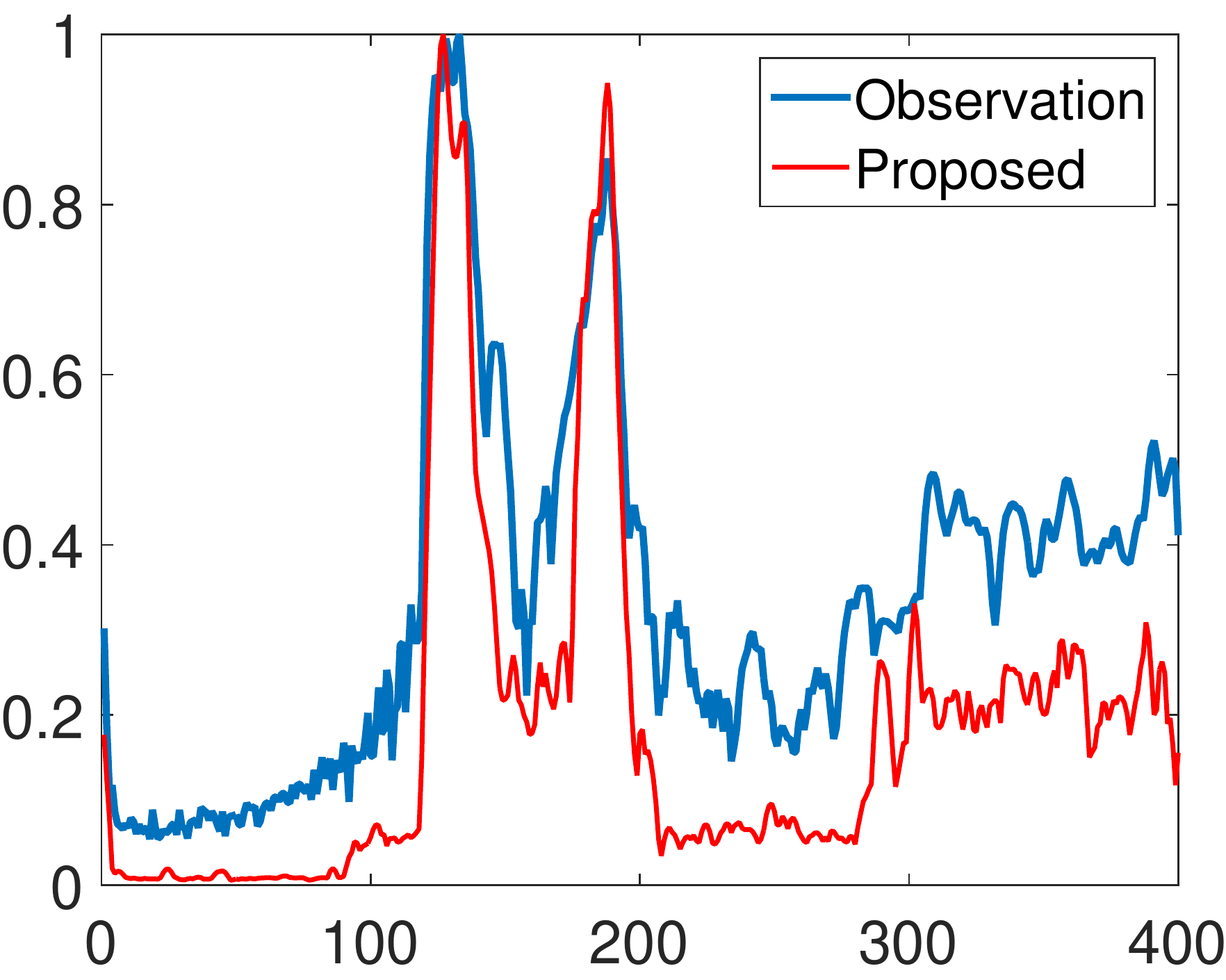}
\caption{\revAQ{Vertical profiles passing through the skin tumor, extracted from the observed and restored images of Fig. \ref{fig:eg5_bmode_y} and \ref{fig:eg5_bmode_new_GGD}.}}
\label{fig:tumor_profiles}
\end{figure}

\paragraph{Group 2 \rev{- Skin melanoma}} 
The second \textit{in vivo} image \rev{(of size $400\times 298$)} represents a skin melanoma tumor acquired in the same conditions as previously, shown in Fig. \ref{fig:eg5_bmode_y}. Water-based gel was placed between the US probe and the skin of the patient. It represents the dark regions in the upper part of the image in Fig. \ref{fig:eg5_bmode_y}. The rest of the tissues corresponds to the skin layers. The number of homogeneous regions was fixed to $K=4$. The number of Monte Carlo iterations was fixed to $20 000$ (including $10 000$ burn-in period) for this example. \rev{The regularization parameters for the $\ell_2$-norm and $\ell_1$-norm constraint optimization problems were set to $1$ and $1.2\times10^3$ by cross-validation.}
Figs. \ref{fig:eg5_bmode_new_Gauss}-\ref{fig:eg5_bmode_new_GGD} display the restored TRFs with the different methods ($\ell_2$, $\ell_1$ optimization algorithms and proposed method). Note that Fig. \ref{fig:eg5_bmode_new_GGD} shows an improved contrast between the tumor and the healthy skin tissue when compared to the \rev{observed B-mode image} in Fig. \ref{fig:eg5_bmode_y}. \revAQ{The tumor boundaries are better defined on the deconvolved image with the proposed method compared to the observed B-mode image. To better visualize the improved transition between the tumor and the healthy skin tissue, we show in Fig. \ref{fig:tumor_profiles} two vertical profiles passing through the tumor, corresponding to the blue line in Fig. \ref{fig:eg5_bmode_y}, extracted from our result and observation. One can remark the sharper slopes obtained on the deconvolved image in the neighbourhood of tumor boundaries, \textit{i.e.} around positions 200 and 300.}
The marginal MAP estimates of the labels for this image are shown in Fig. \ref{fig:zmap_peau}. The four estimated labels correspond to the water-gel (light blue), the tumor (yellow) and the skin tissues (the two shades of red).

\paragraph{Group 3 \rev{- Healthy skin tissue}} 
The last \textit{in vivo} US data represents a healthy skin image shown in Fig. \ref{fig:peau2_bmodey}, \rev{which is of size $832 \times 299$.}
The number of homogeneous regions was set to $K=2$. The number of Monte Carlo iterations was fixed to $6000$ including a burn-in period of $2000$ iterations). \rev{The regularization parameters for the $\ell_2$-norm and $\ell_1$-norm constraint optimization problems were set to $10$ and $1.5\times 10^4$ by cross-validation.} 
The restored TRFs obtained with the different methods ($\ell_2$, $\ell_1$ optimization algorithms and the proposed method) are displayed in Figs. \ref{fig:peau2_bmodenew_l2}-\ref{fig:peau2_bmodenew_lp}. 
The marginal MAP estimation of the label field is shown in Fig. \ref{fig:zmap_peau2}.

In addition to the visual inspection, the deconvolution results were evaluated using the RG and CNR criteria and the CPU time, as reported in Table \ref{tab:QAR}. Despite its higher computational complexity, the visual impression and the numerical results confirm that a better contrast and more defined boundaries between the different tissues is achieved with our method. It is interesting to note that in addition to the restored image, our algorithm also provides a segmentation result. To our knowledge, there is no other existing method in US imaging able to achieve this joint segmentation and deconvolution performance. 

\begin{table*}
\caption{Deconvolution Quality for the real US data}
\label{tab:QAR}
\centering
\begin{tabular}{|c|c|c|c|c|c|c|c|c|c|}
\hline
 Group     & \multicolumn{3}{c|}{\rev{group 1 - Mouse bladder}} & \multicolumn{3}{c|}{\rev{group 2 - Skin melanoma}} & \multicolumn{3}{c|}{\rev{group 3 - Healthy skin tissue}}\\
\hline\hline
 Metrics & RG & CNR & Time (s)  & RG & CNR & Time (s)   & RG  & CNR & Time (s)\\
 \hline\hline
\rev{Observation}& - & 1.08 & -           & - & 1.17 &  -    & - & 1.30&   -   \\
\hline
$\ell_2$ & 3.82 & 1.00 & 0.006  & 3.01 & 1.09   &  0.007  & 1.07   &\textbf{3.01}&  0.007\\
\hline
$\ell_1$ & 3.29 & \textbf{1.11} & 5.07            & 4.63 & 1.19 &  3.53    & 2.09 & 2.47&   22.30   \\
 \hline
Proposed & \textbf{3.94} & 0.94 & 3904.8 & \textbf{10.01} & \textbf{1.35}  &  1303.4   & \textbf{2.59} &2.23 &  6585.8\\
\hline
\end{tabular}
\end{table*} 

\section{\revAQ{Discussions}}
\rev{The main drawback of the proposed method is its computational complexity, which limits its use in real-time applications. However, the proposed algorithm is interesting for numerous off-line applications.} For example, improving the readability of US images (e.g., spatial resolution, contrast, SNR) off-line allows the clinician to better appreciate the anatomical structures, especially when very accurate measurements are required (e.g., for cancer detection) or when very small structures must be identified (e.g., vessel walls). Computer-aided detection, often performed off-line and based on a quantitative analysis of the images, could also take advantage from the deconvolved images provided by our approach, \rev{see, e.g., \cite{MartinoAlessandrini2011}}.
\rev{Finally, we would like to emphasize that the main objective of this paper is to validate the proposed joint segmentation and detection strategy on simulated images with a controlled ground truth and to show its applicability to clinical examples. Performing a deeper clinical evaluation is obviously an interesting and essential perspective that will be conducted in future work.}
\section{\revAQ{Conclusions}}
This paper proposed a Bayesian method for the joint deconvolution and segmentation of ultrasound images. 
This method assumed that the ultrasound image can be divided into regions with statistical homogeneous properties. 
\rev{Based on this assumption, a Potts model was introduced for the image labels. Independent generalized Gaussian priors were also assigned to the tissue reflectivity functions of each homogeneous region of the image.} According to the author's knowledge, it is the first time a joint segmentation and deconvolution method is proposed for ultrasound images. 
The proposed method showed very interesting restoration results when compared to more classical optimization methods based on $\ell_2$-norm or $\ell_1$-norm regularizations. 

Future work includes the estimation of the point spread function within the Bayesian algorithm, resulting into a blind segmentation and deconvolution approach. The spatially varying nature of the PSF could also be considered with more sophisticated block-wise techniques ensuring the continuity and regularity of the estimated tissue reflectivities. \rev{The automatic estimation of the number of classes, which is manually tuned in this work, is also an interesting perspective that could be addressed using a Bayesian non-parametric approach}. Finally, combining our MCMC approach with deterministic optimization methods (such as the PMALA approach \cite{Pereyra2014,Schreck2013,Pereyra2016Survey}), \rev{exploring parallel techniques such as \cite{Gonzalez2011} and applying the algorithm to demodulated I/Q data} are interesting research areas, which should allow the computational cost of our algorithm to be reduced.

\section*{Acknowledgements}
The authors would like to thank Dr. Martino Alessandrini for sharing the code of the EM algorithm used for comparison in this paper. 
The authors would also like to express their gratitude to Nicolas Meyer from the CHU of Toulouse and to Hadj
Batatia from the university of Toulouse for providing us some of the \textit{in vivo} US images used in this paper. Part of this work has been supported by the Chinese Scholarship Council and the thematic trimester on image processing of the CIMI Labex, Toulouse, France, under grant ANR-11-LABX-0040-CIMI within the program ANR-11-IDEX-0002-02.

\vspace{-0.4cm}
\appendices
\rev{\section{Deviations of the conditional distributions of the noise variance and scale parameters}
\label{app:IG}
\paragraph{Inverse gamma distribution}
A univariate inverse gamma distribution with shape parameter $\alpha$ and scale parameter $\beta$ denoted as $\mathcal{IG} (\alpha, \beta)$ has the following pdf
\begin{equation}
p(x)= \frac{\beta^{\alpha}}{\Gamma(\alpha)} x^{-\alpha-1}\exp\left( -\frac{\beta}{x} \right), \quad x \in \mathbb{R^+}.
\end{equation}
The conditional distribution of the noise variance and of the GGD scale parameters of the joint posterior distribution, i.e., \eqref{eq:post_noise} and \eqref{eq:scale_conditional} are inverse gamma distributions that are derived hereinafter.
\paragraph{Conditional distribution of the noise variance}
\begin{align*}
&p(\sigma_n^2|\bfy,\bfx,\bsxi,\bsgamma,\bfz)\\
&\varpropto p(\bfy|\bfx,\sigma_n^2,\bsxi,\bsgamma,\bfz) p(\sigma_n^2) \\
&\varpropto \frac{1}{(2\pi\sigma_{n}^{2})^{\frac{N}{2}}}\exp\left(-\frac{\|\bfy-\bfH\bfx\|_2^2}{2\sigma_n^2}\right)  \times
 \frac{\nu^{\alpha}\exp\left(-\nu/\sigma_n^2\right)}{\Gamma(\alpha) (\sigma_n^2)^{\alpha+1}} \\
 &\varpropto (\sigma_n^2)^{-\alpha-N/2-1}
\times\exp\left[ -\frac{1}{\sigma_n^2}\left(\nu+\frac{1}{2}\|\bfy-\bfH\bfx\|^2_2 \right)\right].
\end{align*}
We can recognize the following inverse gamma distribution
$$
\mathcal{IG}\left(\alpha+N/2,\theta+\frac{1}{2}\|\bfy-\bfH\bfx\|^2_2\right).
$$
\paragraph{Conditional distribution of the scale parameters}
\begin{align*}
p(\gamma_k |\bfx,\bsxi,\bfz,\bsgamma_{-k}) 
&\varpropto p(\bfx_k| \xi_k,\gamma_k, \bfz_k) p(\gamma_k)  \\
&\varpropto a_k^{N_k}\exp\left(-\frac{\|\bfx_k\|_{\xi_k}^{\xi_k}}{\gamma_k} \right)\frac{1}{\gamma_k} \mathcal{I}_{\mathbb{R+}}(\gamma_k) \\
&\varpropto \gamma_k^{-N_k/\xi_k-1} \exp \left(-\frac{\|\bfx_k\|_{\xi_k}^{\xi_k}}{\gamma_k}\right).
\end{align*}
We can recognize the following inverse gamma distribution
$$
\mathcal{IG} \left(\frac{N_k}{\xi_k},\|\bfx_k\|_{\xi_k}^{\xi_k}\right).
$$
}

\section{\rev{Sampling the shape parameters with an RWMH Algorithm}}
\label{app:RWMH}
\rev{In order to sample the shape parameter $\xi_k$ following \eqref{eq:shape_conditional}, we generate a candidate using a proposal and accept or reject this candidate with an appropriate acceptance ratio. The proposal used in this paper is a truncated Gaussian distribution whose mean is $\xi_k^{(t)}$ (the value of the parameter generated at the previous iteration) and whose variance $\delta$ is adjusted in order to obtain a suitable average acceptance ratio, i.e.,} 
\begin{equation}
\xi_{k}^{*} \sim \mathcal{N}(\xi_{k}^{(t)}, \delta ) \mathcal{I}_{(0,3)}(\xi_{k}^{*}).
\end{equation}
This candidate is then accepted or rejected according to the following ratio
\begin{equation}
\rho=\min \left\lbrace \frac{p(\xi_k^{\ast}|\bfx,\bsgamma,\bfz,\bsxi_{-k})}{p(\xi_k^{t}|\bfx,\bsgamma,\bfz,\bsxi_{-k})}, 1 \right\rbrace .
\label{ratio1}
 \end{equation}
We propose to adjust the stepsize $\delta$ every 100 iterations to achieve a reasonable acceptance rate ($30\%-90\%$) \cite{Pereyra2014}. Specifically, if the acceptance ratio during the previous 100 iterations is larger than $90\%$ (respectively smaller than $30\%$), than the variance $\delta$ is decreased (respectively increased) of $20\%$ compared to its previous value. 
Note that to ensure the homogeneity of the Markov chain after the burn-in period, this tuning procedure is only executed during the burn-in period. The stepsize is then fixed during the following iterations.

The algorithm used to sample $\xi_k$ is finally divided into three procedures that are summarized in  Algo. \ref{algrithm2}.
\begin{algorithm}
\label{algrithm2}
\tcc{Initialization}\
Choose an initial value $\xi_0$\;
\tcc{Candidate Generation}\ 
\For{$t=1:N_{MC}$}
{$\xi_k^{\ast}\sim\mathcal{N}(\xi_k^{(t)},\delta)\mathcal{I}_{(0,3)}(\xi_{k}^{*})$\;
\tcc{Accept/Reject Procedure}\ 
\eIf {$rand\leqslant\rho$}
{$\xi_k^{(t+1)}=\xi_k^{\ast}$\; }
{$\xi_k^{(t+1)}=\xi_k^{(t)}$\;
}
Adjust $\delta$ in order to obtain a suitable acceptance rate.
}
\caption{\zhao{Adjusted RWMH Algorithm}}
\end{algorithm} 

\section{\rev{Sampling the TRF using an HMC Algorithm}}
\label{app:HMC}
\subsection{HMC Algorithm}
The main idea of the HMC algorithm is to introduce a vector of momentum variables $\bfp \in
\bbR^N$ that is independent of $\bfx$ and to sample the pair $(\bfx,\bfp)$ instead of just
sampling $\bfx$.
The conditional distribution of $(\bfx,\bfp)$ can be written
$$
p(\bfx,\bfp|\bfy,\sigma_n^2,\bsxi,\bsgamma,\bfz)=p(\bfx|\bfy,\sigma_n^2,\bsxi,\bsgamma,\bfz)p(\bfp).
$$
The Hamiltonian of the system is defined as
\begin{equation*}
H(\bfx,\bfp)\triangleq -\log p(\bfx,\bfp|\bfy,\sigma_n^2,\bsxi,\bsgamma,\bfz) = U(\bfx)+V(\bfp)
\end{equation*}
where $V(\bfp)$ and $U(\bfx)$ are the kinetic and
potential energies of the Hamiltonian system. They are defined as
\begin{align*}
V(\bfp)=\frac{1}{2}\bfp^{T}\bfp \; \; \textrm{and} \; \;
U(\bfx)=-\log[p(\bfx|\bfy,\sigma_n^2,\bsxi,\bsgamma,\bfz)].
\end{align*}

\noindent At the iteration $\# t$, the HMC consists of two steps:
\begin{itemize} 
\item generate a candidate pair ($\bfp^{(\star)},\bfx^{(\star)}$) from the current state ($\bfp^{(t)}, \bfx^{(t)}$) using
a discretizing method, such as the leapfrog and Euler methods;
\item accept or reject the candidate with the probability $\rho$
\begin{equation}
\rho = \textrm{min}\{\textrm{exp}[H(\bfp
^{(t)},\bfx^{(t)})-H(\bfp^{(\star)},\bfx^{(\star)})] ,1\}.
\label{eq:accept}
\end{equation}
\end{itemize} 

\noindent In our experiments, we have considered the leapfrog discretizing method due to its better performance compared to the Euler method, also noticed in \cite{Neal2011}.
The three steps of the leapfrog method are defined as
\begin{eqnarray*}
\bfp_i(t+\epsilon/2)&=&\bfp_i(t)-\frac{\epsilon}{2}\frac{\partial U}{\partial \bfx_{i}}[\bfx(t)]\\
\bfx_i(t+\epsilon)&=&\bfx_i(t)+\epsilon\bfp_i(t+\epsilon/2)\\
\bfp_i(t+\epsilon)&=&\bfp_i(t+\epsilon/2)-\frac{\epsilon}{2}
\frac{\partial U} {\partial \bfx_{i} } [\bfx(t+\epsilon)]
\end{eqnarray*}
where $\epsilon$ is a so-called stepsize and $L$ is the number of leapfrog iterations. 
We should note that $U(\bfx)$ is not differentiable when $\xi_k \leqslant 1$. To deal with this problem, a smoothing approximation has been considered, i.e., $|\cdot| \approx \sqrt{\cdot^2 + \varepsilon}$, with $\varepsilon \ll 1$.
The algorithm based on the leapfrog discretization and this approximation is summarized in Algo. \ref{hmc}.
Compared to other MCMC algorithms, the HMC method has the noticeable advantage to generate efficiently a candidate $\bfx$
even in the case of a high dimensional and complicated distribution.
\begin{algorithm}
\tcc{Initialization}\
$\bfx^{(0)}=\bfy$\;

\For{$t=1:N_{MC}$}{
\tcc{Candidate generation}\
$\bfp^{(t,0)} \sim N(\mathbf{0},\bfI^{N\times N})$\;
\tcc{Leapfrog Method}\
\For{$i=1:L$}
{Set $\bfp^{(t,i)}=\bfp^{(t,i)}-\frac{\epsilon}{2} \frac{\partial U}{\partial \bfx^{(t,i)}} \bfx^{(t,i)}$\;
Set $\bfx^{(t,i)}=\bfx^{(t,i)}+\epsilon \bfp^{(t,i)}$\;
Set $\bfp^{(t,i)}=\bfp^{(t,i)}-\frac{\epsilon}{2} \frac{\partial U}{\partial\bfx}\bfx^{(t,i)}$\;}
$\bfp^{(*)} = \bfp^{(t,L)}$;\\
$\bfx^{(*)} = \bfx^{(t,L)}$;\\
\tcc{Accept/Reject Procedure}\
Compute $\rho$ with \eqref{eq:accept}\\
\eIf{$rand \leqslant \rho$}
{$\bfx^{(t+1)}=\bfx^{(*)}$\;}
{$\bfx^{(t+1)}=\bfx^{(t)}$\;}
Adjust $\epsilon$ in order to obtain a suitable acceptance rate.
}
\caption{Adjusted HMC Algorithm}
\label{hmc}
\end{algorithm}

\subsection{Tuning the parameters $\epsilon$ and $L$}
The performance of the HMC algorithm mainly depends on the values of the parameters $\epsilon$ (stepsize) and $L$ (number of leapfrog steps). Fortunately, these two parameters can be tuned independently in most applications \cite{Neal2011}. It is recommended to select a random number of leapfrog steps $L$ to avoid possible periodic trajectories \cite{Neal2011}. In our algorithm, $L$ is sampled uniformly in the interval $[50,70]$. 
The leapfrog stepsize $\epsilon$ has been adjusted in order to ensure a reasonable average acceptance rate any 100 iterations. 
Specifically, when the acceptance rate is too large, $\epsilon$ should be decreased and vice versa. 
The range of the acceptance rate has been set to $30\%-90\%$ \zhao{in the burn-in period}. Note that the tuning of $\epsilon$ is just carried out during the burn-in period to ensure the Markov chain is homogeneous after the burn-in period. 
The acceptance rate generally belongs to the interval $60\% - 80\%$ when the Markov chain has converged, 
while the acceptance rate is around $25\%$ in standard MH moves for high dimensional target distributions \cite{Girolami_riemannmanifold}.

\bibliographystyle{IEEEtran}
\bibliography{strings_all_ref,Jointbiblios}

\begin{thebibliography}{10}
\providecommand{\url}[1]{#1}
\csname url@samestyle\endcsname
\providecommand{\newblock}{\relax}
\providecommand{\bibinfo}[2]{#2}
\providecommand{\BIBentrySTDinterwordspacing}{\spaceskip=0pt\relax}
\providecommand{\BIBentryALTinterwordstretchfactor}{4}
\providecommand{\BIBentryALTinterwordspacing}{\spaceskip=\fontdimen2\font plus
\BIBentryALTinterwordstretchfactor\fontdimen3\font minus
  \fontdimen4\font\relax}
\providecommand{\BIBforeignlanguage}[2]{{%
\expandafter\ifx\csname l@#1\endcsname\relax
\typeout{** WARNING: IEEEtran.bst: No hyphenation pattern has been}%
\typeout{** loaded for the language `#1'. Using the pattern for}%
\typeout{** the default language instead.}%
\else
\language=\csname l@#1\endcsname
\fi
#2}}
\providecommand{\BIBdecl}{\relax}
\BIBdecl

\bibitem{Atam2011}
A.~P. Dhawan, \emph{Medical Image Analysis}, 2nd~ed.\hskip 1em plus 0.5em minus
  0.4em\relax New Jersey: IEEE Press Series in Biomedical Engineering, 2011,
  ch. Medical Image Modalites: Ultrasound Imaging.

\bibitem{tanter2014ultrafast}
M.~Tanter and M.~Fink, ``Ultrafast imaging in biomedical ultrasound,''
  \emph{IEEE Trans. Ultrason. Ferroelectr. Freq. Control}, vol.~61, no.~1, pp.
  102--119, 2014.

\bibitem{rindal2014understanding}
O.~M.~H. Rindal, J.~P. Asen, S.~Holm, and A.~Austeng, ``Understanding contrast
  improvements from capon beamforming,'' in \emph{Proc. IEEE Int. Ultrasonics
  Symposium (IUS)}, Chicago, IL, USA, 2014, pp. 1694--1697.

\bibitem{Gungor2015}
M.~A. Gungor and I.~Karagoz, ``The homogeneity {MAP} method for speckle
  reduction in diagnostic ultrasound images,'' \emph{Measurement}, vol.~68, pp.
  100--110, 2015.

\bibitem{Michailovich2006}
O.~Michailovich and A.~Tannenbaum, ``Despeckling of medical ultrasound
  images,'' \emph{IEEE Trans. Ultrason. Ferroelectr. Freq. Control}, vol.~53,
  no.~1, pp. 64--78, 2006.

\bibitem{Noble2006_Survey}
J.~A. Noble and D.~Boukerroui, ``Ultrasound image segmentation: A survey,''
  \emph{IEEE Trans. Med. Imag.}, vol.~25, no.~8, pp. 987--1010, 2006.

\bibitem{Pereyra2012}
M.~Pereyra, N.~Dobigeon, H.~Batatia, and J.-Y. Tourneret, ``Segmentation of
  skin lesions in 2-{D} and 3-{D} ultrasound images using a spatially coherent
  generalized {R}ayleigh mixture model,'' \emph{IEEE Trans. Med. Imag.},
  vol.~31, no.~8, pp. 1509--1520, 2012.

\bibitem{MartinoAlessandrini2011}
M.~Alessandrini, S.~Maggio, J.~Poree, L.~D. Marchi, N.~Speciale,
  E.~Franceschini, O.~Bernard, and O.~Basset, ``A restoration framework for
  ultrasonic tissue characterization,'' \emph{IEEE Trans. Ultrason.
  Ferroelectr. Freq. Control}, vol.~58, no.~11, pp. 2344--2360, 2011.

\bibitem{Szabo_book}
T.~L. Szabo, \emph{Diagnostic Ultrasound Imaging: Inside Out}.\hskip 1em plus
  0.5em minus 0.4em\relax Academic Press., 2004.

\bibitem{JamesNg2007}
J.~Ng, R.~Prager, N.~Kingsbury, G.~Treece, and A.~Gee, ``Wavelet restoration of
  medical pulse-echo ultrasound images in an {EM} framework,'' \emph{IEEE
  Trans. Ultrason. Ferroelectr. Freq. Control}, vol.~54, no.~3, pp. 550--568,
  2007.

\bibitem{Jensen1993}
J.~A. Jensen, J.~Mathorne, T.~Gravesen, and B.~Stage, ``Deconvolution of
  \textit{in vivo} ultrasound {B}-mode images,'' \emph{Ultrason. Imaging},
  vol.~15, no.~2, pp. 122--133, Apr. 1993.

\bibitem{Taxt1995}
T.~Taxt, ``Restoration of medical ultrasound imags using two-dimensional
  homomorphic deconvolution,'' \emph{IEEE Trans. Ultrason. Ferroelectr. Freq.
  Control}, vol.~42, no.~4, pp. 543--554, July 1995.

\bibitem{Michailovich2007}
O.~Michailovich and A.~Tannenbaum, ``Blind deconvolution of medical ultrasound
  images: A parametric inverse filtering approach,'' \emph{IEEE Trans. Image
  Process.}, vol.~16, no.~12, pp. 3005--3019, 2007.

\bibitem{Michailovich2005}
O.~Michailovich and D.~Adam, ``A novel approach to the 2-{D} blind
  deconvolution problem in medical ultrasound,'' \emph{IEEE Trans. Med. Imag.},
  vol.~24, pp. 86--104, 2005.

\bibitem{NageOLeary1998}
J.~G. Nagy and D.~P. O'Leary, ``Restoring images degraded by spatially variant
  blur,'' \emph{SIAM J. Sci. Comput.}, vol.~19, no.~4, pp. 1063--1082, Jul.
  1998.

\bibitem{Jirik2008}
R.~Jirik and T.~Taxt, ``Two dimensional blind {B}ayesian deconvolution of
  medical ultrasound images,'' \emph{IEEE Trans. Ultrason. Ferroelectr. Freq.
  Control}, vol.~55, no.~10, pp. 2140--2153, 2008.

\bibitem{Yu2012}
C.~Yu, C.~Zhang, and L.~Xie, ``A blind deconvolution approach to ultrasound
  imaging,'' \emph{IEEE Trans. Ultrason. Ferroelectr. Freq. Control}, vol.~59,
  no.~2, pp. 271--280, 2012.

\bibitem{Bar2004}
L.~Bar, N.~Sochen, and N.~Kiryati, ``Variational pairing of image segmentation
  and blind restoration,'' in \emph{Computer Vision - ECCV 2004}, ser. Lecture
  Notes in Computer Science, T.~Pajdla and J.~Matas, Eds.\hskip 1em plus 0.5em
  minus 0.4em\relax Springer Berlin Heidelberg, 2004, vol. 3022, pp. 166--177.

\bibitem{MumfordShah2011}
L.~Bar, T.~Chan, G.~Chung, M.~Jung, N.~Kiryati, R.~Mohieddine, N.~Sochen, and
  L.~Vese, ``Mumford and shah model and its applications to image segmentation
  andimage restoration,'' in \emph{Handbook of Mathematical Methods in
  Imaging}, O.~Scherzer, Ed.\hskip 1em plus 0.5em minus 0.4em\relax Springer
  New York, 2011, pp. 1095--1157.

\bibitem{Chan2014}
R.~Chan, H.~Yang, and T.~Zeng, ``A two-stage image segmentation method for
  blurry images with poisson or multiplicative gamma noise,'' \emph{SIAM
  Journal on Imaging Sciences}, vol.~7, no.~1, pp. 98--127, 2014.

\bibitem{Ayasso2010}
H.~Ayasso and A.~Mohammad-Djafari, ``Joint {NDT} image restoration and
  segmentation using {G}auss-{M}arkov-{P}otts prior models and variational
  {B}ayesian computation,'' \emph{IEEE Trans. Image Process.}, vol.~19, no.~9,
  pp. 2265--2277, 2010.

\bibitem{Storath2015}
M.~Storath, A.~Weinmann, J.~Frikel, and M.~Unser, ``Joint image reconstruction
  and segmentation using the {P}otts model,'' \emph{Inv. Prob.}, no.~2, pp.
  1--29, 2015.

\bibitem{Paul2013}
G.~Paul, J.~Cardinale, and I.~F. Sbalzarini, ``Coupling image restoration and
  segmentation: A generalized linear model/{B}regman perspective,'' \emph{Int.
  J. Comput. Vis.}, vol. 104, no.~1, pp. 69--93, 2013.

\bibitem{Mignotte2006}
M.~Mignotte, ``A segmentation-based regularization term for image
  deconvolution,'' \emph{IEEE Trans. Image Process.}, vol.~15, no.~7, pp.
  1973--1984, 2006.

\bibitem{Hruska2009}
D.~P. Hruska and M.~L. Oelze, ``Improved parameter estimates based on the
  homodyned {K} distribution,'' \emph{IEEE Trans. Ultrason. Ferroelectr. Freq.
  Control}, vol.~56, no.~11, pp. 2471--2481, 2009.

\bibitem{LarrueNaka2011}
A.~Larrue and J.~Noble, ``Nakagami imaging with small windows,'' in \emph{Proc.
  IEEE International Symposium on Biomedical Imaging (ISBI)}, Chicago, IL, USA,
  2011, pp. 887 -- 890.

\bibitem{Bernard2007}
O.~Bernard, B.~Touil, J.~D'hooge, and D.~Friboulet, ``Statistical modeling of
  the radio-frequency signal for partially- and fully-developed speckle based
  on a generalized {G}aussian model with application to echocardiography,''
  \emph{IEEE Trans. Ultrason. Ferroelectr. Freq. Control}, vol.~54, no.~10, pp.
  2189--2194, 2007.

\bibitem{MartinoAlessandrini2011_AI}
M.~Alessandrini, A.~Palladini, L.~D. Marchi, and N.~Speciale, ``Expectation
  maximization for joint deconvolution and statistics estimation,''
  \emph{Acoustical Imaging}, vol.~30, no.~11, pp. 335--343, 2011.

\bibitem{NZHAO2015}
\BIBentryALTinterwordspacing
N.~Zhao, A.~Basarab, D.~Kouam\'e, and J.-Y. Tourneret, ``Joint segmentation and
  deconvolution of ultrasound images using a hierarchical {B}ayesian model
  based on generalized {G}aussian priors,'' 2015. [Online]. Available:
  \url{http://arxiv.org/abs/1412.2813}
\BIBentrySTDinterwordspacing

\bibitem{Pereyra2013}
M.~Pereyra, N.~Dobigeon, H.~Batatia, and J.-Y. Tourneret, ``Estimating the
  granularity coefficient of a {P}otts-{M}arkov random field within a {M}arkov
  chain {M}onte {C}arlo algorithm,'' \emph{IEEE Trans. Image Process.},
  vol.~22, no.~6, pp. 2385--2397, 2013.

\bibitem{Murray2006}
I.~Murray, Z.~Ghahramani, and D.~J.~C. MacKay, ``{MCMC} for doubly-intractable
  distributions,'' in \emph{In Proc. 22nd Annu. Conf. Uncertainty Artif. Intell
  (UAI)}, Cambridge, MA, USA, 2006, pp. 356--366.

\bibitem{Besag1974}
J.~Besag, ``Spatial interaction and the statistical analysis of lattice
  systems,'' \emph{J. Roy. Stat. Soc. Ser. B}, vol.~36, no.~2, pp. 192--236,
  1974.

\bibitem{Lotfi2010}
L.~Chaari, J.-C. Pesquet, J.-Y. Tourneret, P.~Ciuciu, and A.~Benazza-Benyahia,
  ``A hierarchical {B}ayesian model for frame representation,'' \emph{IEEE
  Trans. Signal Process.}, vol.~58, no.~11, pp. 5560--5571, 2010.

\bibitem{Hastings1970}
W.~K. Hastings, ``Monte {C}arlo sampling methods using {M}arkov chains and
  their applications,'' \emph{Biometrika}, vol.~57, no.~1, pp. 97--109, 1970.

\bibitem{Neal2011}
R.~M. Neal, \emph{Handbook of {M}arkov chain {M}onte {C}arlo}, ser. Chapman and
  Hall/CRC Handbooks of Modern Statistical Methods, 2011, ch. MCMC using
  Hamiltonian dynamics.

\bibitem{hoffman-gelman2013}
M.~D. Hoffman and A.~Gelman, ``The no-{U}-turn sampler: Adaptively setting path
  lengths in {H}amiltonian {M}onte {C}arlo,'' \emph{Journal of Machine Learning
  Research}, vol.~15, no. Apr, pp. 1593--1623, 2014.

\bibitem{beskos2013}
\BIBentryALTinterwordspacing
A.~Beskos, N.~Pillai, G.~Roberts, J.-M. Sanz-Serna, and A.~Stuart, ``Optimal
  tuning of the hybrid monte carlo algorithm,'' \emph{Bernoulli}, vol.~19,
  no.~5A, pp. 1501--1534, 11 2013. [Online]. Available:
  \url{http://dx.doi.org/10.3150/12-BEJ414}
\BIBentrySTDinterwordspacing

\bibitem{GKail2012}
G.~Kail, J.-Y. Tourneret, F.~Hlawatsch, and N.~Dobigeon, ``Blind deconvolution
  of sparse pulse sequences under a minimum distance constraint: A partially
  collapsed {G}ibbs sampler method,'' \emph{IEEE Trans. Signal Process.},
  vol.~60, pp. 2727 -- 2743, Mar. 2012.

\bibitem{Wang2004}
Z.~Wang, A.~C. Bovik, H.~R. Sheikh, and E.~P. Simoncelli, ``Image quality
  assessment: From error visibility to structural similarity,'' \emph{IEEE
  Trans. Image Process.}, vol.~13, no.~4, pp. 600--612, 2004.

\bibitem{Mahloojifar2010}
B.~M. Asl and A.~Mahloojifar, ``Eigenspace-based minimum variance beamforming
  applied to medical ultrasound imaging,'' \emph{IEEE Trans. Ultrason.
  Ferroelectr. Freq. Control}, vol.~57, no.~11, pp. 2381--2390, 2010.

\bibitem{ACJensen2012}
A.~C. Jensen, S.~P. Nasholm, C.-I.~C. Nilsen, A.~Austeng, and S.~Holm,
  ``Applying {T}homson's multitaper approach to reduce speckle in medical
  ultrasound imaging,'' \emph{IEEE Trans. Ultrason. Ferroelectr. Freq.
  Control}, vol.~59, no.~10, pp. 2178--2185, 2012.

\bibitem{Gelman1992}
A.~Gelman and D.~Rubin, ``Inference from iterative simulation using multiple
  sequences,'' \emph{Stat. Sci.}, vol.~7, no.~4, pp. 457--511, 1992.

\bibitem{BeckTeboulle2009}
A.~Beck and M.~Teboulle, ``A fast iterative shringkage-thresholding algorithm
  for linear inverse problems,'' \emph{SIAM J. Imag. Sci.}, no.~1, pp.
  183--202, March 2009.

\bibitem{TwIST_Bioucas-Dias07}
J.~M. Bioucas-Dias and M.~A.~T. Figueiredo, ``A new {T}w{IST}: Two-step
  iterative shrinkage/thresholding algorithms for image restoration,''
  \emph{IEEE Trans. Image Process.}, vol.~16, no.~12, pp. 2992--3004, 2007.

\bibitem{GEM_Bioucas-Dias2006}
J.~M. Bioucas-Dias, ``Bayesian wavelet-based image deconvolution: A {GEM}
  algorithm exploiting a class o heavy-tailed priors,'' \emph{IEEE Trans. Image
  Process.}, vol.~15, no.~4, pp. 937--951, 2006.

\bibitem{Figueredo2007GPSR}
M.~A.~T. Figueiredo, R.~D. Nowak, and S.~J. Wright, ``Gradient projection for
  sparse reconstruction: Application to compressed sensing and other inverse
  problems,'' \emph{IEEE Trans. Image Process.}, vol.~1, no.~4, pp. 586--597,
  Dec. 2007.

\bibitem{Field1996}
J.~A. Jensen, ``{F}ield: A program for simulating ultrasound systems,''
  \emph{Med. Biol. Eng. Comput.}, vol.~34, pp. 351--353, 1996.

\bibitem{Flandrin2004}
G.~R. P.~Flandrin and P.~Goncalves, ``Empirical mode decomposition as a filter
  bank,'' \emph{IEEE Signal Process. Lett.}, vol.~11, no.~2, p. 112–114, 2004.

\bibitem{Pereyra2014}
\BIBentryALTinterwordspacing
M.~Pereyra, ``Proximal {M}arkov chain {M}onte {C}arlo algorithms,'' 2015.
  [Online]. Available: \url{http://arxiv.org/abs/1306.0187}
\BIBentrySTDinterwordspacing

\bibitem{Schreck2013}
\BIBentryALTinterwordspacing
A.~Schreck, G.~Fort, S.~L. Corff, and E.~Moulines, ``A shrinkage-thresholding
  {M}etropolis adjusted {L}angevin algorithm for {B}ayesian variable
  selection,'' 2014. [Online]. Available: \url{http://arxiv.org/abs/1312.5658}
\BIBentrySTDinterwordspacing

\bibitem{Pereyra2016Survey}
M.~Pereyra, P.~Schniter, E.~Chouzenoux, J.-C. Pesquet, J.-Y. Tourneret,
  A.~O.~H. III, and S.~McLaughlin, ``A survey of stochastic simulation and
  optimization methods in signal processing,'' \emph{IEEE J. Sel. Topics Signal
  Process.}, vol.~10, no.~2, pp. 224--241, 2016.

\bibitem{Gonzalez2011}
J.~Gonzalez, Y.~Low, A.~Gretton, and C.~Guestrin, ``Parallel {G}ibbs sampling:
  from colored fields to thin junction trees,'' in \emph{Proc. Artificial
  Intelligence and Statistics (AISTATS)}, Ft. Lauderdale, FL, May 2011.

\bibitem{Girolami_riemannmanifold}
M.~Girolami, B.~Calderhead, and S.~A. Chin, ``Riemann manifold {L}angevin and
  {H}amiltonian {M}onte {C}arlo methods,'' \emph{J. Roy. Stat. Soc. Ser. B},
  vol.~73, pp. 123--214, 2011.

\end{thebibliography}
\end{document}